\documentclass[twoside,11pt]{article}
\usepackage{jmlr2e_old,amssymb,amsmath,color}
\usepackage{url,times}

\usepackage{epsfig}
\usepackage{graphics}

\RequirePackage{natbib}

\newcommand\pscal[2]{ \left\langle #1 , #2 \right\rangle }

\newcommand{\mysec}[1]{Section~\ref{sec:#1}}
\newcommand{\myapp}[1]{Appendix~\ref{app:#1}}
\newcommand{\eq}[1]{Eq.~(\ref{eq:#1})}
\newcommand{\myfig}[1]{Figure~\ref{fig:#1}}

\newcommand{\BEAS}{\begin{eqnarray*}}
\newcommand{\EEAS}{\end{eqnarray*}}
\newcommand{\BEA}{\begin{eqnarray}}
\newcommand{\EEA}{\end{eqnarray}}
\newcommand{\BEQ}{\begin{equation}}
\newcommand{\EEQ}{\end{equation}}
\newcommand{\BIT}{\begin{itemize}}
\newcommand{\EIT}{\end{itemize}}
\newcommand{\BNUM}{\begin{enumerate}}
\newcommand{\ENUM}{\end{enumerate}}
\newcommand{\BA}{\begin{array}}
\newcommand{\EA}{\end{array}}

\newcommand{\var}{\mathop{\rm var}}
\newcommand{\cov}{{\rm cov}}
\newcommand{\Diag}{\mathop{\rm Diag}}

\newcommand{\rb}{\mathbb{R}}

\newcommand{\tr}{{\rm tr}}
\newcommand{\idm}{I}

\def \E{{\mathbb E}}
\def \P{{\mathbb P}}

\def \hS{\hat{\S}}
\def \J { \mathbf{J} }
\def \S{ \Sigma }

\def \f{ \mathbf{f}}
\def \fJ{ \mathbf{f}_\J  }
\def \fj{ \mathbf{f}_j }
\def \g{ \mathbf{g}}
\def \gJ{ \mathbf{g}_\J  }
\def \gj{ \mathbf{g}_j }
\def \w{ \mathbf{w} }
\def \b{ \mathbf{b}  }
\def \h{ \mathbf{h}  }
\def \sm{ \sigma_{\min}^2}
\def \sM{ \sigma_{\max}^2}
\def \wJ{ \mathbf{w}_\J  }
\def \wj{ \mathbf{w}_j }
\def \F {\mathcal{F}}
\def \Fi { {\mathcal{F}_i} }
\def \Fj {{\mathcal{F}_j}}
\def \FJ {{\mathcal{F}_\J}}

\def \X{ \bar{X} }
\def \Y{ \bar{Y} }

\newcounter{hyp}
\newenvironment{hyp}{\refstepcounter{hyp}\begin{itemize}
  \item[({\bf{A}\arabic{hyp}})]}{\end{itemize}}
  
\newcommand{\hypref}[1]{({\bf{A}\ref{hyp:#1}})}
\newcommand{\hypreff}[2]{({\bf{A}\ref{hyp:#1}-\ref{hyp:#2}})}

\title{Consistency of the Group Lasso\\ and Multiple Kernel Learning}

\author{\name Francis R. Bach \email francis.bach@mines.org \\
    \addr INRIA - Willow project  \\
     D\'epartement d'Informatique, Ecole Normale Sup\'erieure\\
     45, Rue d'Ulm\\
     75230 Paris, France
       }

\begin{document}
\maketitle
 
\begin{abstract}
We consider the least-square regression problem with regularization by a block $\ell_1$-norm, i.e., a
sum of  Euclidean norms over spaces of dimensions larger than one. 
This problem, referred to as the group Lasso, extends the usual regularization by the $\ell_1$-norm where all spaces have dimension one, where it
is commonly referred to as the Lasso. In this paper, we study the asymptotic model
consistency of the group Lasso. We derive necessary and sufficient
conditions for the consistency of group Lasso under practical assumptions, such as model misspecification. When the linear
predictors and  Euclidean norms are replaced by functions and reproducing kernel Hilbert norms, the problem is usually referred to
as multiple kernel learning and is commonly used for learning from heterogeneous data sources and for
non linear variable selection. Using tools from functional analysis, and in particular covariance operators, 
we extend the consistency results to this infinite dimensional case and also propose an
adaptive scheme
to
obtain a consistent model estimate, even when the necessary 
 condition required for the non adaptive scheme is not satisfied.
\end{abstract}


\section{Introduction}
Regularization has emerged as a dominant theme in machine 
learning and  statistics. It provides an intuitive and principled tool
for learning from high-dimensional data. 
Regularization by squared Euclidean norms or squared Hilbertian norms has been thoroughly studied in various settings, from approximation
theory to statistics, leading to efficient practical algorithms based on linear algebra
 and very general theoretical consistency results~\citep{tikhonov,wahba,hastie,steinwart,smale}. 
 
In recent years, regularization by non Hilbertian norms  has generated considerable interest in linear supervised learning,
where the goal is to predict a response as a linear function of  covariates; 
 in particular, regularization by the $\ell_1$-norm (equal to the sum of absolute values), a method commonly referred to as the
 \emph{Lasso}~\citep{lasso,osborne}, allows to perform variable selection. However, 
 regularization by non Hilbertian norms cannot be solved empirically
 by simple linear algebra and instead leads to general convex optimization problems and much of the early effort
 has been dedicated to algorithms to solve the optimization problem efficiently. In particular, the \emph{Lars}
 algorithm of~\citet{lars} allows to find the entire regularization path (i.e., the set of solutions for all values
 of the regularization parameters) at the cost of a single matrix inversion. 
 
As the consequence of the optimality conditions, 
regularization by the $\ell_1$-norm  leads to \emph{sparse} solutions, i.e., loading vectors with many zeros. Recent works
\citep{Zhaoyu,yuanlin,zou,martin} have looked precisely
at the model consistency of the Lasso, i.e., if we know that the data were generated from a sparse loading vector, does
the Lasso actually recover it when the number of observed data points grows? In the case of a fixed number of covariates, the Lasso does recover
the sparsity pattern if and only if a certain simple condition on the generating covariance matrices is verified~\citep{yuanlin}. 
In particular, in low correlation settings,
the Lasso is indeed consistent. However, in presence of strong correlations, the Lasso cannot be consistent, shedding light on
potential problems of such procedures for variable selection. Adaptive versions where data-dependent weights
are added to the $\ell_1$-norm  then allow
to keep the consistency in all situations~\citep{zou}.

A related Lasso-type procedure is the \emph{group Lasso}, where the covariates are assumed to be clustered in groups, and instead of
summing the absolute values of each individual loading, the sum of Euclidean norms of the loadings in each group is used. 
Intuitively, this should drive all the weights in one group to zero \emph{together}, and thus lead to group selection~\citep{grouped}. 
In \mysec{grouplasso}, we extend the consistency results of the Lasso to the group Lasso, showing that similar correlation
conditions are necessary and sufficient conditions for consistency. The passage from groups of size
one to groups of larger sizes leads however to a slightly weaker result as we can not get a single
necessary and sufficient condition (in \mysec{refined}, we show that the stronger result similar to the Lasso is not true
as soon as one group has dimension larger than one).
Also, in our proofs, we relax the  assumptions usually
made for such consistency results, i.e., that the model is completely well-specified (conditional expectation of the response
which is linear in the
covariates and constant conditional variance). In the context of \emph{misspecification}, which is a common situation when
applying methods such as the ones presented in this paper, we simply prove convergence
to the best linear predictor (which is assumed to be sparse), both in terms of loading vectors and sparsity patterns.

The group Lasso essentially replaces groups of size one by groups of size larger than one. It is natural in this context
to allow the size of each group to grow unbounded, i.e., to replace the sum of Euclidean norms by a sum 
of appropriate Hilbertian norms. When the Hilbert spaces are reproducing kernel Hilbert spaces (RKHS),
this procedure turns out to be equivalent to learn the best convex combination 
of a set of basis kernels, where each kernel corresponds to one  Hilbertian norm used for regularization~\citep{skm}.
This framework, referred to as \emph{multiple kernel learning}~\citep{skm}, has applications in kernel selection, data fusion from
heterogeneous data sources and non linear variable selection~\citep{genomic_fusion}. In this latter case,  multiple kernel learning
can exactly be seen as variable selection in a \emph{generalized additive model}~\citep{hastie_GAM}.
We extend  the consistency results of the group Lasso to this non parametric  case, by using covariance operators and
appropriate notions of functional analysis. These notions allow to carry out the analysis
entirely in \emph{``primal/input''} space, while the algorithm has to work in \emph{``dual/feature'' } space
to avoid infinite dimensional optimization.
Throughout the paper, we will always go back and forth between primal and dual formulations, primal formulation for analysis
and dual formulation for algorithms.

The paper is organized as follows: in \mysec{grouplasso}, we present the consistency results for the group Lasso, while in \mysec{mklsec},
we extend these to Hilbert spaces. Finally, we present the adaptive schemes in \mysec{adaptive} and illustrate our
set of results with simulations on synthetic examples in \mysec{simulations}.

\section{Consistency of the Group Lasso}
\label{sec:grouplasso}
We consider the problem of predicting a response $Y \in \rb$ from covariates $X \in \rb^{p}$, where
$X$ has a block structure with $m$ blocks, i.e.,
$X = (X_1^\top,\dots,X_m^\top)^\top$ with each $X_j \in \rb^{p_j}$, $j=1,\dots,1m$, and $\sum_{j=1}^m p_j = p$. Unless otherwise specified, $\|X\|$ will denote the Euclidean norm of a vector $X$.
The only assumptions that we make on the joint distribution $P_{XY}$ of $(X,Y)$ are the following:

\begin{hyp}
\label{hyp:var}
$X$ and $Y$ have finite fourth order moments: $\E  \| X\|^4 < \infty$ and
$\E \| Y\|^4 < \infty$.
\end{hyp}

\begin{hyp}
\label{hyp:inv}
 The joint covariance matrix  $\S_{XX} = \E XX^\top -
 (\E X) ( \E X)^\top \in \rb^{ p \times p} $ is invertible.
\end{hyp}

\begin{hyp}
\label{hyp:model}
We let  $ (\w,\b) \in \rb^p \times \rb $  denote any minimizer of $\E (Y-X^\top w -b)^2$. We assume
that $\E( (Y - \w^\top X -\b)^2  | X)$ is almost surely greater than $\sm>0$.
We let denote $\J = \{ j, \wj \neq 0\}$
the sparsity pattern of $\w$.\footnote{
 Note that throughout this paper, we use boldface fonts for population quantities.
}
\end{hyp}
The assumption \hypref{model} does not state that $\E(Y|X)$ is an affine function of $X$ and that the conditional
variance is constant, as it is commonly done in
most works dealing with consistency for linear supervised learning. We simply assume that  given the best affine
predictor of $Y$ given $X$ (defined by $\w \in \rb^p$ and $\b \in \rb$), there is still a strictly positive amount of variance in $Y$. If \hypref{inv} is satisfied, then 
the full loading vector
$\w$ is uniquely defined and is equal to
 $\w = (\S_{XX}^\top)^{-1} \S_{XY}$, where $\S_{XY} = \E (XY) - (\E X)(\E Y) \in \rb^p$. Note that throughout this paper, we do include a non regularized constant term $b$ but since we use a square loss it will optimized out in closed form by centering the data. Thus all our consistency statements will be stated only for the loading vector $w$; corresponding results for $b$ then immediately follow.

We often use the notation
$\varepsilon = Y - \w^\top X - \b $.
In terms of   covariance matrices, our assumption \hypref{model} leads to:
$\S_{\varepsilon\varepsilon|X} = \E( \varepsilon\varepsilon|X)  \geqslant \sm $ and $\S_{\varepsilon X}
=  0$
(but $\varepsilon$ might not in general be independent from $X$).

\paragraph{Applications of grouped variables}
In this paper, we assume that the groupings of the univariate variables is known and fixed, i.e., the group structure is given and we wish to achieve sparsity at the level of groups. This has numerous applications, e.g., in speech and signal processing, where groups may represent different frequency bands~\citep{speech}, or bioinformatics~\citep{genomic_fusion} and computer vision~\citep{Varma07c,graphkernel} where each group may correspond to different data sources or data types. Note that those different data sources are sometimes referred to as \emph{views}~\citep[see, e.g.,][]{views}.

Moreover, we always assume that the number $m$ of groups is fixed and finite. Considering cases where $m$ is allowed to grow with the number
of observed data points, in the line of~\citet{yuinfinite}, is outside the scope of this paper.

\paragraph{Notations} Throughout this paper, we consider the block covariance matrix $\S_{XX}$ with $m^2$ blocks
$\Sigma_{X_i X_j}$, $i,j=1,\dots,m$. We refer to the submatrix composed of all blocks indexed by sets $I$, $J$ as $\Sigma_{X_I X_J}$.
Similarly, our loadings are vectors defined following block structure, $w =(w_1^\top,\dots,w_m^\top)^\top$ and we denote
$w_I$ the elements indexed by $I$. Moreover we denote $1_q$ the vector in $\rb^q$ with constant components equal to one, and $\idm_q$ the identity matrix of size $q$.

\subsection{Group Lasso}
We consider \emph{independent and identically distributed}
 (i.i.d.) data $(x_i,y_i) \in \rb^p \times \rb$, $i=1,\dots,n$, sampled from $P_{XY}$ and
the data are given in the form of matrices $\Y \in \rb^n$ and $\X \in \rb^{n \times p}$ and we write
$\X = (\X_1,\dots,\X_m)$ where each $\X_j \in \rb^{ n \times p_j}$ represents the data associated with group $j$. Throughout this paper, we make the same i.i.d. assumption; dealing with 
non identically distributed or dependent data and extending our results in those situations are left for future research.

We consider the following optimization problem:
$$\min_{w \in \rb^p, \ b \in \rb } \ 
\frac{1}{2n} \| \Y - \X w - b 1_n  \|^2 + \lambda_n \sum_{j=1}^m d_j \|w_j\|,
$$
where $d \in \rb^m $ is a vector of strictly positive fixed weights. Note that considering
weights in the block $\ell_1$-norm is important in practice as those have an influence regarding the consistency of the estimator
(see \mysec{adaptive} for further details).
Since $b$ is not regularized, we can minimize in closed form with respect to $b$, by setting $b = \frac{1}{n} 1_n^\top(  \Y -  \X w )$. This leads to the following reduced optimization problem in $w$:
\BEQ
\label{eq:problem}
\min_{w \in \rb^p } \ 
\frac{1}{2} \hS_{YY} -  \hS_{XY}^\top w + \frac{1}{2} w^\top
 \hS_{XX} w + \lambda_n \sum_{j=1}^m d_j \|w_j\|,
\EEQ
 where $\hS_{YY} = \frac{1}{n} \Y^\top \Pi_n   \Y $, $\hS_{XY}
 = \frac{1}{n} \X^\top \Pi_n \Y$ and $\hS_{XX} =  \frac{1}{n} \X^\top \Pi_n  \X$ are empirical  covariance matrices (with the centering matrix $\Pi_n$ defined as
 $\Pi_n = \idm_n - \frac{1}{n} 1_n 1_n^\top$).
We denote $\hat{w}$ any minimizer of
\eq{problem}. We refer to $\hat{w}$ as the \emph{group Lasso} estimate\footnote{We use the convention that all
``hat'' notations correspond to data-dependent and thus $n$-dependent quantities, so we do not need the
explicit dependence on $n$.}. Note that with probability tending to one, if \hypref{inv} is satisfied (i.e.,
if $\S_{XX}$ is invertible), there is a unique minimum. 

 Problem (\ref{eq:problem}) is a non-differentiable convex optimization problem, for which classical tools
 from  convex optimization~\citep{boyd} lead to the following optimality conditions
 (see proof by~\citet{grouped} and in \myapp{opt}): 
\begin{proposition}
\label{prop:opt}
A vector $w \in \rb^p$ with sparsity pattern $J = J(w) = \{ j, \ w_j \neq 0\}$ is optimal for problem (\ref{eq:problem}) if
and only if
\BEA
\label{eq:opt1}
 \forall j \in J^c, && \ \left\| \hS_{X_j X} w -  \hS_{X_j Y} \right\| \leqslant \lambda_n d_j, \\
\label{eq:opt2}
  \forall j \in J, &&\  \hS_{X_j X} w - \hS_{X_j Y} = - w_j \frac{ \lambda_n d_j}{\| w_j\|}.
\EEA
\end{proposition}

\subsection{Algorithms}
Efficient \emph{exact} algorithms exist for the regular Lasso, i.e., for the case where all group dimensions
$p_j$ are equal to one. They are based on
the piecewise linearity of the set of solutions as a function of the regularization parameter $\lambda_n$~\citep{lars}. For
the group Lasso, however, the path is only piecewise differentiable, and following such
a path is not as efficient as for the Lasso. Other algorithms have been designed to solve
problem~(\ref{eq:problem}) for a single value of $\lambda_n$, in the original group Lasso setting~\citep{grouped} and
in the multiple kernel setting~\citep{skm,bach_thibaux,sonnenburg,rakoto}.
In this paper, we study path consistency of the group Lasso and of multiple kernel learning, and
in simulations we use the publicly available code for the algorithm of~\citet{bach_thibaux},
 that computes an approximate but entire
 path, by following
the piecewise smooth path with predictor-corrector methods.

\subsection{Consistency Results}

We consider the following two conditions:
\BEQ
\label{eq:condition}
 \max_{ i \in \J^c } \frac{1}{d_i} \left\|  \S_{X_i X_\J  } \S_{X_\J  X_\J }^{-1}
\Diag  ( d_j/ \| \w_j \|) \wJ  \right\| < 1,
\EEQ
\BEQ
\label{eq:condition-weak}
 \max_{ i \in \J^c } \frac{1}{d_i} \left\|  \S_{X_i X_\J  } \S_{X_\J  X_\J }^{-1}
\Diag  ( d_j/ \| \w_j \|) \wJ  \right\| \leqslant 1,
\EEQ
where $\Diag  ( d_j/ \| \w_j \|)$ denotes the block-diagonal matrix (with block sizes $p_j$) in which each diagonal block is equal to 
$ \frac{d_j}{\|\w_j\|} I_{p_j}$ (with
$I_{p_j}$ the identity matrix of size $p_j$), and $\wJ$ denotes the concatenation of the loading vectors indexed
by $\J$. Note that the conditions involve the covariance between all active groups $X_j$, $j \in \J$ and all non active groups $X_i$, $i\in\J^c$.

 These are conditions on both the input (through the joint
covariance matrix $\S_{XX}$) and on the weight vector $\w$. Note that, 
when all blocks have size 1, this corresponds to the 
conditions derived for the Lasso~\citep{Zhaoyu,yuanlin,zou}. Note also the difference between the \emph{strong condition}
(\ref{eq:condition}) and the \emph{weak condition}
(\ref{eq:condition-weak}). For the Lasso, with our assumptions, \citet{yuanlin} has shown that
the strong condition (\ref{eq:condition})  is necessary and sufficient for path consistency of the Lasso;
i.e., the path of solutions consistently contains
an estimate which is both consistent for the $2$-norm (regular consistency) and the $\ell_0$-norm (consistency
of patterns), if and only if condition (\ref{eq:condition})  is satisfied.

 In the case
of the group Lasso, even with a finite fixed number of groups, our results are not as strong, as we can only get
the strict condition as sufficient and the weak condition as necessary. In \mysec{refined}, we show that this cannot
be improved in general.
More precisely the following theorem, proved in \myapp{theo1}, shows that if the condition (\ref{eq:condition})  is satisfied, any
regularization parameter that satisfies a certain decay conditions will lead to a consistent estimator; thus the strong condition
(\ref{eq:condition}) is sufficient for path consistency:
\begin{theorem}
\label{theo:theo1}
Assume \hypreff{var}{model}.
If condition (\ref{eq:condition}) is satisfied, then for any sequence $\lambda_n$ such that
 $\lambda_n  \to 0$ and $\lambda_n n^{1/2} \to + \infty$, then the group Lasso estimate $\hat{w}$ defined in \eq{problem}
 converges in probability
 to $\w$ and the group sparsity pattern $J(\hat{w}) = \{ j, \hat{w}_j \neq 0 \}$ converges in probability to $\J$ (i.e.,
 $\P( J(\hat{w})  = \J ) \to 1$).
\end{theorem}

The following theorem, proved in \myapp{theo2}, states that if there is a consistent solution on the path, then the weak condition
(\ref{eq:condition-weak}) must  be satisfied.

\begin{theorem}
\label{theo:theo2}
Assume  \hypreff{var}{model}.
If there exists a (possibly data-dependent) sequence $\lambda_n$ such 
that $\hat{w}$ converges to $\w$ and $J(\hat{w})$ converges to $ \J$ 
 in probability, then  condition~(\ref{eq:condition-weak})  is  satisfied.
\end{theorem}
 
On the one hand, Theorem~\ref{theo:theo1} states that under  the ``low correlation between variables in $\J$ and variables in $\J^c$'' 
condition (\ref{eq:condition}), the group Lasso is indeed consistent.
 On the other hand, the result (and the similar one for the Lasso) is rather disappointing regarding the  
 applicability of the group Lasso as a practical group
  selection method, as Theorem~\ref{theo:theo2} states that 
  if the weak correlation condition (\ref{eq:condition-weak}) is not satisfied, we cannot have consistency. 
  
Moreover, this is to be contrasted with a thresholding procedure of the joint least-square estimator, which is also consistent
with no conditions (but the invertibility of $\S_{XX}$),
if the threshold is properly chosen (smaller than the smallest norm $\|\w_j\|$ for $j \in \J$ or with appropriate decay conditions). 
However, the Lasso and group
Lasso do not have to set such a threshold; moreover, further analysis show that the Lasso has additional advantages over regular regularized least-square procedure~\citep{yuinfinite}, and empirical evidence shows that in the finite sample case, they do perform better~\citep{lasso},
in particular in the case where the number $m$ of groups is allowed to grow. In this paper we focus on the extension
from uni-dimensional groups to multi-dimensional groups for finite number of groups $m$ and leave the possibility of letting $m$ grow with $n$ for future research.

Finally, by looking carefully at condition (\ref{eq:condition}) and (\ref{eq:condition-weak}), we can see 
  that if we were to increase the weight $d_j$ for $j \in \J^c$ and decrease the weights otherwise, we 
  could always be consistent: this however requires the (potentially empirical) knowledge of $\J$ and this is exactly the idea behind the adaptive scheme that we present in \mysec{adaptive}. Before looking at these extensions, we discuss in the next Section,   qualitative differences between our results and the corresponding ones for the Lasso.

\subsection{Refinements of Consistency Conditions}
\label{sec:refined}

Our current results state that the strict condition (\ref{eq:condition}) is sufficient for joint consistency of the group Lasso,
while the weak condition (\ref{eq:condition-weak}) is only necessary. When all groups have dimension one, then the strict condition turns out to be also necessary~\citep{yuanlin}. 

 The main technical reason for those differences is that in dimension one,
the set of vectors of unit norm is finite (two possible values), and thus regular squared norm consistency leads to estimates of the signs  of the loadings
(i.e., their normalized versions $\hat{w}_j/\|\hat{w}_j\|$)
which are ultimately constant.
When groups have size larger than one, then $\hat{w}_j/\|\hat{w}_j\|$ will not be ultimately constant (just consistent) and this
added dependence on data leads to the following refinement of 
Theorem~\ref{theo:theo1} (see proof in \myapp{theo1-refined}):
\begin{theorem}
\label{theo:theo1-refined}
Assume \hypreff{var}{model}.
Assume the weak condition (\ref{eq:condition-weak}) is satisfied and that for all $i 
\in \J^c$ such that $\frac{1}{d_i} \left\|  \S_{X_i X_\J  } \S_{X_\J  X_\J }^{-1}
\Diag  ( d_j/ \| \w_j \|) \wJ  \right\| =1$, we have
\BEQ
\label{eq:new}
\Delta^\top \S_{ X_\J X_i  }
 \S_{X_{i} X_\J  } \S_{X_\J  X_\J  }^{-1} \Diag \left[ d_j / \|\w_j\| \left( \idm_{p_j} - \frac{ \w_j \w_j^\top}{ \w_j^\top \w_j}
\right) \right]  \Delta > 0,
\EEQ
with $\Delta =  - \S_{X_\J X_\J  }^{-1} \Diag( d_j / \|\w_j\|) \w_\J$.
 Then for any sequence $\lambda_n$ such that
 $\lambda_n   \to 0$ and $\lambda_n n^{1/4} \to + \infty$, then the group Lasso estimate $\hat{w}$ defined in \eq{problem}
 converges in probability
 to $\w$ and the group sparsity pattern $J(\hat{w}) = \{ j, \hat{w}_j \neq 0 \}$ converges in probability to $\J$.
\end{theorem}
 This theorem is of lower practical significance than Theorem~\ref{theo:theo1} and Theorem~\ref{theo:theo2}. It merely shows
that the link between strict/weak conditions and sufficient/necessary conditions are in a sense tight (as soon as there exists $j \in \J$
such that $p_j>1$, 
it is easy to exhibit
examples where  \eq{new} is or is not satisfied). The previous theorem does not contradict the fact that condition
(\ref{eq:condition}) is necessary for path-consistency in the Lasso case: indeed, if $w_j$ has dimension one, then
$\idm_{p_j} - \frac{ \w_j \w_j^\top}{ \w_j^\top \w_j}$ is always equal to zero, and thus \eq{new} is never satisfied.
Note that when condition (\ref{eq:new}) is an equality, we could still refine 
the condition by using higher orders in the asymptotic expansions presented in \myapp{theo1-refined}.

We can also further refined the \emph{necessary} condition results in Theorem~\ref{theo:theo2}:
as stated in Theorem~\ref{theo:theo2}, the group Lasso estimator may be both consistent in terms of norm and sparsity patterns only if the  condition (\ref{eq:condition-weak})  is satisfied. However, if we require only the consistent sparsity pattern estimation, then we may allow the convergence of the regularization parameter $\lambda_n$ to a strictly positive limit $\lambda_0$. In this situation, we may consider the following population problem: 
\BEQ
\label{eq:problem-lambda0}
\min_{w \in \rb^p }     \frac{1}{2} (w-\w)^\top
 \S_{XX}  (w-\w) + \lambda_0 \sum_{j=1}^m d_j \|w_j\|.
\EEQ
If there exists $\lambda_0 >0$ such that the solution has the correct sparsity pattern, then the group Lasso estimate with $\lambda_n \to \lambda_0$, will have a consistent sparsity pattern. The following proposition, which can be proved with standard M-estimation arguments, make this precise:
\begin{proposition}
\label{prop:refined}
 Assume \hypreff{var}{model}.
If $\lambda_n$ tends to $\lambda_0>0$, then the group Lasso estimate $\hat{w}$ is sparsity-consistent if and only if the solution of \eq{problem-lambda0} has the correct sparsity pattern.
\end{proposition}

Thus, even when condition  (\ref{eq:condition-weak})   is not satisfied, we may have consistent estimation of the sparsity pattern but inconsistent estimation of the loading vectors. We provide in \mysec{simulations} such examples.

\subsection{Probability of Correct Pattern Selection}
\label{sec:proba}
In this section, we focus on regularization parameters that tend to zero, at the rate $n^{-1/2}$, i.e., $\lambda_n = \lambda_0 n^{-1/2}$ with $\lambda_0>0$. For this particular setting, we can actually compute the limit of the probability of correct pattern selection (proposition proved in Appendix~\ref{app:probab}). Note that in order to obtain a simpler result, we assume constant conditional variance of $Y$ given $\w^\top X$:
\begin{proposition}
\label{prop:probab}
Assume  \hypreff{var}{model} and $\var(Y|\w^\top x) = \sigma^2$ almost surely. Assume  moreover $\lambda_n = \lambda_0 n^{-1/2}$ with $\lambda_0>0$. Then, the group Lasso
$\hat{w}$ converges in probability to $\w$ and  the probability of correct sparsity pattern selection has the following limit:
\BEQ
\label{eq:probab}
\P\left( \max_{i \in \J^c} \frac{1}{d_i}
\left \| \frac{\sigma}{ \lambda_0 }  
 { {t}}_i -    \S_{X_i X_\J  } \S_{X_\J  X_\J }^{-1}
\Diag  ( \frac{d_j}{\| \w_j \|} ) \wJ  \right\| \leqslant 1 \right),
\EEQ
where $ { {t}}$ is normally distributed with mean zero and  covariance matrix
  $\S_{X_{\J^c} X_{\J^c} | X_\J}
  = \S_{X_{\J^c} X_{\J^c}}-
\S_{ X_{\J^c} X_\J} \S_{X_\J X_\J}^{-1} \S_{X_\J  X_{\J^c}}$ (which is the conditional covariance matrix of
$ X_{\J^c}$ given $X_\J$).
\end{proposition}
The previous theorem states that the probability of correct selection tends to the mass under a non degenerate multivariate distribution of the intersection of   cylinders. 
Under our assumptions, this set is never empty and thus the limiting probability is strictly positive, i.e., there is (asymptotically) always a positive probability of estimating the correct pattern of groups.

Moreover, additional insights may be gained from Proposition~\ref{prop:probab}, namely in terms of the dependence on $\sigma$, $\lambda_0$ and the tightness of the consistency conditions. First, when $\lambda_0$ tends to infinity, then the limit defined in \eq{probab} tends to one if the strict consistency condition  (\ref{eq:condition}) is satisfied, and tends to zero if one of the conditions is strictly not met. This corroborates the results of Theorem~\ref{theo:theo1} and~\ref{theo:theo2}. Note however, that only an extension of Proposition~\ref{prop:probab} to $\lambda_n$ that may deviate from a $n^{-1/2}$ would actually lead to a proof of  Theorem~\ref{theo:theo1}, which is a subject of ongoing research.

Finally, \eq{probab} shows that $\sigma$ has a smoothing effect on the probability of correct pattern selection, i.e., if condition  (\ref{eq:condition}) is satisfied, then this probability is a decreasing function of $\sigma$ (and an increasing function of $\lambda_0$). Finally, the stricter the inequality in \eq{condition}, the larger the probability of correct rank selection, which is illustrated in \mysec{simulations} on synthetic examples.

\subsection{Loading Independent Sufficient Condition}

Condition (\ref{eq:condition}) depends on the loading vector $\w$ and on the sparsity pattern $\J$, which are both a priori
unknown. In this section,
we consider sufficient conditions that do not depend on the loading vector, but only on the sparsity pattern $\J$ and of
course on
the covariance matrices. The following condition is sufficient for consistency of
the group Lasso, for all possible  loading vectors $\w$ with sparsity pattern $\J$:

\BEQ
\label{eq:condition2}
C(\Sigma_{XX},d,\J) =
 \max_{ i \in \J^c } \ \ \ \ \max_{\forall j \in \J, \ \| u_j\| = 1}    \left\| \frac{1}{d_i} \S_{X_i X_\J  } \S_{X_\J  X_\J }^{-1}
\Diag ( d_j ) u_\J  \right\| < 1.
\EEQ

As opposed to the Lasso case, $C(\Sigma_{XX},d,\J)$
cannot be readily computed in closed form, but we have the following upper bound: 
$$ C(\Sigma_{XX},d,\J)
\leqslant 
 \max_{ i \in \J^c } \frac{1}{d_i} \sum_{j \in \J}
   d_j \left\|   \sum_{k \in \J}   \S_{X_i X_{k} }  
   \left(\S_{X_\J  X_\J }^{-1}\right)_{kj} \right\|,
$$
where for a matrix $M$, $\| M\|$ denotes its maximal singular value (also known as its spectral norm).
This
leads to the following sufficient condition for consistency of the group Lasso~\citep[which extends the condition
of][]{yuanlin}:

\BEQ
\label{eq:condition3}
\max_{ i \in \J^c } \frac{1}{d_i} \sum_{j \in \J}
   d_j \left\|   \sum_{k \in \J}   \S_{X_i X_{k} }  
   \left(\S_{X_\J  X_\J }^{-1}\right)_{kj} \right\|   < 1.
\EEQ
Given a set of weights $d$, better sufficient conditions than
\eq{condition3} may be obtained by solving a semidefinite programming problem~\citep{boyd}:
\begin{proposition} The quantity
$ \displaystyle
\max_{\forall j \in \J, \ \| u_j\| = 1}    \left\|  \S_{X_i X_\J  } \S_{X_\J  X_\J }^{-1}
\Diag ( d_j ) u_\J  \right\|^2$ is upperbounded
by 
\BEQ
\label{eq:cvx}
\max_{ M \succcurlyeq 0 ,\  \tr M_{ii} = 1 }  \tr M   \left(
  \Diag(d_j)\S_{X_\J  X_\J }^{-1} \S_{X_\J  X_i }
\S_{X_i X_\J  } \S_{X_\J  X_\J }^{-1} \Diag(d_j) \right),
\EEQ
where $M$ is a matrix defined by blocks following the block structure of $\S_{X_\J X_\J}$. Moreover, 
the bound is also equal to
$$
 \min_{\lambda \in \rb^m, \  \Diag(d_j)\S_{X_\J  X_\J }^{-1} \S_{X_\J  X_i }
\S_{X_i X_\J  } \S_{X_\J  X_\J }^{-1} \Diag(d_j) \preccurlyeq \Diag(\lambda) } \sum_{j=1}^m\lambda_j.
$$
\end{proposition}
\begin{proof}
We let denote $M = uu^\top \succcurlyeq 0 $. Then if all $u_j$ for $j \in \J$ have norm 1, then we have
$\tr M_{jj}=1$ for all $j \in \J$. This implies the convex relaxation. The second problem is easily obtained as the convex dual
of the first problem~\citep{boyd}.
\end{proof}
Note that for the Lasso, the convex bound in \eq{cvx} is tight and leads to the bound
 given above in \eq{condition3}~\citep{yuanlin,martin}.
For the Lasso, \citet{Zhaoyu} consider several particular patterns of dependencies using \eq{condition3}. Note that this condition
(and not the condition in \eq{condition2}) is independent from the dimension and thus does not readily
lead to rules of thumbs allowing to set the weight
 $d_j$ as a function of the dimension $p_j$; several rules of thumbs have been suggested, that loosely
depend on the dimension on the blocks, in the context of the linear group Lasso~\citep{grouped} 
or multiple kernel learning~\citep{bach_thibaux}; we argue in this paper, that weights should also depend
on the response as well (see \mysec{adaptive}).

\subsection{Alternative Formulation of the Group Lasso}
\label{sec:alt}
Following~\citet{skm}, we can instead consider regularization by the square of the block $\ell_1$-norm:
$$
\min_{w \in \rb^p, \ b \in \rb } \ 
\frac{1}{2n} \| \Y - \X w - b 1_n  \|^2 + \frac{1}{2} \mu_n \left( \sum_{j=1}^m d_j \|w_j\| \right)^2.
$$
This leads to the same path of solutions, but it is better behaved because each variable which is not zero is still
regularized by the squared norm. The alternative version has also two advantages: (a) it has very close links
to more general frameworks for learning the kernel matrix from data~\citep{gert}, and (b) it 
is essential in our proof of consistency in the functional case.
We also get the equivalent formulation to \eq{problem}, by minimizing in closed form with respect to $b$, to obtain:
\BEQ
\label{eq:problem-alt}
\min_{w \in \rb^p} \ 
\frac{1}{2} \hS_{YY} -  \hS_{YX} w + \frac{1}{2} w^\top
 \hS_{XX} w +  \frac{1}{2} \mu_n \left( \sum_{j=1}^m d_j \|w_j\| \right)^2.
\EEQ
The following proposition gives the optimality conditions for the convex optimization problem defined
in \eq{problem-alt} (see proof in \myapp{opt-alt}):
\begin{proposition}
\label{prop:opt-alt}
A vector $w \in \rb^p$ with sparsity pattern $J = \{ j, \ w_j \neq 0\}$ is optimal for problem (\ref{eq:problem-alt}) if
and only if
\BEA
\label{eq:opt1-alt}
\forall j \in J^c ,& &   \left\| \hS_{X_j X} w -  \hS_{X_j Y}  \right\|  \leqslant  \mu_n d_j
 \textstyle \left(\sum_{i=1}^n d_i \|w_i\| \right) \displaystyle ,  \\
\label{eq:opt2-alt}
\forall j \in J ,& &   
  \hS_{X_{j} X }w  -  \hS_{X_{j} Y} =  - \mu_n  \textstyle \left(\sum_{i=1}^n d_i \|w_i\| \right) \displaystyle  \frac{d_j w_j}{ \|w_j\|}.
\EEA
\end{proposition}
Note the correspondence at the optimum between optimal solutions of the two optimization problems in \eq{problem} and
\eq{problem-alt} through  $\lambda_n =\mu_n \textstyle \left(\sum_{i=1}^n d_i \|w_i\| \right)$.
As far as consistency results are concerned, Theorem~\ref{theo:theo2} immediately applies to the alternative formulation because the regularization paths are the same.
For Theorem~\ref{theo:theo1}, it does not readily apply. But since the relationship between $\lambda_n$ and $\mu_n$ at
optimum is   $\lambda_n =\mu_n \textstyle \left(\sum_{i=1}^n d_i \|w_i\| \right)$  and that $\sum_{i=1}^n d_i \|\hat{w}_i\|$ converges to
a constant whenever $\hat{w}$ is consistent, it does apply as well with minor modifications (in particular, to deal with the case where $\J$ is empty, which requires
$\mu_n = \infty$).

\section{Covariance Operators and Multiple Kernel Learning}
\label{sec:mklsec}

We now extend the previous consistency results to the case of non-parametric estimation, where each group is a potentially infinite dimensional
space of functions. Namely, the non parametric group Lasso aims at estimating a sparse linear combination
of functions of separate random variables, and can then be seen as a variable selection method
in a generalized additive model~\citep{hastie_GAM}. Moreover, as shown in \mysec{mkl}, the non-parametric group Lasso
 may also be seen as equivalent to learning a convex combination of kernels, a framework referred to 
as multiple kernel learning (MKL).
 In this context it is customary to have a single input space with several kernels (and hence Hilbert spaces) defined on the same input space~\citep{gert,skm}. Our framework accomodates this case as well, but our assumption \hypref{compact} regarding the invertibility of the joint correlation operator states that the kernels cannot span Hilbert spaces which intersect.

In this nonparametric context, covariance operators constitute appropriate tools for the statistical analysis and are
becoming standard in the  theoretical analysis of kernel methods~\citep{fukumizu,gretton,kenji,capo}. The following section reviews
important concepts. For more details, see~\citet{baker} and~\citet{fukumizu}.

\subsection{Review of Covariance Operator Theory}
\label{sec:covop}
In this section, we first consider a single set $\mathcal{X}$ and a positive definite kernel $k:\mathcal{X}\times \mathcal{X} \to \rb$, associated
with the reproducing kernel Hilbert space (RKHS) $\mathcal{F}$ of 
functions from $\mathcal{X}$ to $\rb$ (see, e.g., \citet{scholkopf-smola-book} or
\citet{rkhs} for an introduction to RKHS theory). The Hilbert space and its dot product $\langle \cdot, \cdot \rangle_\F$ 
are such that for all $x \in \mathcal{X}$, then $k(\cdot,x) \in \F$ and for all $f \in \F$, 
$\langle k(\cdot,x),f \rangle_\F = f(x)$, which leads to the \emph{ reproducing property}
$\langle k(\cdot,x), k(\cdot,y) \rangle_\F = k(x,y)$ for any $(x,y) \in \mathcal{X} \times \mathcal{X}$.

\paragraph{Covariance operator and norms}

Given a random variable $X$ on $\mathcal{X}$ with bounded second order moment, i.e., such that $\E k(X,X) < \infty$, 
 we can define the   covariance operator as the bounded linear operator
$\S_{XX}$ from  $\F$ to $\F$ such that for all $(f,g) \in \F \times \F$,
$$\langle f , \S_{XX} g \rangle_\F = \cov( f(X),g(X)) = \E(f(X)g(X)) - (\E f(X))(\E g(X)) .$$
The operator $\S_{XX}$ is  \emph{auto-adjoint}, \emph{non-negative} and \emph{Hilbert-Schmidt}, i.e., for any
orthonormal basis $(e_p)_{p\geqslant 1}$ 
of $\F$, then $\sum_{p=1}^\infty  \| \S_{XX} e_p \|_\F^2$ is finite; in this case, the value does not depend
on the chosen basis and is referred to as the square of the Hilbert-Schmidt norm.
 The norm that we use by default in this paper is the operator norm $
\|\Sigma_{XX}\|_\F = \sup_{ f \in \F, \ \|f\|_\F=1} \| \S_{XX} f \|_\F$, which is dominated by the Hilbert-Schmidt norm.
Note that in the finite dimensional case where $\mathcal{X}=\rb^p$, $p>0$ and the kernel is linear, the  covariance operator is exactly the covariance matrix, and the Hilbert-Schmidt norm
is the Frobenius norm, while the operator norm is the maximum singular value (also referred to as the spectral norm).

The null space of the covariance operator is the space of functions $f\in \F$ such that $\var f(X)=0$, i.e., such that $f$ is constant on the support of $X$.

\paragraph{Empirical estimators}
Given data $x_i \in \mathcal{X}, i=1,\dots,n$ sampled i.i.d. from $P_X$, then the empirical estimate $\hS_{XX}$ of $\S_{XX}$ is
defined such that $\langle f, \hS_{XX} g \rangle_\F$ is the empirical   covariance between $f(X)$ and $g(X)$, which leads to:
$$\hS_{XX} = \frac{1}{n} \sum_{i=1}^n k(\cdot,x_i) \otimes k(\cdot,x_i)
-\frac{1}{n} \sum_{i=1}^n   k(\cdot,x_i) \otimes \frac{1}{n} \sum_{i=1}^n   k(\cdot,x_i)
,
$$
where $u \otimes v$ is the operator defined by
$\langle f , ( u \otimes v ) g \rangle_\F  = \langle
f ,u \rangle_\F  \langle
g ,v \rangle_\F$.
If we further assume that the fourth order moment is finite, i.e., $\E k(X,X)^2 < \infty$, then 
the estimate is uniformly consistent i.e., $\| \hS_{XX} -\S_{XX}\|_\F =
O_p(n^{-1/2})$ (see~\citet{kenji} and Appendix~\ref{app:cov}), which generalizes the usual result
of finite dimension.\footnote{A random variable $Z_n$ is said to be of order $O_p(a_n)$ if for
any $\eta>0$, there exists $M>0$ such that $\sup_n \P( |Z_n| > M a_n ) < \eta$. See~\citet{VanDerVaart}
for further definitions and properties of asymptotics in probability.}

\paragraph{Cross-covariance and joint covariance operators}
Covariance operator theory can be extended to cases with more than one random variables~\citep{baker}. In our situation, we have
$m$ input spaces $\mathcal{X}_1,\dots,\mathcal{X}_m$ and $m$ random variables $X=(X_1,\dots,X_m)$ and $m$
RKHS $\F_1,\dots,\F_m$ associated with $m$ kernels $k_1,\dots,k_m$.

If we assume that  $\E  k_j(X_j,X_j) < \infty$, for all $j=1,\dots,m$, then we can naturally define the cross-covariance
operators $\S_{X_i X_j}$ from $\F_j$ to $\F_i$ such that $\forall (f_i,f_j) \in \F_i \times \F_j$,
$$\langle f_i , \S_{X_i X_j} f_j \rangle_\Fi = \cov( 
f_i(X_i), f_j(X_j)) = 
\E(f_i(X_i) f_j(X_j)) -  (\E f_i(X_i) ) ( \E f_j(X_j)).$$
These are also Hilbert-Schmidt operators, and if we further 
assume that  $\E k_j(X_j,X_j)^2 < \infty$, for all $j=1,\dots,m$, then the natural empirical estimators
converges to the population quantities in Hilbert-Schmidt and operator norms at rate $O_p(n^{-1/2})$. We can now define
a joint block covariance operator on $\F = \F_1 \times \cdots \times \F_m$ following the block structure
of covariance matrices in \mysec{grouplasso}. As in the finite dimensional case, it leads to a joint covariance
operator $\S_{XX}$ and we can refer to sub-blocks as $\S_{X_I X_J}$ for the blocks indexed by $I$ and $J$.

Moreover, we can define the bounded (i.e., with finite operator norm) correlation operators through
$\S_{X_i X_j} = \S_{X_i X_i}^{1/2} C_{X_i X_j}  \S_{X_j X_j}^{1/2}$~\citep{baker}. Throughout 
this paper we will make the assumption that those operators $C_{X_i X_j}$
 are \emph{compact} for $i\neq j$: compact operators can be characterized
as limits of finite rank operators or as operators that can be diagonalized on a countable basis with spectrum composed
of a sequence tending to zero~\citep[see, e.g.,][]{brezis80analyse}. This implies
that the joint operator $C_{XX}$, naturally
defined on $\F = \F_1 \times \cdots \times \F_m$, is of the form ``identity plus compact''. It thus has a minimum and a
maximum eigenvalue which are both between $0$ and $1$~\citep{brezis80analyse}. If those eigenvalues are strictly greater than zero, then the
operator is invertible, as are all the square sub-blocks. Moreover, the joint correlation operator is lower-bounded by a strictly positive constant times the identity operator.

\paragraph{Translation invariant kernels}
A particularly interesting ensemble of RKHS in the context of 
nonparametric estimation is the set of translation invariant kernels defined 
over $\mathcal{X} = \rb^p$, where $p\geqslant 1$, of the form $k(x,x') = q(x'-x)$ where $q$ is a function on $\rb^p$ with pointwise nonnegative integrable Fourier transform (which implies that $q$ is continuous).
In this case, the associated RKHS is $  \mathcal{F} = \{ q_{1/2} \ast g , \ g \in L^2(\rb^p) \}$, 
where $q_{1/2}$ denotes the inverse Fourier transform of the square root of the Fourier transform of $q$
and $\ast$ denotes the convolution operation, and
$L^2(\rb^p)$ denotes the space of square integrable functions. The norm is thenequal to 
$$\| f\|_\F^2 = \int \frac{ |F(\omega) |^2}{ Q(\omega) } d \omega,$$
where $F$ and $Q$ are the Fourier transforms of $f$ and $q$~\citep{wahba,scholkopf-smola-book}. 
Functions in the RKHS are  functions with appropriately integrable derivatives.
In this paper, when using infinite dimensional kernels, we use the Gaussian kernel
$k(x,x') = q(x-x') = \exp( -   b \| x - x' \|^2 )$.

\paragraph{One-dimensional Hilbert spaces}
In this paper, we also consider real random variables $Y$ and $\varepsilon$ embedded in the natural Euclidean structure of real numbers (i.e.,
we consider the linear kernel on $\rb$). In this setting the covariance operator $\S_{X_jY}$ from $\rb$ to $\F_j$ can be canonically
identified as an element of $\F_j$. Throughout this paper, we always use this identification.

\subsection{Problem Formulation}

We assume in this section and in the remaining of the paper
 that for each $j=1,\dots,m$, $X_j \in \mathcal{X}_j$ where $\mathcal{X}_j$ is any set on which we have
 a reproducible kernel Hilbert spaces
$\F_j$, associated with the positive kernel $k_j:\mathcal{X}_j \times \mathcal{X}_j \to \rb$.
We now make the following assumptions, that extends the assumptions \hypref{var},
\hypref{inv} and \hypref{model}. For each of them, we detail the main implications as well as common natural sufficient
conditions. The first two conditions \hypref{rkhs} and \hypref{compact} depend solely on  the input variables, while the two other ones, \hypref{model-cov} and \hypref{range} consider the relationship between $X$ and $Y$.

\begin{hyp}
\label{hyp:rkhs}
For each $j=1\,\dots,m$, 
$\F_j$ is a separable reproducing kernel Hilbert space associated with kernel $k_j$, and the random variables $k_j(\cdot,X_j)$ are not constant and have finite
fourth-order moments, i.e., $\E k_j(X_j,X_j)^2 < \infty$.
\end{hyp}

This is a non restrictive assumption in many situations; for example, when (a) $\mathcal{X}_j = \rb^{p_j}$ and the kernel function (such as
the Gaussian kernel) is bounded, or when (b) $\mathcal{X}_j$ is a compact subset of $\rb^{p_j}$ and the kernel is any continuous function
such as linear or polynomial.
 This implies notably, as shown in \mysec{covop}, that we can define covariance, cross-covariance and correlation
 operators that are all Hilbert-Schmidt~\citep{baker,kenji} and can all be estimated at rate $O_p(n^{-1/2})$ in operator norm.

\begin{hyp}
\label{hyp:compact}

All cross-correlation operators are compact and  the joint correlation operator $C_{XX}$ is invertible.
\end{hyp}
This is also a condition uniquely on the input spaces and not on $Y$. Following~\citet{kenji}, a simple sufficient condition is that
we have measurable spaces and distributions with joint density $p_X$ (and
marginal distributions $p_{X_i}(x_i)$ and $p_{X_i X_j}(x_i,x_j)$) and that the \emph{mean square contingency} between all pairs of variables is finite, i.e.
$$
 \E\left\{ \frac{ p_{X_iX_j}(x_i,x_j) }{p_{X_i }(x_i )p_{X_j}(x_j) } - 1 \right\} < \infty.
$$
The contingency is a measure of statistical dependency~\citep{renyi}, and thus this sufficient condition
simply states that two variables $X_i$ and $X_j$ cannot be too dependent. In the context of multiple kernel learning
for heterogeneous data fusion, this corresponds to having sources which are heterogeneous enough.
On top of compacity we impose the invertibility of the joint correlation operator; we use this assumption to make sure that the
functions $\f_1,\dots,\f_m$  are unique.
This ensures the non existence of any set of functions $f_1,\dots,f_m$ in the closures of
$\F_1,\dots,\F_m$,
  such that $\var f_j(X_j) >0$ and a linear combination
is constant on the support of the random variables. In the context of generalized additive models, this assumption is referred to as
the empty \emph{concurvity space} assumption~\citep{hastie_GAM}.

\begin{hyp}
\label{hyp:model-cov}
There exists functions $\f=(\f_1,\dots,\f_m) \in \F = \F_1 \times \dots \times \F_m$,
$\b \in \rb$,
 and a function $\h$ of $X=(X_1,\dots,X_m)$ such that
$\E(Y|X) = \sum_{j=1}^m \f_j(X_j) + \b +  \h(X)$ with  $\E h(X)^2 < \infty$, $\E h(X) = 0$ and $\E \h(X) f_j(X_j) =0$ for all $j=1,\dots,m$ and $f_j
\in \F_j$.
We assume
that $\E( (Y - \f(X) - \b)^2  | X)$ is almost surely greater than $\sm>0$ and smaller than $\sM < \infty$.
We let denote $\J = \{ j, \fj \neq 0\}$
the sparsity pattern of $\f$.
\end{hyp}
This assumption on the conditional expectation of $Y$ given $X$ is not the most general and follows common assumptions
in approximation theory~(see, e.g.,~\citet{capo,smale} and references therein). It allows misspecification, but
it essentially requires that the conditional expectation of $Y$ given sums of measurable functions of 
$X_j$ is attained at functions in the RKHS, and not merely measurable functions. Dealing with more general assumptions in the line of~\citet{spam} requires
to consider consistency for norms weaker than the RKHS norms~\citep{capo,steinwart}, and is left
for future research. Note also, that to simplify proofs, we assume a finite upper-bound $\sM$ on the residual variance.

\begin{hyp}
\label{hyp:range}

 For all $j \in \{1,\dots,m\}$, there exists  $\gj \in \mathcal{F}_j$ such that $\fj = \S_{X_j X_j}^{1/2} \gj$, i.e., each $\fj$ is in the range
of $\S_{X_j X_j}^{1/2}$. 
\end{hyp}
This technical condition, already used by~\citet{capo}, which concerns all RKHS independently, ensures that we obtain consistency for the norm of the RKHS (and not another weaker norm) for the least-squares
estimates. Note also that it implies that $\var f_j(X_j) > 0 $, i.e., $f_j$ is not constant on the support of $X_j$.

This assumption might be checked (at least) in two ways; first, if $(e_p)_{p\geqslant 1}$
is a sequence of eigenfunctions of $\Sigma_{X X}$, associated with strictly positive eigenvalues $\lambda_p > 0$, then $f$ is in the range of $\S_{XX}$ if and only if $f$ is constant outside the support of the random variable $X$ and $\sum_{p \geqslant 1} \frac{1}{\lambda_p}
\langle f,e_p \rangle^2  $ is finite (i.e, the decay of the sequence $\langle f,e_p \rangle^2 $ is strictly faster than $\lambda_p$).

We also provide another sufficient condition that sheds additional light on this technical condition which is always true for finite dimensional Hilbert spaces. For the common
situation where $\mathcal{X}_j = \rb^{p_j}$, $P_{X_j}$ (the marginal distribution of $X_j$) has a density $p_{X_j}(x_j)$
with respect to the Lebesgue measure and the kernel is of the form $k_j(x_j,x_j') = q_j(x_j-x_j')$,
we have the following proposition (proved in \myapp{range}):
\begin{proposition}
\label{prop:range}
Assume  $\mathcal{X}=\rb^p$  and $X$ is a random variable on $\mathcal{X}$ with distribution
$P_{X}$ that has a \emph{strictly positive} density $p_{X}(x)$ with respect to the Lebesgue measure. Assume $k(x,x')=
q(x-x')$ for a function $q \in L^2(\rb^p)$ has an integrable pointwise positive Fourier transform, 
with associated RKHS $\mathcal{F}$.  If $f$ can be written as
$f = q \ast  g$ (convolution of $q$ and $g$) with  $\int_{\rb^p} g(x) dx = 0 $ and $\int_{\rb^p} \frac{g(x)^2}{p_{X}(x)} dx < \infty$, then $f \in  \F$ is in the range of
the square root $\S_{XX}^{1/2}$ of the covariance operator.
\end{proposition}
The previous proposition gives natural conditions regarding $f$ and $p_X$. Indeed,  the condition
$\int \frac{g(x)^2}{p_{X}(x)} dx  < \infty$ corresponds to a natural support condition, i.e., $f$ should be zero where $X$
has no mass, otherwise, we will not be able to estimate $f$; note the similarity
with the usual condition regarding the variance of importance sampling estimation~\citep{bremaud}.  Moreover, $f$ should be even smoother than a regular function in the RKHS (convolution by $q$ instead of the square root of $q$). Finally, we provide in \myapp{covgauss} detailed covariance structures for Gaussian kernels with Gaussian variables.

\paragraph{Notations}
Throughout this section, we refer to
functions $f = (f_1,\dots,f_m) \in \mathcal{F} = \mathcal{F}_1 \times \dots \times \mathcal{F}_m$ and the
joint covariance operator $\S_{XX}$. In the following, we always use the norms of the RKHS. When considering operators,
we use the operator norm. We also refer to a subset of $f$ indexed by $J$ through $f_J$. Note that the Hilbert norm $\| f_J\|_{\F_J}$ is equal to
$\| f_J\|_{\F_J} = ( \sum_{j \in J} \| f_j \|_\Fj )^{1/2}$. Finally, given a nonnegative auto-adjoint operator $S$, we let denote $S^{1/2}$ its nonnegative autoadjoint square root~\citep{baker}.

\subsection{Nonparametric Group Lasso}

Given i.i.d data $(x_{ij},y_i)$, $i=1,\dots,n$, $j=1,\dots,m$, where  each $x_{ij} \in \mathcal{X}_j$, our goal is to estimate 
consistently the functions $\fj$ and which of them are zero. We let denote  $\Y \in \rb^n$ the vector of responses.
 We consider the following optimization problem:
$$
\min_{f \in \F , \ b \in \rb } \ 
\frac{1}{2n} \sum_{i=1}^n \left( y_i - \sum_{j=1}^m f_j(x_{ij}   )  -b  \right)^2
+ \frac{\mu_n}{2} \left(  \sum_{j=1}^m d_j \|f_j\|_\Fj \right)^2.
$$
By minimizing with respect to $b$ in closed form,  we  obtain a similar formulation to \eq{problem-alt}, where empirical covariance matrices
are replaced by empirical covariance operators:
\BEQ
\label{eq:problem-alt-cov}
\min_{f \in \F} \ 
\frac{1}{2} \hS_{YY} -  \langle f, \hS_{XY} \rangle_\F  + \frac{1}{2} \langle f,
 \hS_{XX} f \rangle_\F +  \frac{\mu_n}{2} \left( \sum_{j=1}^m d_j \|f_j\|_\Fj \right)^2.
\EEQ
We let denote  $\hat{f}$ any minimizer of \eq{problem-alt-cov}, and we refer to it as the non parametric
group Lasso estimate, or also the multiple kernel learning estimate. By
Proposition~\ref{prop:opt-alt-dual}, the previous problem has indeed minimizers, and by 
Proposition~\ref{prop:opt-alt-dual-unique} this global minimum is unique with probability tending to one.

Note that formally, the finite and infinite dimensional formulations in \eq{problem-alt} 
and \eq{problem-alt-cov} are the same, and this is the main reason why covariance operators
are very practical tools for the analysis. Furthermore, we have the corresponding proposition regarding optimality conditions
(see proof in  \myapp{opt-alt-cov}):
\begin{proposition}
\label{prop:opt-alt-cov}
A function $f \in \mathcal{F}$ with sparsity pattern $J = J(f) =  \{ j, \ f_j \neq 0\}$ is optimal for problem (\ref{eq:problem-alt-cov}) if
and only if
\BEA
\label{eq:opt1-alt-cov}
\forall j \in J^c ,& &   \left\| \hS_{X_j X} f -  \hS_{X_j Y}  \right\|_\Fj  \leqslant  \mu_n d_j
 \textstyle \left(\sum_{i=1}^n d_i \|f_i\|_\Fi \right) \displaystyle ,  \\
\label{eq:opt2-alt-cov}
\forall j \in J ,& &   
  \hS_{X_{j} X }f  -  \hS_{X_{j} Y} =  - \mu_n  \textstyle \left(\sum_{i=1}^n d_i \|f_i\|_\Fi \right) \displaystyle  \frac{d_j f_j}{ \|f_j\|_\Fj}.
\EEA
\end{proposition}
A consequence (and in fact the first part of the proof) is that an optimal function $f$ must be in the range of
$\hS_{XY}$ and $\hS_{XX}$, i.e., an optimal $f$ is supported by the data; that is, each $f_j$ is a linear combination of functions
$k_j(\cdot,x_{ij})$, $i=1,\dots,n$. This is a rather circumvoluted way of presenting the representer theorem~\citep{wahba}, but this is
the easiest for the theoretical analysis of consistency. However, to actually compute the estimate $\hat{f}$ from data, we need the usual formulation
with dual parameters (see \mysec{mkl}).

Moreover, one important
conclusion  is that all our optimization problems in spaces of functions can be in fact
transcribed into finite-dimensional problems. In particular, all notions from multivariate differentiable
calculus may be used without particular care regarding the infinite dimension.

\subsection{Consistency Results}

We consider the following strict and weak conditions, which  correspond to condition (\ref{eq:condition})
and (\ref{eq:condition-weak}) in the finite
dimensional case:
\BEQ
\label{eq:condition-cov}
 \max_{ i \in \J^c } \frac{1}{d_i} \left\| \S_{X_i X_i }^{1/2} C_{X_i X_\J  }  C_{X_\J  X_\J }^{-1}
\Diag ( d_j/ \| \fj \|_\Fj  ) \gJ  \right\|_\Fi < 1,
\EEQ
\BEQ
\label{eq:condition-cov-weak}
 \max_{ i \in \J^c } \frac{1}{d_i} \left\| \S_{X_i X_i }^{1/2} C_{X_i X_\J  }  C_{X_\J  X_\J }^{-1}
\Diag ( d_j/ \| \fj \|_\Fj  ) \gJ  \right\|_\Fi \leqslant 1,
\EEQ
where $\Diag  ( d_j/ \| \fj \|_\Fj)$ denotes the block-diagonal operator   with operators
$ \frac{d_j}{\|\fj\|_\Fj} I_{\mathcal{F}_j}$ on the diagonal.  Note that this is well-defined because
$C_{XX}$ is invertible and that it reduces to \eq{condition} and \eq{condition-weak} when the input spaces $\mathcal{X}_j$, $j=1,\dots,m$ are of the form $\rb^{p_j}$ and the kernels are linear. The main reason is rewriting the conditions in terms of correlation operators rather than covariance operators is that correlation operators are invertible by assumption, while covariance operators are not as soon as the Hilbert spaces have infinite dimensions.
The following theorems give necessary and sufficient conditions for the path consistency of
the nonparametric group Lasso (see proofs in \myapp{theo1-cov} and \myapp{theo2-cov}):

\begin{theorem}
\label{theo:theo1-cov}
Assume \hypreff{rkhs}{range} and
that $\J$ is not empty.
If condition (\ref{eq:condition-cov}) is satisfied, then for any sequence $\mu_n$ such that
 $\mu_n  \to 0$ and $\mu_n n^{1/2} \to + \infty$,   any sequence of nonparametric group Lasso estimates $\hat{f}$ converges in probability
 to $\f$ and the sparsity pattern $J(\hat{f}) = \{ j, \hat{f}_j \neq 0 \}$ converges in probability to $\J$.
\end{theorem}

\begin{theorem}
\label{theo:theo2-cov}
Assume \hypreff{rkhs}{range}  and
that $\J$ is not empty.
If there exists a (possibly data-dependent) sequence $\mu_n$ such $\hat{f}$ converges
to $\f$ and $\hat{J}$ converges to $\J$ in probability, then condition (\ref{eq:condition-cov-weak})  is satisfied.
\end{theorem}

Essentially, the results in finite dimension also hold when groups have infinite dimensions. We leave the extensions
of the refined results in \mysec{refined} to future work.
Condition (\ref{eq:condition-cov}) might be hard to check in practice since it involves inversion
of correlation operators; see \mysec{estimation} for an estimate
from data.

\subsection{Multiple Kernel Learning Formulation}

\label{sec:mkl}

Proposition~\ref{prop:opt-alt-cov} does not readily lead to an algorithm for computing the estimate $\hat{f}$. In this section,
following~\citet{skm}, we link the group Lasso to the multiple kernel learning framework~\citep{gert}.
Problem (\ref{eq:problem-alt-cov}) is an optimization problem on a 
potentially infinite dimensional space of functions. However, the following proposition shows that it
reduces to a finite dimensional problem that we now precise~(see proof in \myapp{opt-alt-dual}):
\begin{proposition}
\label{prop:opt-alt-dual}
The dual of problem (\ref{eq:problem-alt-cov}) is
\BEQ
\label{eq:dual}
\max_{ \alpha \in \rb^n, \ \alpha^\top 1_n=0 } 
\left\{
-\frac{1}{2n} \| \Y - n\mu_n \alpha \|^2  
- \frac{1}{2 \mu_n} \max_{ i=1,\dots,m } \frac{ \alpha^\top {K}_i \alpha}{d_i^2} \right\},
\EEQ
where $(K_i)_{ab} = k_i(x_a,x_b)$ are the
kernel matrices in $\rb^{n \times n}$, for $i=1,\dots,m$. Moreover, the dual variable $\alpha \in \rb^n$ is optimal if and only if $\alpha^\top 1_n = 0 $ and there exists $\eta \in \rb_+^m$ such that
$\sum_{j=1}^m \eta_j d_j^2 = 1$ and
\BEA
\label{eq:opt2-altdual}
\left( \sum_{j=1}^m \eta_j {K}_j  + n \mu_n  \idm_n \right) \alpha =  \Y, \\
\label{eq:opt3-altdual}
\forall j \in \{1,\dots,m\}, \ \frac{\alpha^\top {K}_j \alpha }{d_j^2} <   \max_{ i=1,\dots,m } \frac{\alpha^\top {K}_i \alpha }{d_i^2}
\Rightarrow \eta_j = 0.
\EEA
The optimal function may then be written as $f_j  = \eta_j \sum_{ i=1}^n \alpha_i k_j(\cdot,x_{ij})$.
\end{proposition}
Since the problem in \eq{dual} is strictly convex, there is a unique dual solution $\alpha$.
Note that \eq{opt2-altdual} corresponds to the optimality conditions for the least-square problem:
$$\min_{ f \in \F} 
\frac{1}{2} \hS_{YY} -  \langle f, \hS_{XY} \rangle_\F  + \frac{1}{2} \langle f,
 \hS_{XX} f \rangle_\F +  \frac{1}{2} \mu_n  \sum_{j, \ \eta_j>0}  \frac{\|f_j\|_\Fj^2}{\eta_i},
$$
whose dual problem is:
$$
\max_{ \alpha \in \rb^n, \ \alpha^\top 1_n = 0  } 
\left\{
-\frac{1}{2n} \| \Y - n\mu_n \alpha \|^2  
- \frac{1}{2 \mu_n} \alpha^\top \left(\sum_{j=1}^m \eta_i {K}_i \right) \alpha  \right\},
$$
and  unique solution is $\alpha = ( \sum_{j=1}^m \eta_j K_j + n \mu_n \idm_n)^{-1} \Y$.
That is, the solution of the MKL problem leads to dual parameters $\alpha$ and set of weights $\eta \geqslant 0$
such that $\alpha$ is the solution to the least-square problem with kernel $K = \sum_{j=1}^m \eta_j K_j$. 
\citet{skm} has shown in a very similar context (hinge loss instead of the square loss)
that the optimal $\eta$ in Proposition~\ref{prop:opt-alt-dual}
can be obtained as the minimizer of the optimal value of the regularized least-square problem with kernel matrix $\sum_{j=1}^m \eta_j K_j$, i.e.:
$$ J(\eta) = 
\max_{ \alpha \in \rb^n, \ \alpha^\top 1_n = 0  }  \left\{
-\frac{1}{2n} \| \Y - n\mu_n \alpha \|^2  
- \frac{1}{2 \mu_n}  \alpha^\top \left(\sum_{j=1}^m \eta_j K_j \right)
\alpha \right\},
$$
with respect to $\eta \geqslant 0 $ such that  $\sum_{j=1}^m \eta_j d_j^2 = 1$. This formulation allows to
derive probably approximately correct error bounds~\citep{gert,bousquet}. Besides, this formulation
allows $\eta$ to be negative, as long as the matrix  $\sum_{j=1}^m \eta_j K_j$ is positive semi-definite. However, theoretical
advantages of such a possibility still remain unclear.

Finally, we state a corollary of Proposition~\ref{prop:opt-alt-dual} that shows that under our assumptions regarding the correlation operator, we have a unique solution to the non parametric groups Lasso problem with probability tending to one~(see proof in \myapp{opt-alt-dual-unique}):
\begin{proposition}
\label{prop:opt-alt-dual-unique}
Assume \hypreff{rkhs}{compact}. 
The  problem (\ref{eq:problem-alt-cov}) has a unique solution with probability tending to one.
\end{proposition}

\subsection{Estimation of Correlation Condition (\ref{eq:condition-cov}) }
\label{sec:estimation}
Condition (\ref{eq:condition}) is simple to compute while the non parametric condition 
(\ref{eq:condition-cov}) might be hard to check even
if  all densities are known (we provide however in \mysec{simulations} a specific example where we can compute in closed form all covariance operators). The following proposition shows that we can consistently estimate
the quantities $\left\| \S_{X_i X_i }^{1/2} C_{X_i X_\J  }  C_{X_\J  X_\J }^{-1}
\Diag ( d_j/ \| \fj \|_\Fj  ) \gJ  \right\|_\Fi$ given an i.i.d. sample~(see proof
in \myapp{correst}):
\begin{proposition}
\label{prop:correst}
Assume \hypreff{rkhs}{range}, and
  $\kappa_n \to 0 $
 and $\kappa_n n^{1/2} \to \infty$.
Let 
$$ \alpha =  \Pi_n
 \left( \sum_{j \in \J}    \Pi_n {K}_j  \Pi_n  + n \kappa_n \idm_n \right)^{-1}   \Pi_n \Y $$
 and $\hat{\eta}_j =  \frac{1}{d_j} ( \alpha^\top K_j \alpha)^{1/2}$.   
Then, for all $i \in \J^c$, the norm $\left\| \S_{X_i X_i }^{1/2} C_{X_i X_\J  }  C_{X_\J  X_\J }^{-1}
\Diag ( d_j/ \| \fj \|  ) \gJ  \right\|_\Fi$ is consistently estimated by:
\BEQ
\label{eq:correst}
\left\|
 (\Pi_n{K}_i \Pi_n)^{1/2}
\left(
\sum_{j\in \J}  \Pi_n {K}_j   \Pi_n + n \kappa_n \idm_n
\right)^{-1}
\left(
\sum_{j\in \J} \frac{1}{ \hat{\eta}_j}   \Pi_n {K}_j  \Pi_n
\right) \alpha
\right\|.
\EEQ
\end{proposition}

\section{Adaptive Group Lasso and Multiple Kernel Learning}
\label{sec:adaptive}
In previous sections, we have shown that specific necessary and sufficient conditions are needed for
path consistency of the group Lasso and multiple kernel learning. The following procedures, adapted from the
adaptive Lasso of~\citet{zou}, lead to  two-step procedures that always achieve both consistency, with no condition such
as \eq{condition} or \eq{condition-cov}. As before, results are a bit different when groups have finite sizes and groups may have infinite sizes.

\subsection{Adaptive Group Lasso}
The following theorem extends the similar theorem of~\citet{zou}, and shows that we can get both $O_p(n^{-1/2})$ consistency and correct pattern estimation:
\begin{theorem}
\label{theo:adapGL}
Assume \hypreff{var}{model} and $\gamma>0$. Let $\hat{w}^{LS} = \hS_{XX}^{-1} \hS_{XY}$ denote the (unregularized) least-square estimate.
Let
$\hat{w}^A$ denote any minimizer
of 
$$
\frac{1}{2} \hS_{YY} -  \hS_{YX } w  + \frac{1}{2} w^\top 
 \hS_{XX} w   + \frac{\mu_n }{2}  \left( \sum_{j=1}^m  \| \hat{w}^{LS}_j  \|^{-\gamma} \| w_j \| \right)^2.
$$
If $ n^{-1/2} \gg \mu_n \gg n^{-1/2-\gamma/2}$, then 
 $\hat{w}^A$ converges in probability to $\w$, $J(\hat{w}^A)$ converges in probability to $\J$, and $n^{1/2}( \hat{w}^A_\J - \w_\J)$ tends in distribution to a normal distribution with mean zero and covariance matrix $\S_{X_\J X_\J}^{-1}$.
\end{theorem}
This theorem, proved in Appendix~\ref{app:adapGL}, shows that the adaptive group Lasso exhibit all important asymptotic properties, both in terms of errors and selected models. In the nonparametric case, we obtain a weaker result.

\subsection{Adaptive Multiple Kernel Learning}

We first begin with the consistency of the least-square estimate (see proof in \myapp{ls}):
\begin{proposition}
\label{prop:ls}
Assume \hypreff{rkhs}{range}.
The unique minimizer $\hat{f}^{LS}_{\kappa_n}$ of 
$$
\frac{1}{2} \hS_{YY} -  \langle\hS_{XY}, f\rangle_\F  + \frac{1}{2} \langle f ,
 \hS_{XX} f \rangle _\F  + \frac{\kappa_n }{2}   \sum_{j=1}^m   { \| f_j \|_\Fj^2 } ,
$$
converges in probability to $f$ if $\kappa_n \to 0 $ and $\kappa_n n^{1/2} \to 0$.
Moreover, we have $\| \hat{f}^{LS}_{\kappa_n} - f \|_\F = O_p(\kappa_n^{1/2} + \kappa_n^{-1} n^{-1/2})$.
\end{proposition}

Since the least-square estimate is consistent and we have an upper bound
 on its convergence rate, we follow~\citet{zou} and use it to defined adaptive weights
$d_j$ for which we get both sparsity and regular consistency without any conditions on the value of the correlation operators.

\begin{theorem}
\label{theo:adapt}
Assume  \hypreff{rkhs}{range} and $\gamma>1$.
Let $\hat{f}^{LS}_{n^{-1/3}}$ be the least-square estimate with regularization parameter proportional to
 $ n^{-1/3}$, as defined in
Proposition~\ref{prop:ls}. Let
$\hat{f}^A$ denote any minimizer
of 
$$
\frac{1}{2} \hS_{YY} -  \langle \hS_{XY }, f\rangle_\F  + \frac{1}{2} \langle f ,
 \hS_{XX} f \rangle_\F   + \frac{\mu_0 n^{-1/3} }{2}  \left( \sum_{j=1}^m  \| (\hat{f}^{LS}_{\kappa_n} )_j  \|_\Fj^{-\gamma} \| f_j \|_\Fj \right)^2.
$$
Then $\hat{f}^A$ converges to $\f$ and $J(\hat{f}^A)$ converges to $\J$ in probability.
\end{theorem}

Theorem~\ref{theo:adapt} allows to set up a specific vector of weights $d$. This provides a  principled way to define
data adaptive weights, that allows to solve (at least theoretically) the
 potential consistency problems of the usual MKL framework (see \mysec{simulations} for illustration on
 synthetic examples). Note that we have no result concerning the $O_p(n^{-1/2})$ consistency of our procedure (as we have for the finite dimensional case) and obtaining precise convergence rates is the subject of ongoing research.

The following proposition gives the expression for the solution of the least-square problem,
necessary for the computation of adaptive weights in Theorem~\ref{theo:adapt}.
\begin{proposition}
The solution of the least-square problem in Proposition~\ref{prop:ls} is given by
$$\forall j \in \{1,\dots,m\}, \ 
f_j^{LS} =  \sum_{i=1}^n \alpha_i k_j(\cdot,x_{ij}) \mbox{ with } \alpha =  \Pi_n
 \left( \sum_{j=1}^m   \Pi_n K_j  \Pi_n  + n \kappa_n \idm_n \right)^{-1}   \Pi_n\Y, $$
with norms $ \| \hat{F}_j^{LS} \|_\Fj =  \left( \alpha^\top K_j \alpha \right)^{1/2}$, $j=1,\dots,m$.
\end{proposition}
Other weighting schemes have been suggested, based on various heuristics. A notable one (which we use in simulations)
is the normalization of kernel matrices by their trace~\citep{gert}, which leads to
$d_j = ( \tr \hS_{X_j X_j} )^{1/2} = ( \frac{1}{n} \tr \Pi_n{K_j}\Pi_n )^{1/2}$.
\citet{bach_thibaux} have observed empirically that such normalization might
lead to suboptimal solutions and consider weights $d_j$ that grow with the empirical ranks of the kernel matrices. In 
this paper, we give theoretical arguments that
indicate that weights which do depend on the data are more appropriate and work better (see \mysec{simulations}
for examples).

\section{Simulations}
\label{sec:simulations}

In this section, we illustrate the consistency results obtained in this paper with a few simple simulations on
synthetic examples.

\subsection{Groups of Finite Sizes}

\begin{figure}
\begin{center}
\includegraphics[scale=.5]{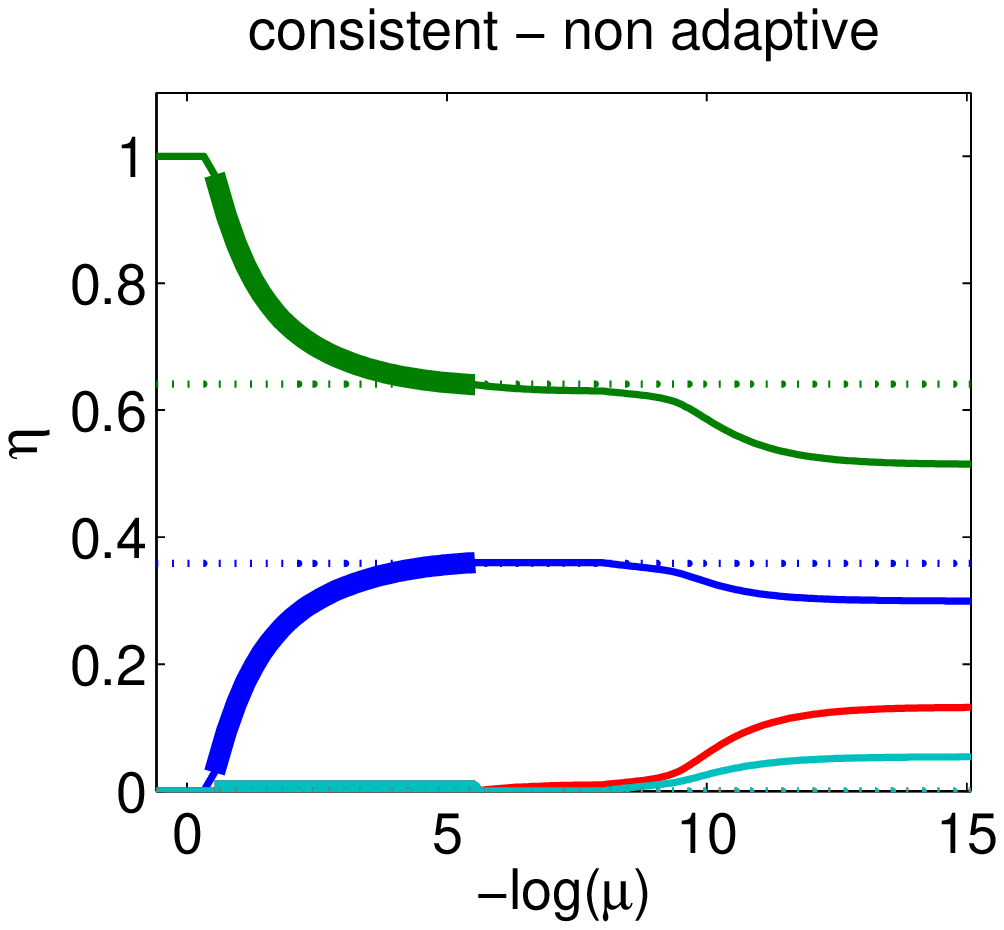} \hspace*{.5cm}
\includegraphics[scale=.5]{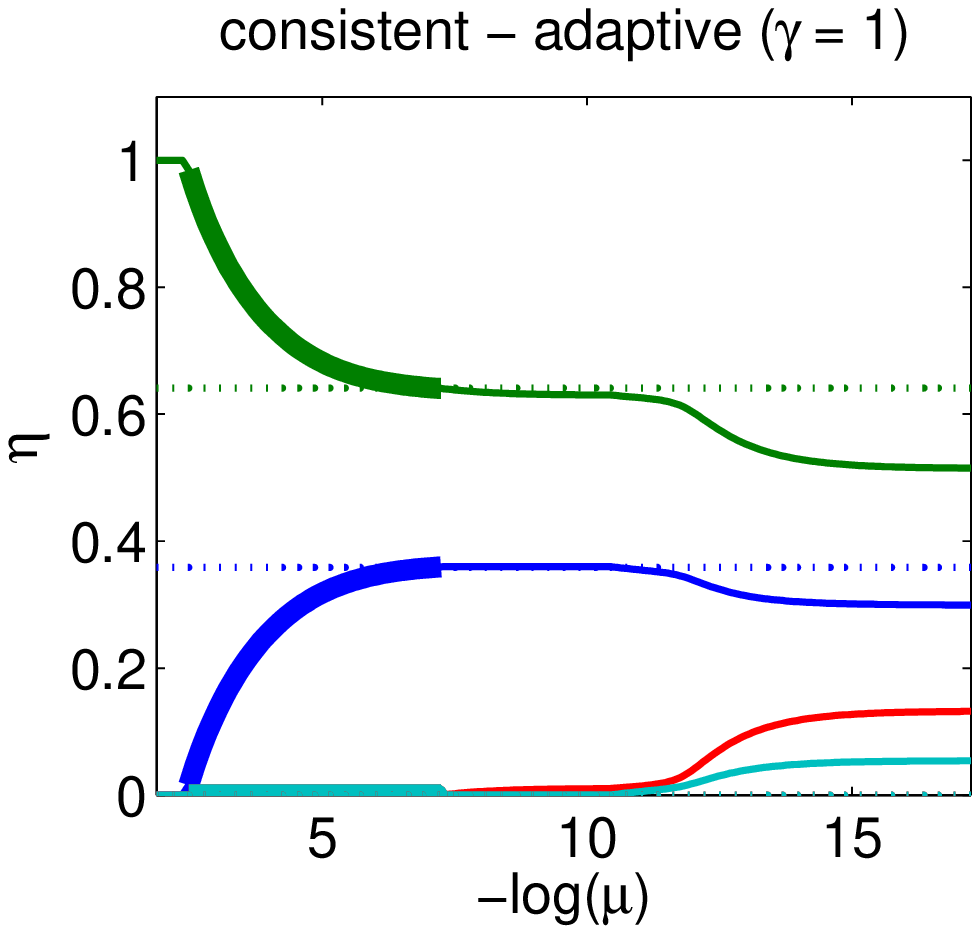}

\vspace*{.25cm}

\includegraphics[scale=.5]{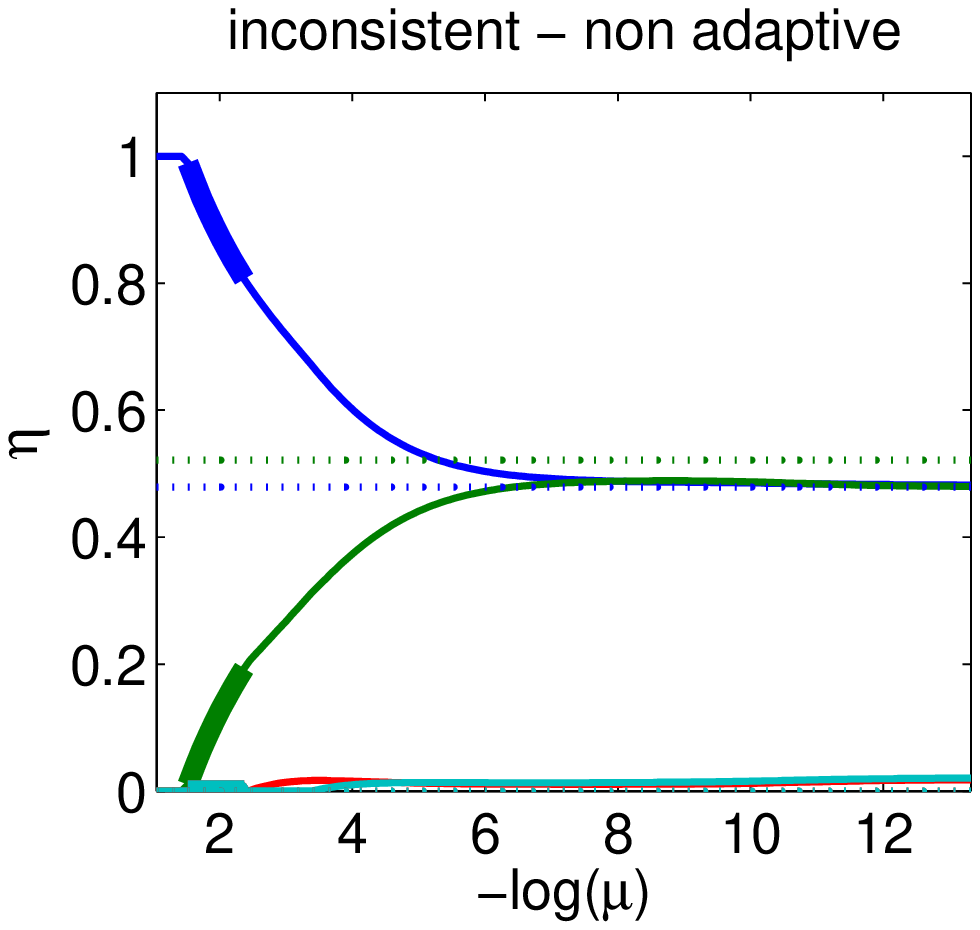}  \hspace*{.5cm}
\includegraphics[scale=.5]{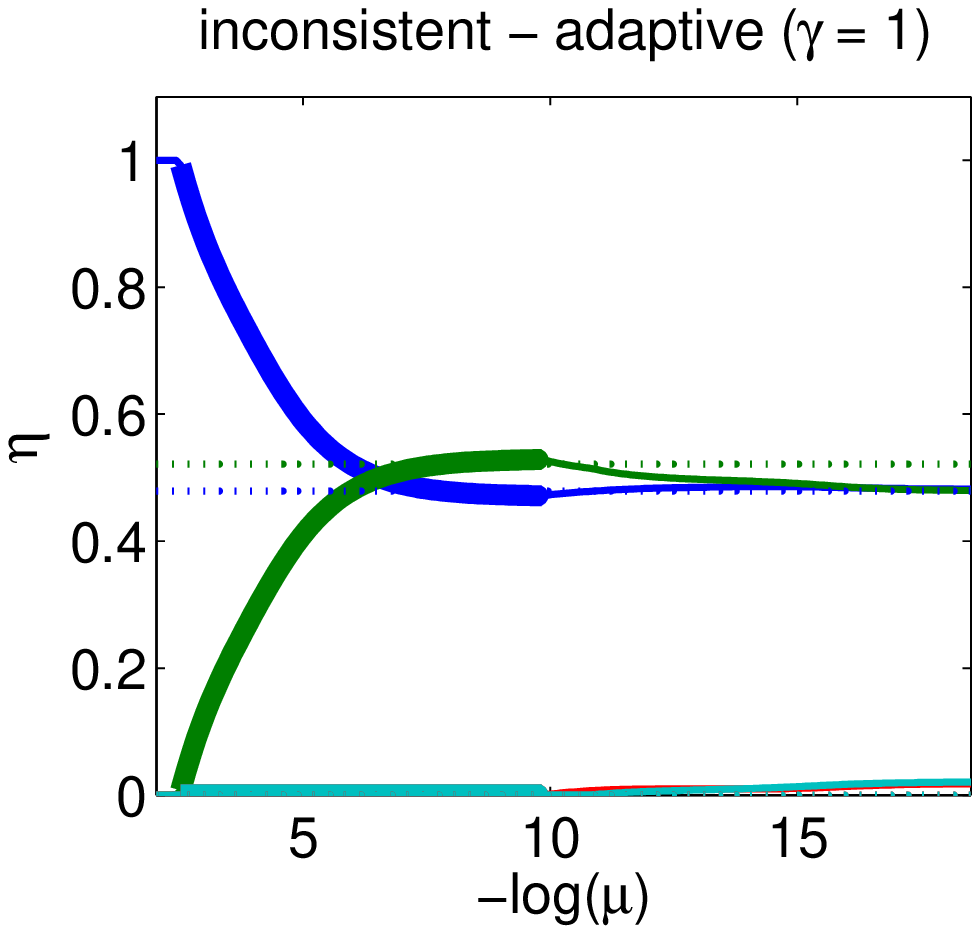}

\vspace*{.25cm}

\includegraphics[scale=.5]{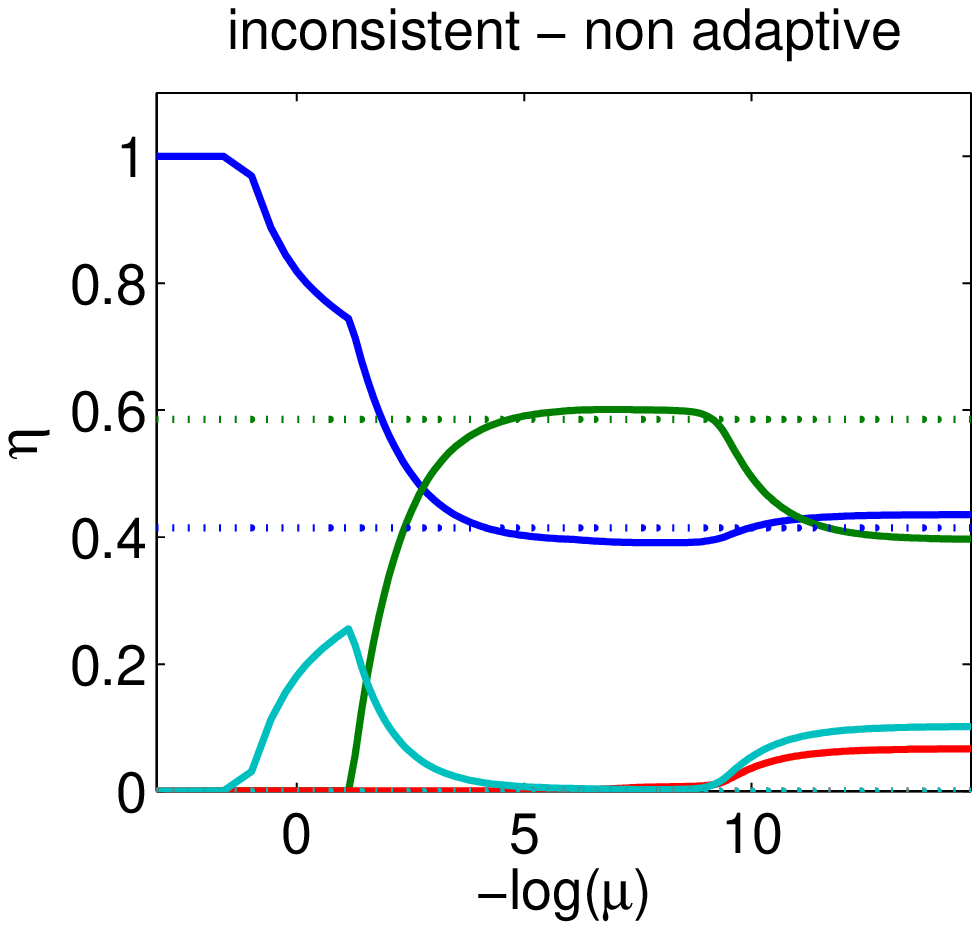}  \hspace*{.5cm}
\includegraphics[scale=.5]{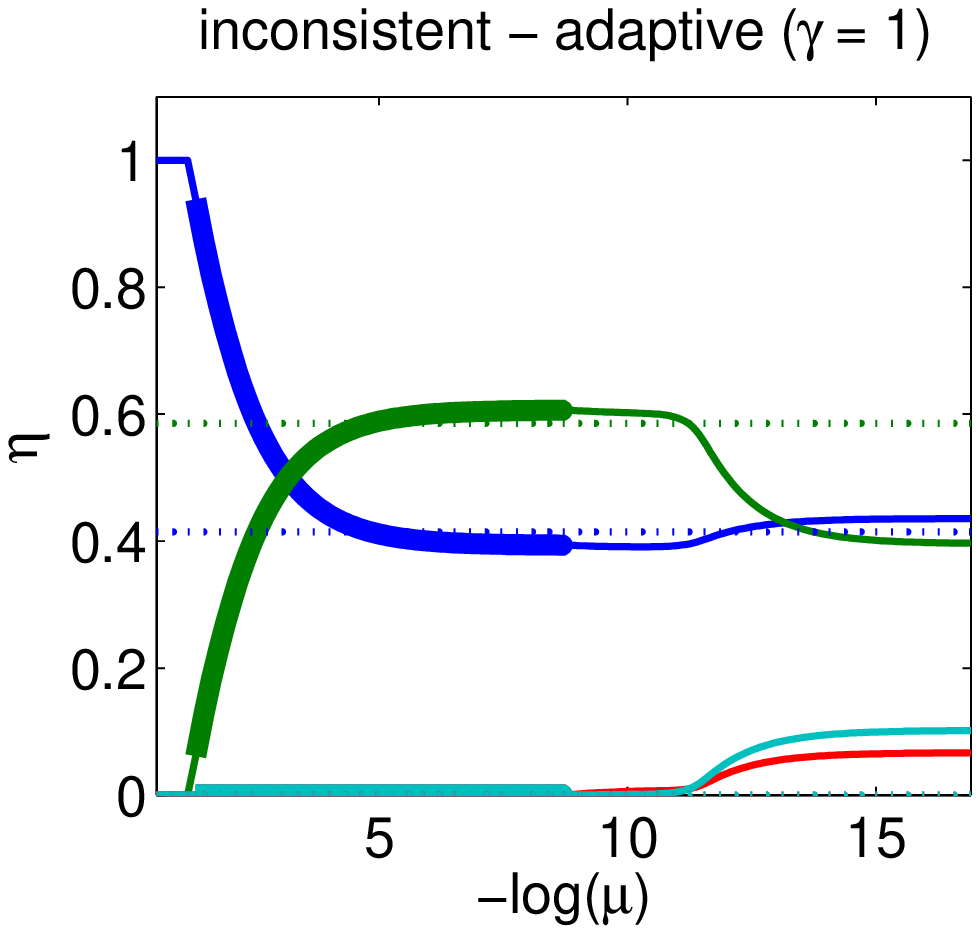}

\vspace*{-.5cm}

\end{center}
\caption{Regularization paths for the group Lasso for two weighting schemes (\emph{left}: non adaptive, \emph{right}:  adaptive) and three different population densities (\emph{top}: strict consistency condition satisfied, \emph{middle}: weak condition not satisfied, no model consistent estimates, \emph{bottom}: weak condition not satisfied, some model consistent estimates but without regular consistency). For each of the plots, 
 plain curves correspond to values of estimated $\hat{\eta}_j$, dotted curves to population values $\eta_j$, and bold curves
to model consistent estimates. }
\label{fig:grouplasso-figures}
\end{figure}

 In the finite dimensional group case, we sampled $X \in \rb^p$ from a normal distribution with zero mean vector and a covariance matrix of size $p=8$
for $m=4$ groups of size $p_j=2$, $j=1,\dots,m$, generated as follows: (a) sample an $p\times p$ matrix $G$ with independent standard normal distributions, (b) form $\S_{XX} = GG^\top$, (c) for each $j \in \{1,\dots,m\}$, rescale  $X_j \in \rb^2$ so that  $\tr \S_{X_j X_j} = 1$.
We selected ${\rm Card} ( \J) = 2$ groups at random and sampled non zero loading vectors as follows: (a) sample each loading from from independent standard normal distributions,  (b) rescale those to unit norm, (c) rescale those by a scaling which is uniform at random between $\frac{1}{3}$ and $1$. Finally, we chose a constant noise level of standard deviation $\sigma$ equal to
$0.2$ times $(\E (w^\top X)^2)^{1/2}$ and sampled $Y$ from a conditional normal distribution with constant variance. The joint distribution on $(X,Y)$ thus defined satisfies with probability one assumptions \hypreff{var}{model}.

For cases when the correlation conditions (\ref{eq:condition})
and (\ref{eq:condition-weak}) were or were not satisfied,
we consider two different weighting schemes, i.e., different ways of setting the weights $d_j$ of the block $\ell_1$-norm: unit weights (which correspond to the unit trace weighting scheme) and adaptive weights as defined
in \mysec{adaptive}. 

In \myfig{grouplasso-figures}, we plot the regularization paths corresponding to 200 i.i.d. samples,
computed by the algorithm of~\citet{bach_thibaux}. We only
plot the values of the estimated variables $\hat{\eta}_j, j =1,\dots,m$ for the alternative formulation in
\mysec{alt},
 which are proportional to $\|\hat{w}_j\|$ and normalized so that
$\sum_{j=1}^m \hat{\eta}_j   = 1$. We compare them to the population values $\eta_j$: both in terms of values, and in terms of
their sparsity pattern ($\eta_j$ is zero for the weights which are equal to zero). \myfig{grouplasso-figures} illustrates several of our theoretical results: (a) the top row corresponds to a situation where the strict consistency condition is satisfied and thus we obtain model consistent estimates with also a good estimation of the loading vectors (in the figure, only the good behavior of the norms of these loading vectors are represented); (b) the right column corresponds to the adaptive weighting schemes which also always achieve the two type of consistency; (c) in the middle and bottom rows, the consistency condition was not satisfied, and in the bottom row the condition of Proposition~\myfig{grouplasso-figures} that ensures that we can get model consistent estimates without regular consistency, is met, while it is not in the middle row: as expected, in the bottom row, we get some model consistent estimates but with bad norm estimation.

\begin{figure}
\begin{center}
\includegraphics[scale=.5]{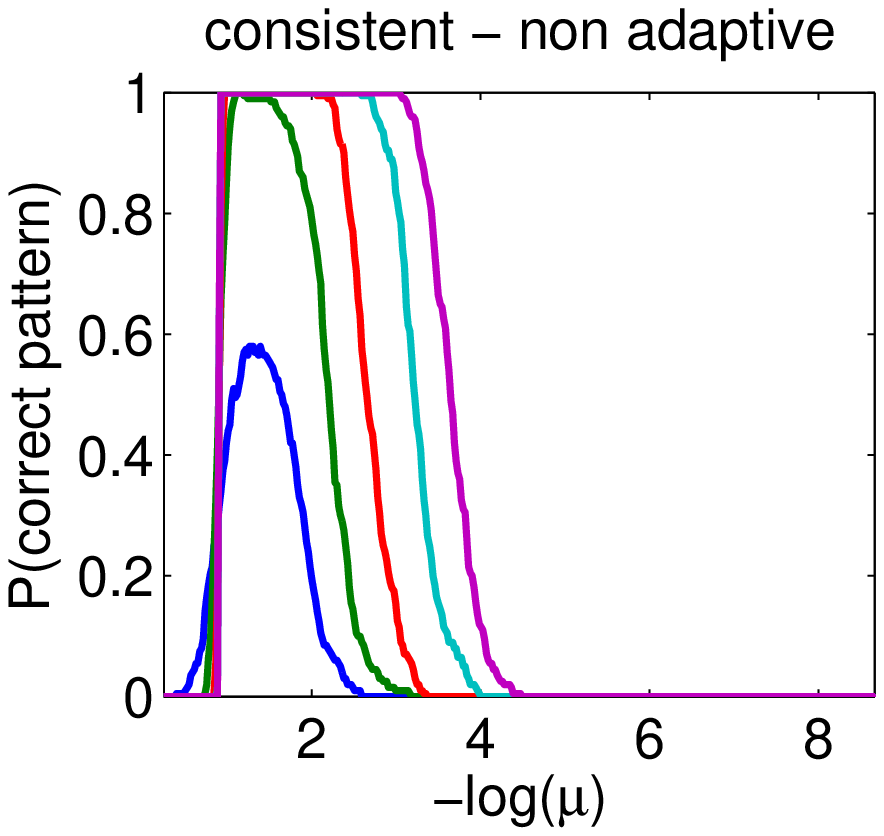} \hspace*{.5cm}
\includegraphics[scale=.5]{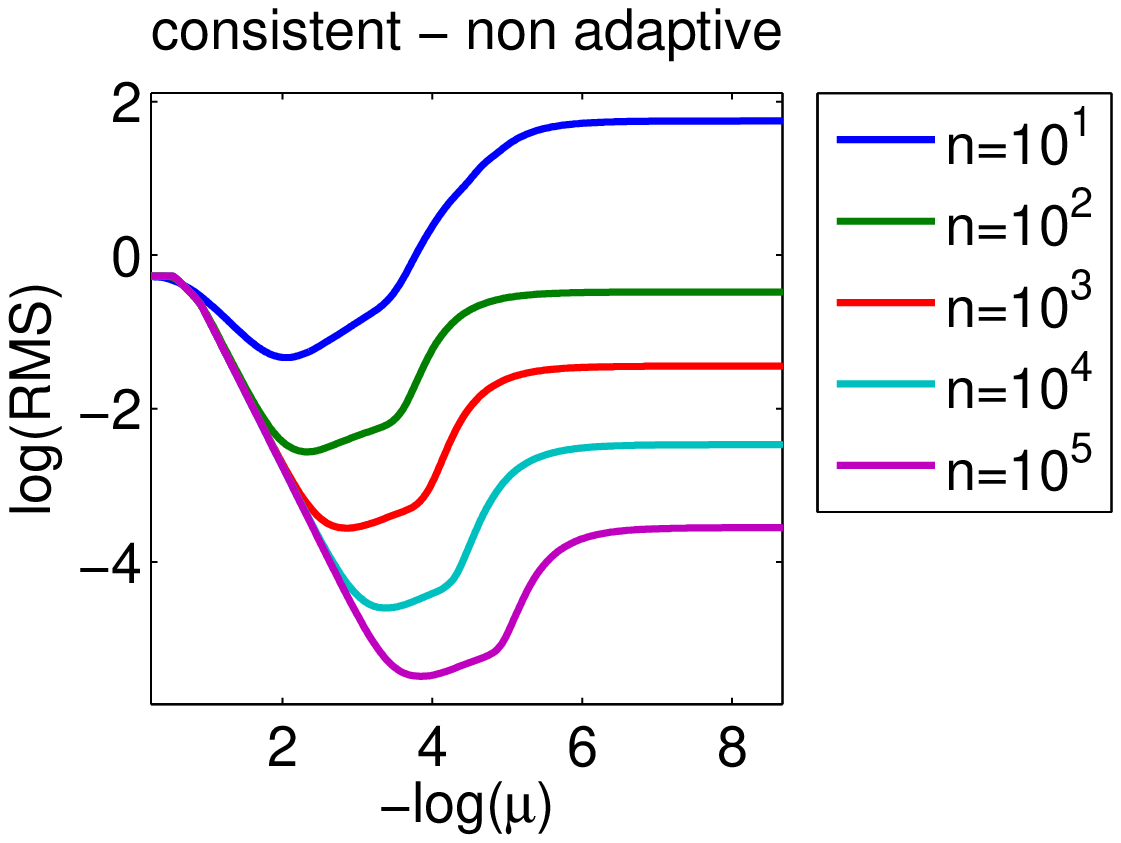}

\vspace*{.25cm}
 
\includegraphics[scale=.5]{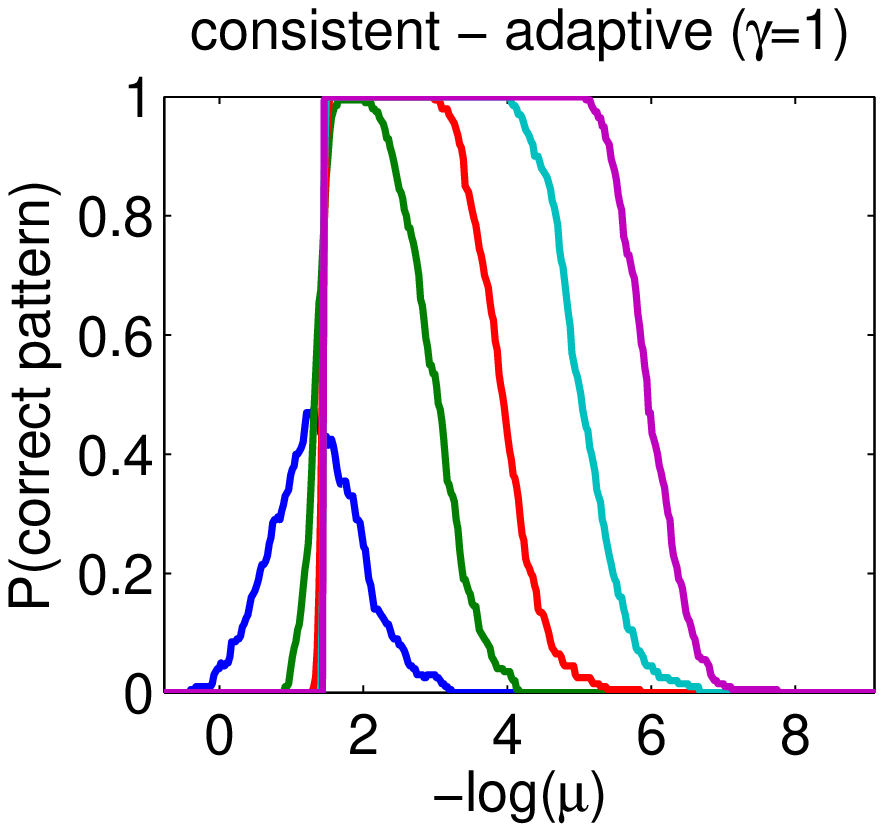} \hspace*{.5cm}
\includegraphics[scale=.5]{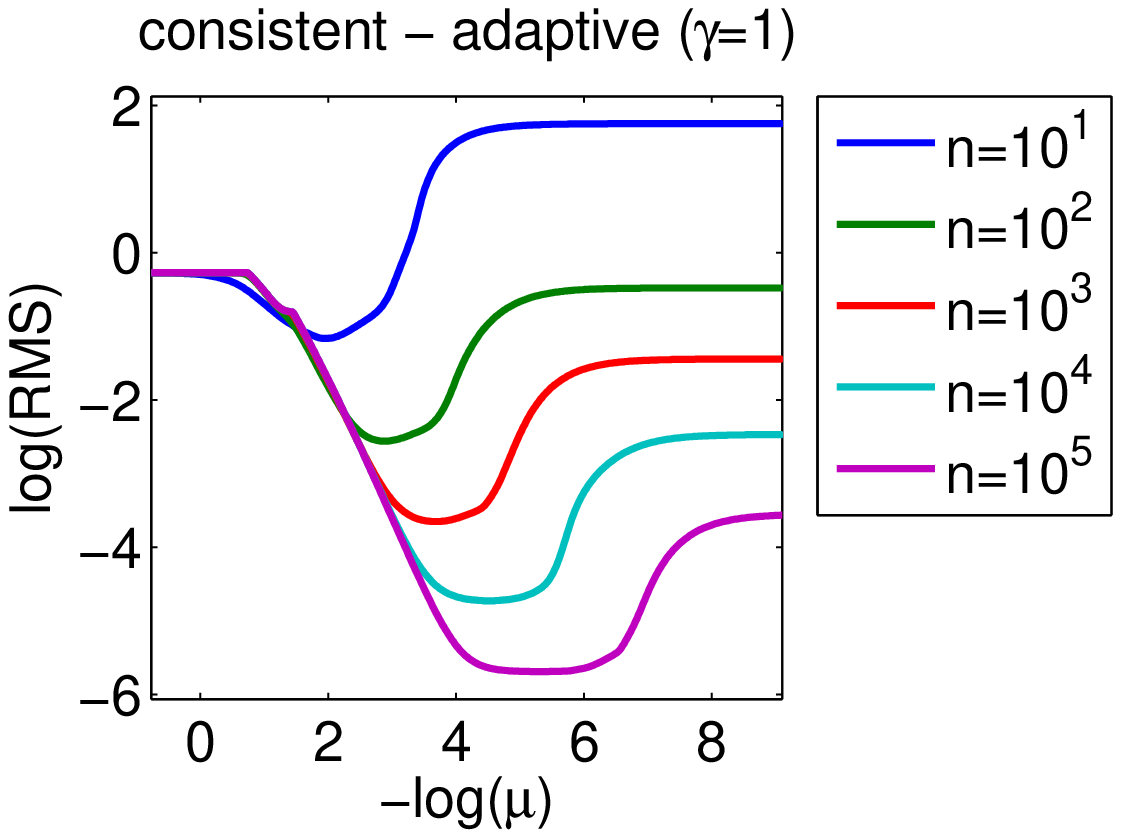}

\vspace*{-.75cm}

\end{center}
\caption{ Synthetic example where consistency condition in \eq{condition} is satisfied (same example as the top of \myfig{grouplasso-figures}: probability of correct pattern selection (\emph{left}) and  logarithm of the expected mean squared estimation error (\emph{right}), for several number of samples as a function of the regularization parameter, for regular regularization (\emph{top}), adaptive regularization with $\gamma=1 $ (\emph{bottom}).
}
\label{fig:grouplasso-ranks-1}
\end{figure}

\begin{figure}
\begin{center}
\includegraphics[scale=.5]{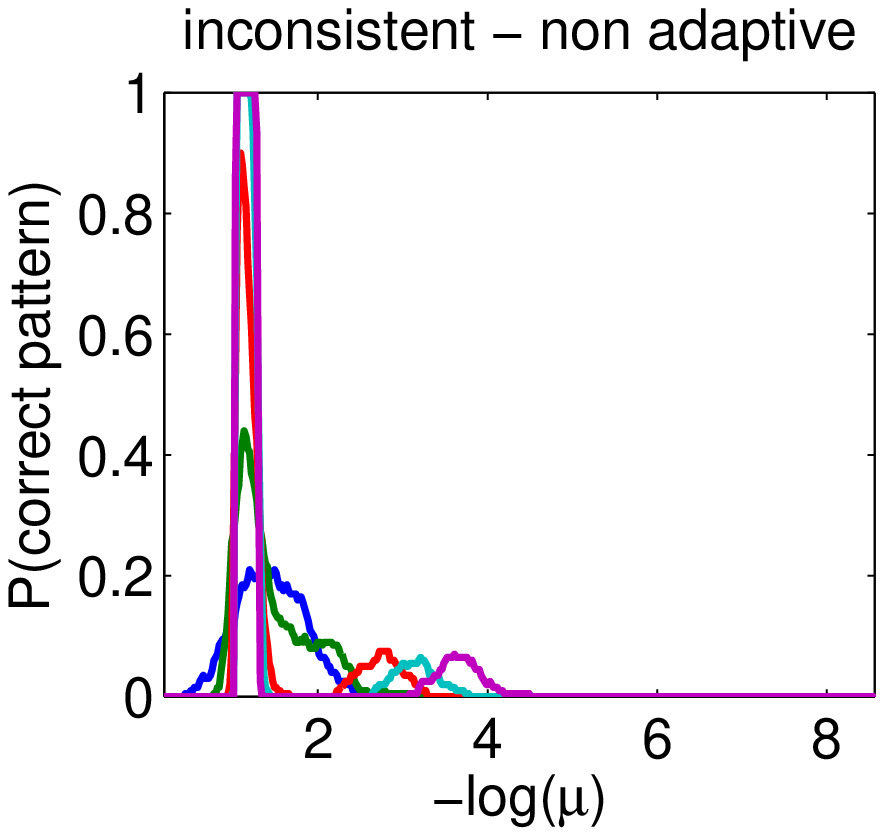} \hspace*{.5cm}
\includegraphics[scale=.5]{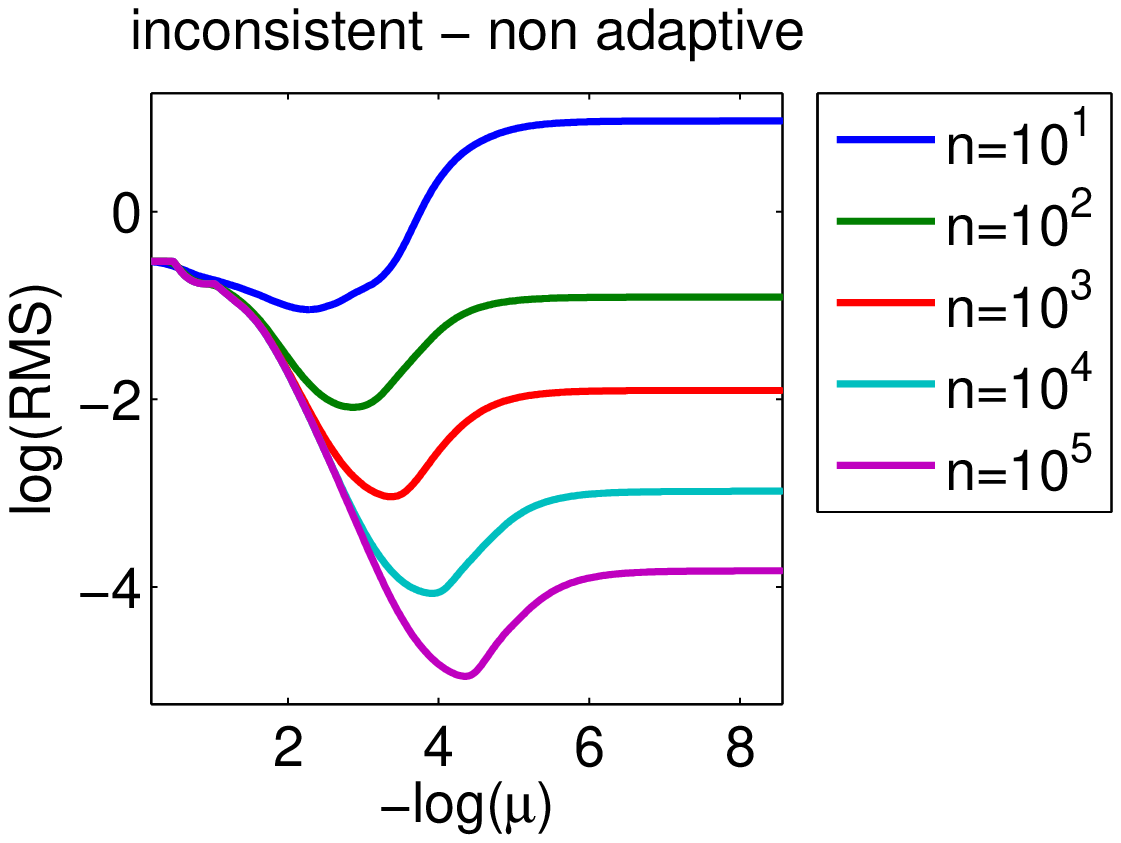}

\vspace*{.25cm}

\includegraphics[scale=.5]{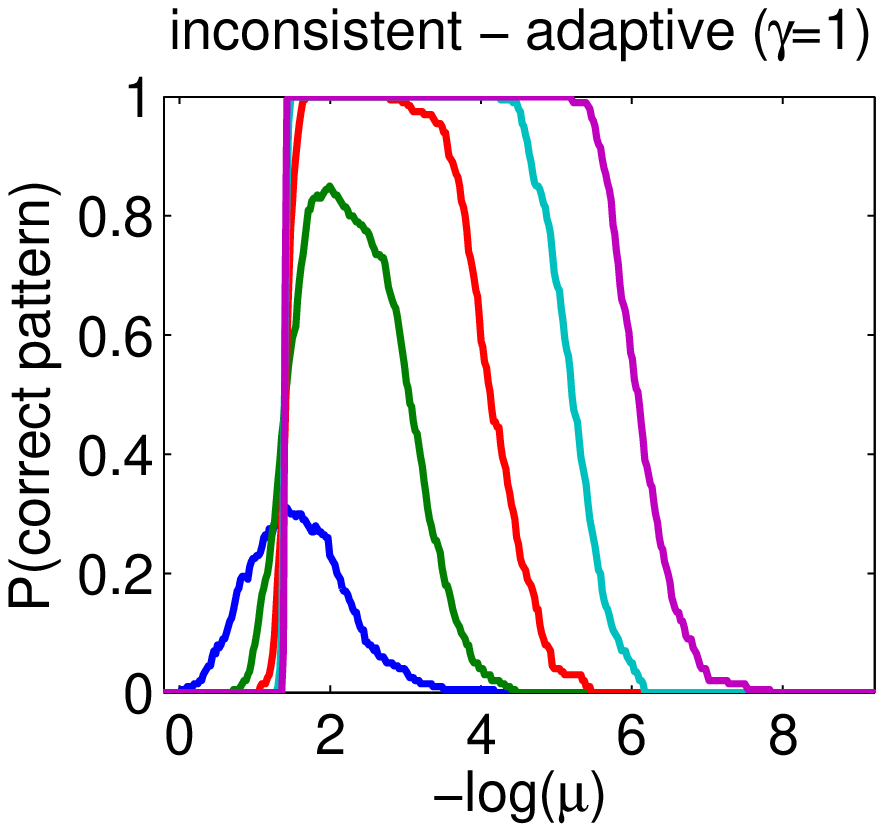} \hspace*{.5cm}
\includegraphics[scale=.5]{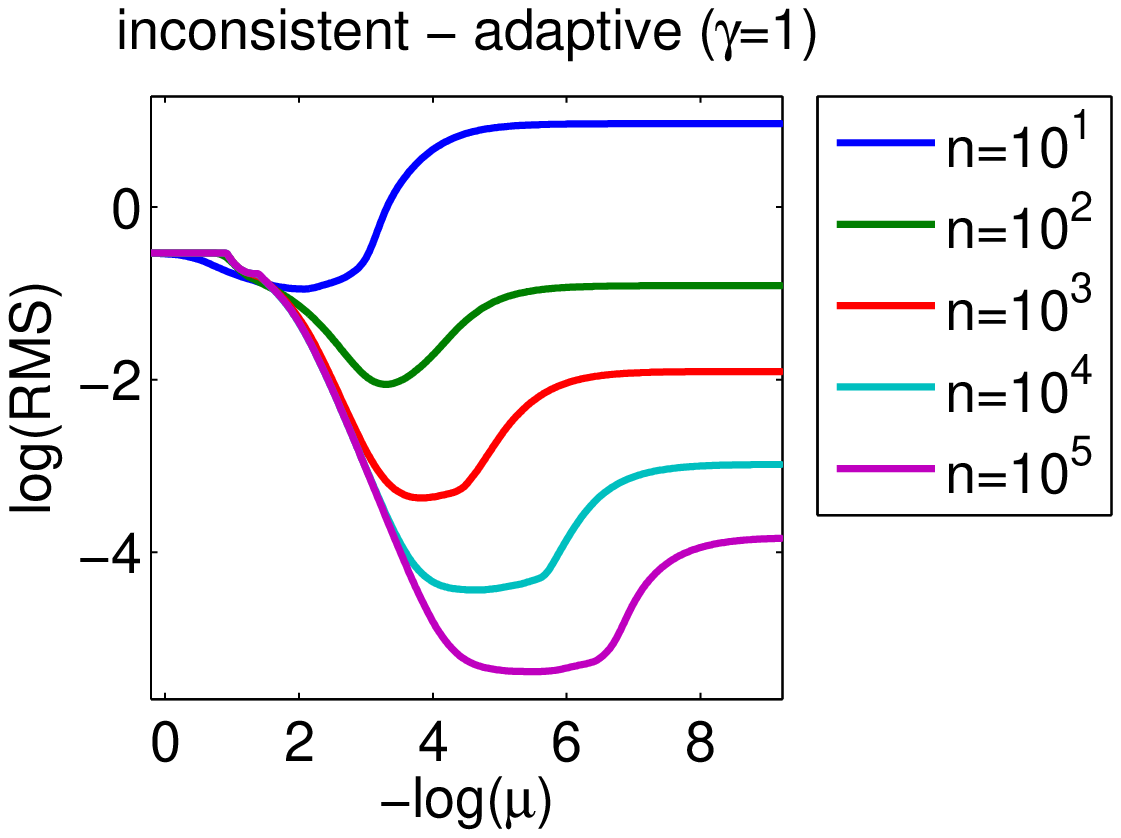}

\vspace*{-.75cm}

\end{center}
\caption{ Synthetic example where consistency condition in \eq{condition-weak} is not satisfied (same example as the middle of \myfig{grouplasso-figures}: probability of correct pattern selection (\emph{left}) and  logarithm of the expected mean squared estimation error (\emph{right}), for several number of samples as a function of the regularization parameter, for regular regularization (\emph{top}), adaptive regularization with $\gamma=1 $ (\emph{bottom}).
}
\label{fig:grouplasso-ranks-2}
\end{figure}

\begin{figure}
\begin{center}
\includegraphics[scale=.5]{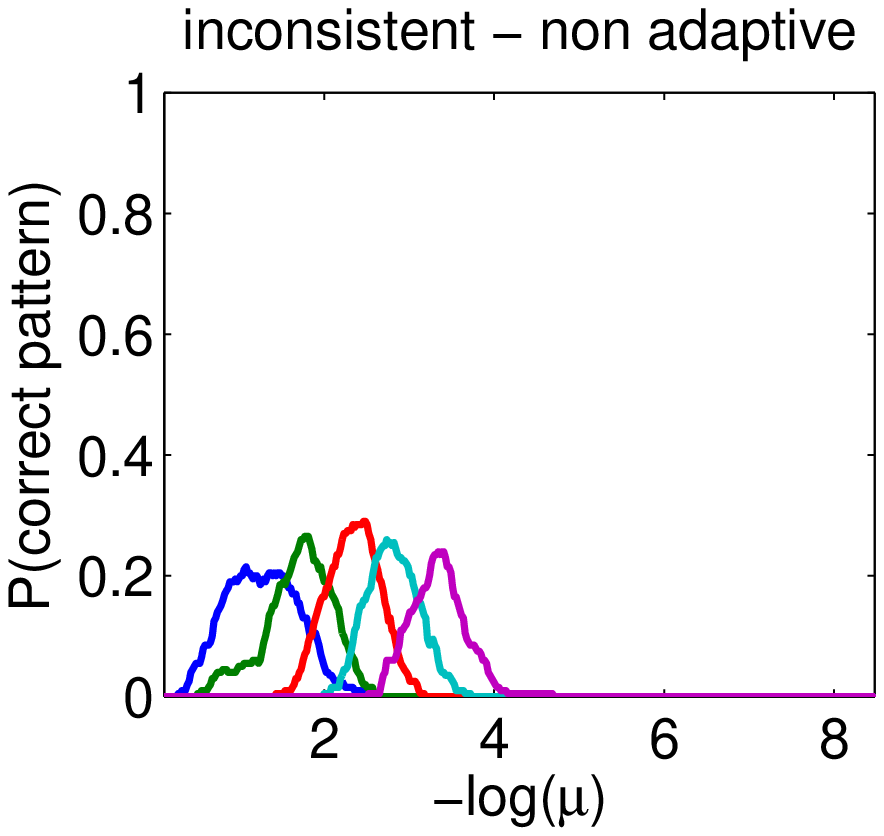} \hspace*{.5cm}
\includegraphics[scale=.5]{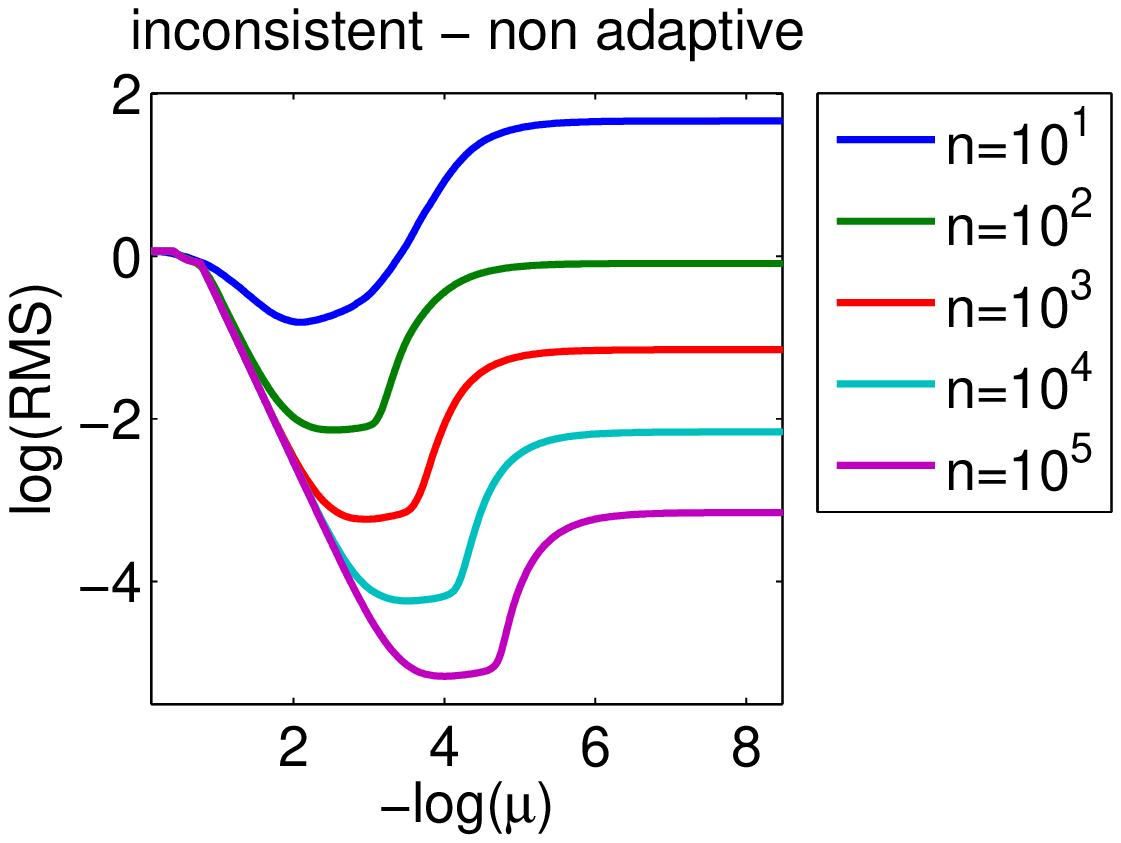}

\vspace*{.25cm}
 
\includegraphics[scale=.5]{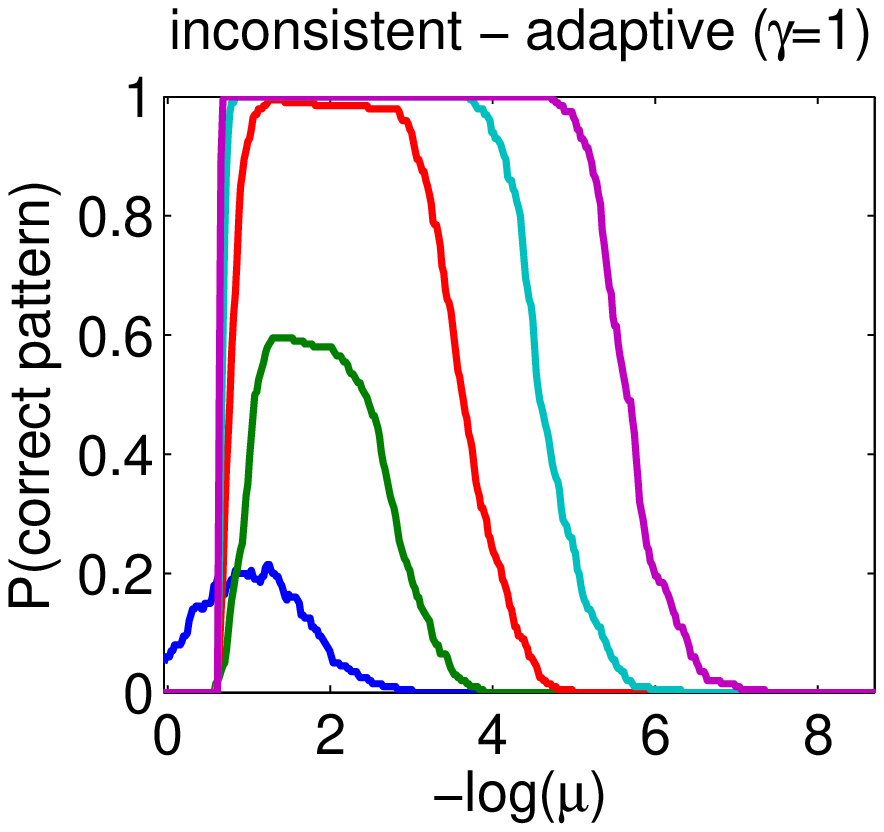} \hspace*{.5cm}
\includegraphics[scale=.5]{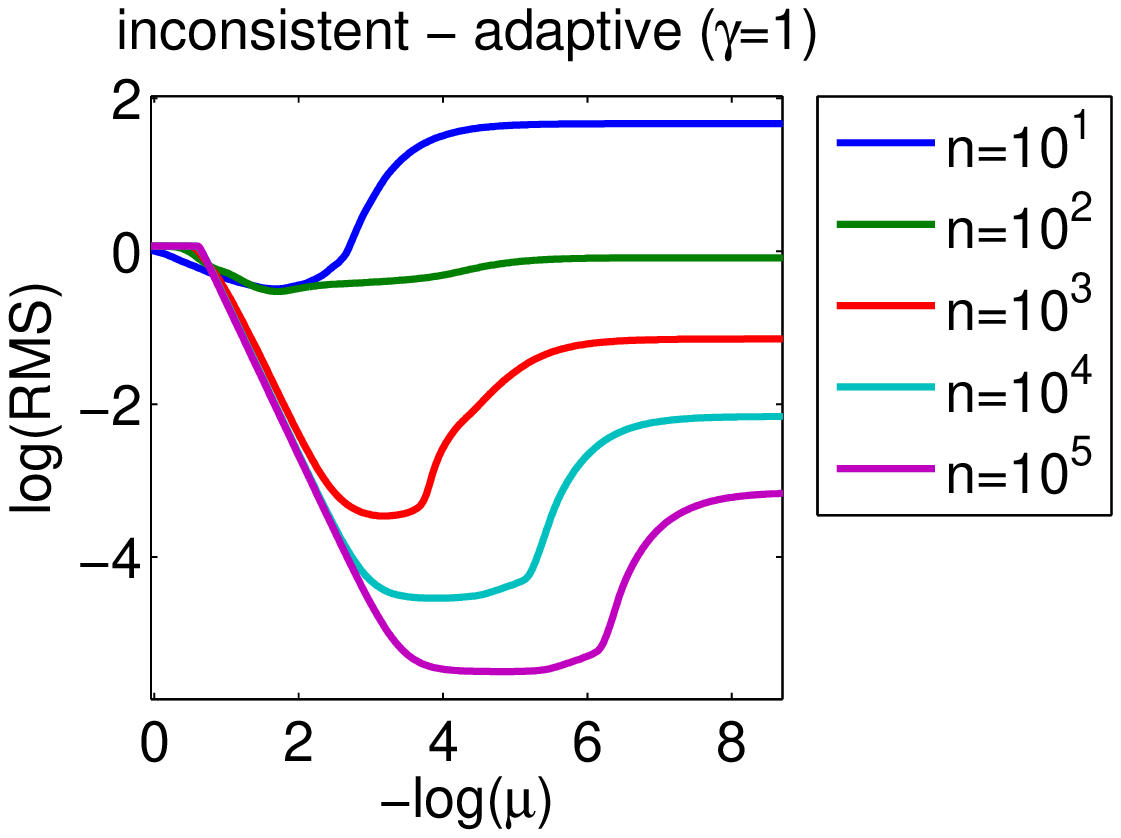}

\vspace*{-.75cm}

\end{center}
\caption{ Synthetic example where consistency condition in \eq{condition-weak} is not satisfied (same example as the bottom of \myfig{grouplasso-figures}: probability of correct pattern selection (\emph{left}) and  logarithm of the expected mean squared estimation error (\emph{right}), for several number of samples as a function of the regularization parameter, for regular regularization (\emph{top}), adaptive regularization with $\gamma=1 $ (\emph{bottom}).
}
\label{fig:grouplasso-ranks-3}
\end{figure}

In \myfig{grouplasso-ranks-1},~\ref{fig:grouplasso-ranks-2} and~\ref{fig:grouplasso-ranks-3}, we consider the three joint distributions used in \myfig{grouplasso-figures} and compute regularization paths for several number of samples ($10$ to $10^5$) with 200 replications. This allows us to estimate both the probability of correct pattern estimation $\P(J(\hat{w} = \J)$ which is considered in \mysec{proba}, and the logarithm of the expected error $\log \E \| \hat{w} - \w \|^2$.

From \myfig{grouplasso-ranks-1}, it is worth noting (a) the regular spacing between the probability of correct pattern selection for several equally spaced (in log scale) numbers of samples, which corroborates the asymptotic result in \mysec{proba}. Moreover, (b) in both row, we get model consistent estimates with increasingly smaller norms as the number of samples grow. Finally, (c) the mean square errors are smaller for the adaptive weighting scheme.

From \myfig{grouplasso-ranks-2}, it is worth noting that (a) in the non adaptive case, we have two regimes for the probability of correct pattern selection: a regime corresponding to Proposition~\ref{prop:probab} where this probablity can take values in $[0,1)$ for increasingly smaller regularization parameters (when $n$ grows); and a regime corresponding to non vanishing limiting regularization parameters corresponding to Proposition~\ref{prop:refined}: we have model consistency without regular consistency. Also, (b) the adaptive weighting scheme allows both consistencies. In \myfig{grouplasso-ranks-2} however, the second regime (correct model estimates, inconsistent estimation of loadings) is not present.

\begin{figure}
\begin{center}
\includegraphics[scale=.65]{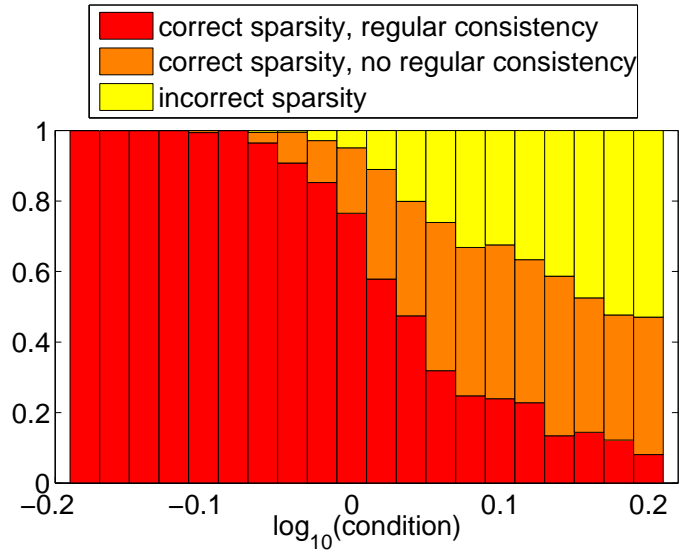}

\end{center}
\caption{Consistency of estimation vs. consistency condition. See text for details.}
\label{fig:grouplasso-histograms}
\end{figure}

In \myfig{grouplasso-histograms},  we sampled 10,000 different covariance matrices and loading vectors using the procedure described above. For each of these we computed the regularization paths from 1000 samples, and we classify each path into three categories: (1) existence of model consistent estimates with estimation error $\|\hat{w}-\w\|$ less than $10^{-1}$, (2) existence of model consistent estimates but none with estimation error $\|\hat{w}-\w\|$ less than $10^{-1}$ and (3) non existence of model consistent estimates. In \myfig{grouplasso-histograms} we plot the proportion of each of the three class as a function of the logarithm of $\max_{ i \in \J^c } \frac{1}{d_i} \left\|  \S_{X_i X_\J  } \S_{X_\J  X_\J }^{-1}
\Diag  ( d_j/ \| \w_j \|) \wJ  \right\|$. The position of the previous value with respect to 1 is indicative of the expected model consistency. When it is less than one, then we get with overwhelming probability model consistent estimates with good errors. As the condition gets larger than one, we get fewer such good estimates and more and more model inconsistent estimates.

\subsection{Nonparametric Case}
 In the infinite dimensional group case, we sampled $X \in \rb^m$ from a normal distribution with zero mean vector and a covariance matrix of size $m=4$, generated as follows: (a) sample a $m\times m$ matrix $G$ with independent standard normal distributions, (b) form $\S_{XX} = GG^\top$, (c) for each $j \in \{1,\dots,m\}$, rescale  $X_j \in \rb$ so that  $\S_{X_j X_j} = 1$.
 
\begin{figure}
\begin{center}
\includegraphics[scale=.61]{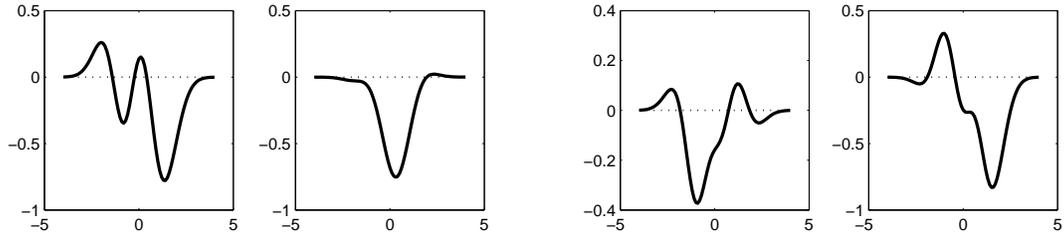}

\vspace*{-.5cm}

\end{center}
\caption{Functions to be estimated in the synthetic non parametric group Lasso experiments (left: consistent case, right: inconsistent case).}
\label{fig:functions}
\end{figure}

We use the same Gaussian kernel for each variables, $k(x,x') = e^{-(x-x')^2}$. In this situation, as shown in \myapp{covgauss} we can compute in closed form the eigenfunctions and eigenvalues of the marginal covariance operators. We then sample function from random independent components on the first 10 eigenfunctions. Examples are given in \myfig{functions}.

\begin{figure}
\begin{center}
\includegraphics[scale=.55]{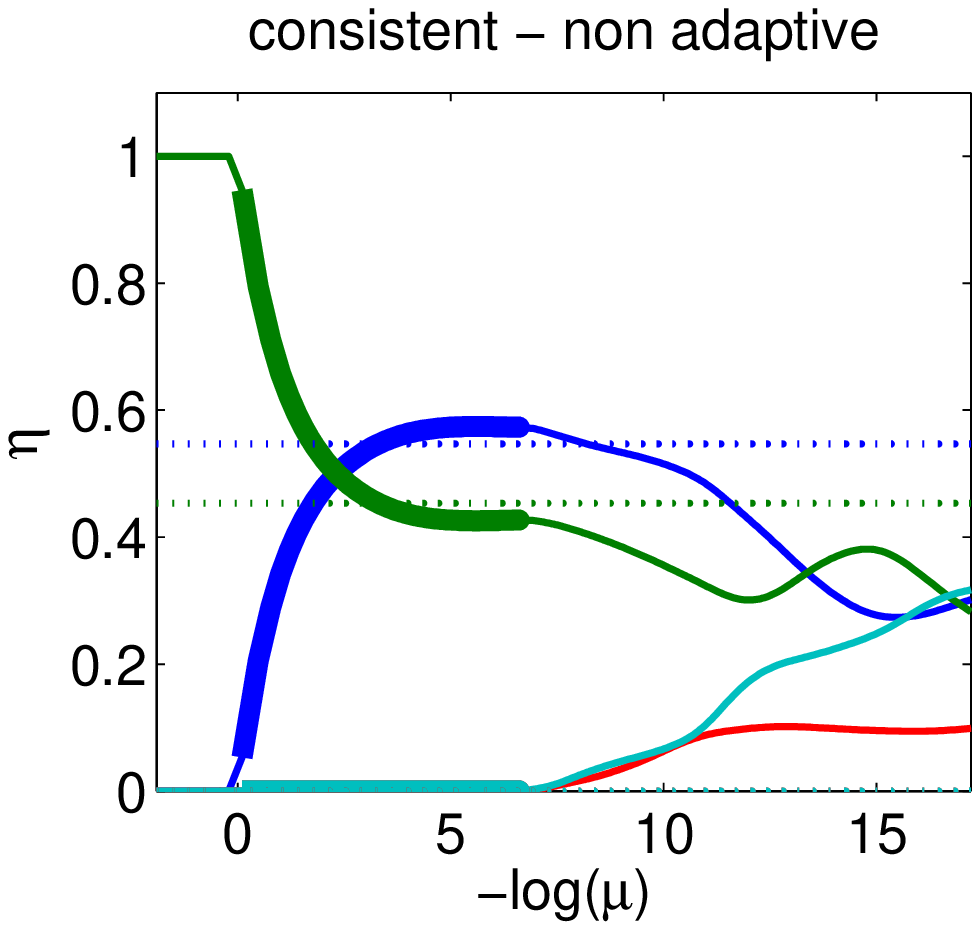} \hspace*{.5cm}
\includegraphics[scale=.55]{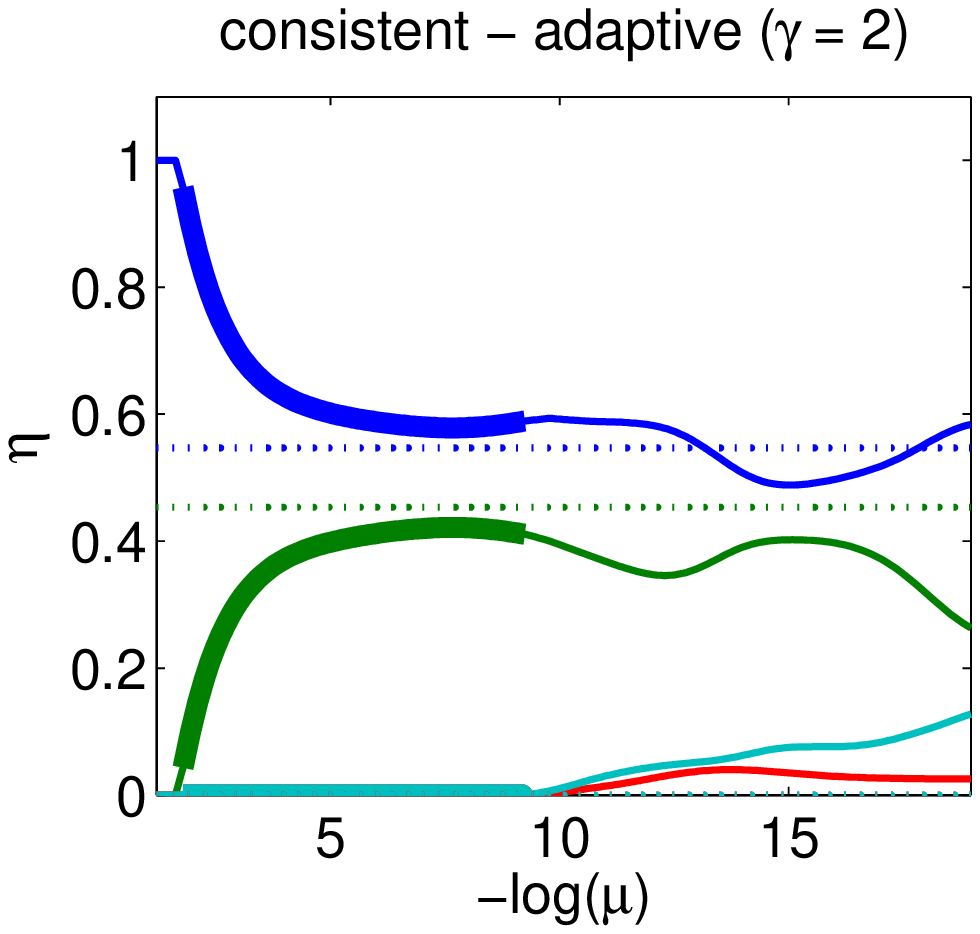}

\vspace*{.25cm}

\includegraphics[scale=.55]{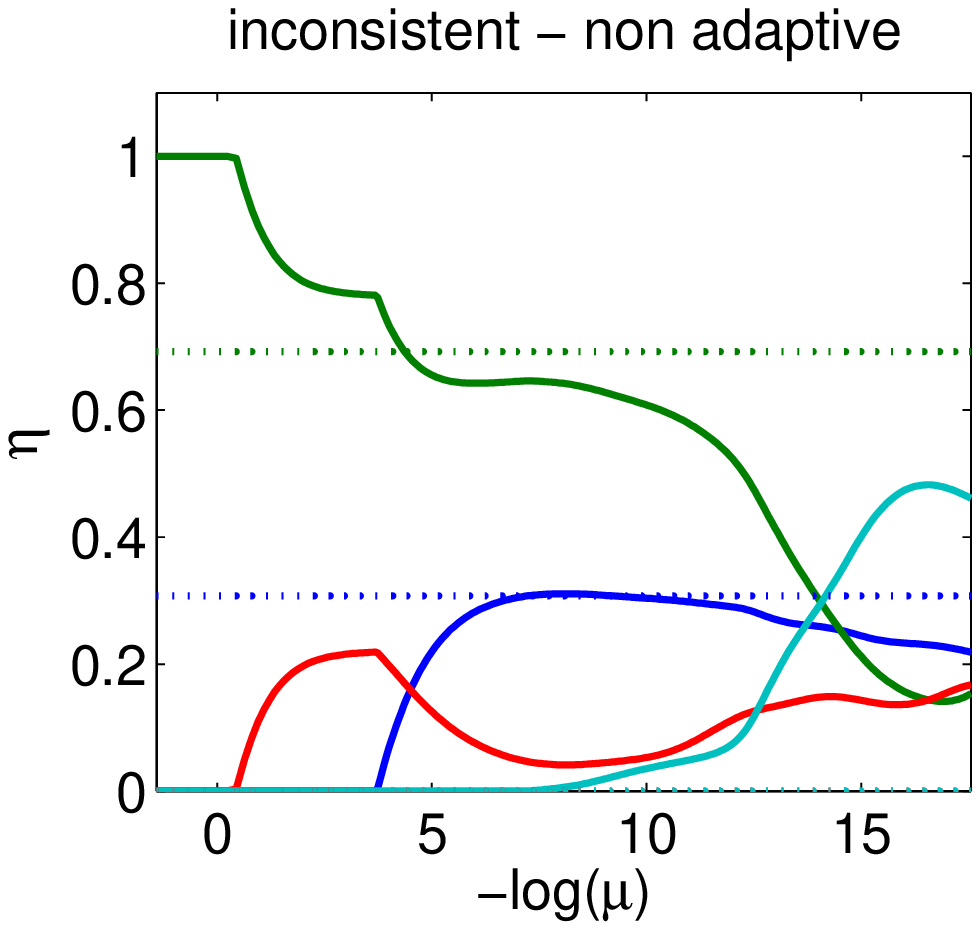}  \hspace*{.5cm}
\includegraphics[scale=.55]{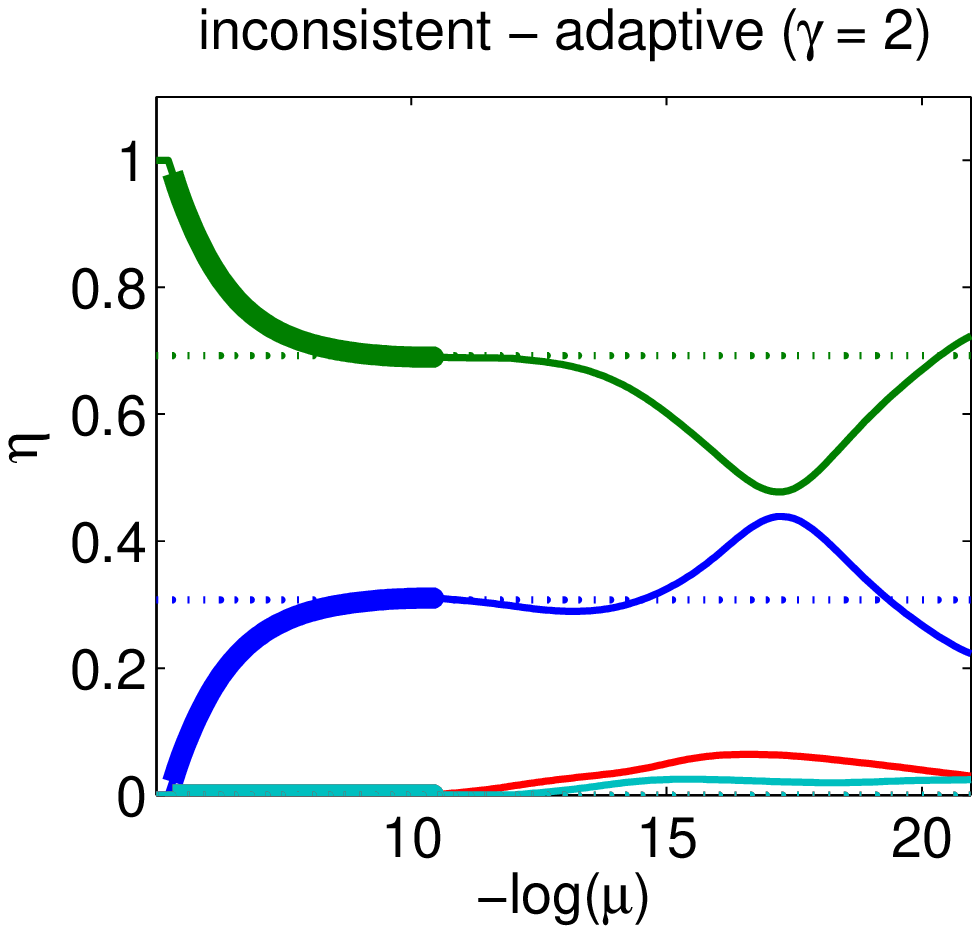}

\vspace*{-.5cm}

\end{center}
\caption{Regularization paths for the group Lasso for two weighting schemes (\emph{left}: non adaptive, \emph{right}:  adaptive) and two different population densities (\emph{top}: strict consistency condition satisfied, \emph{bottom}: weak condition not satisfied. For each of the plots, 
 plain curves correspond to values of estimated $\hat{\eta}_j$, dotted curves to population values $\eta_j$, and bold curves
to model consistent estimates. }
\label{fig:mkl-figures}
\end{figure}

In \myfig{mkl-figures}, we plot the regularization paths corresponding to 1000 i.i.d. samples,
computed by the algorithm of~\citet{bach_thibaux}. We only
plot the values of the estimated variables $\hat{\eta}_j, j =1,\dots,m$ for the alternative formulation in
\mysec{alt},
 which are proportional to $\|\hat{w}_j\|$ and normalized so that
$\sum_{j=1}^m \hat{\eta}_j   = 1$. We compare them to the population values $\eta_j$: both in terms of values, and in terms of
their sparsity pattern ($\eta_j$ is zero for the weights which are equal to zero). \myfig{mkl-figures} illustrates several of our theoretical results: (a) the top row corresponds to a situation where the strict consistency condition is satisfied and thus we obtain model consistent estimates with also a good estimation of the loading vectors (in the figure, only the good behavior of the norms of these loading vectors are represented); (b) in the bottom row, the consistency condition was not satisfied, and we do not get good model estimates. Finally,  (b) the right column corresponds to the adaptive weighting schemes which also always achieve the two type of consistency.However, such schemes should be used with care, as there is one added free parameter (the regularization
parameter $\kappa$ of the least-square estimate used to define the weights): if chosen too large, all adaptive weights are equal, and thus 
there is no adaptation, while if chosen too small, the least-square estimate may overfit.

\section{Conclusion}
In this paper, we have extended some of the theoretical results of the Lasso to the group Lasso, for finite dimensional
groups and infinite dimensional groups. In particular, under practical assumptions regarding the distributions
the data are sampled from, we have provided necessary and sufficient conditions for model consistency of the group Lasso and its nonparametric
version, multiple kernel learning.

The current work could be extended in several ways: first, a more detailed study of the limiting distributions of the group Lasso and
adaptive group Lasso estimators could be carried and then extend the analysis of~\citet{zou} or~\citet{agreg} and \citet{multi}, in particular regarding
convergence rates. Second, our results
should extend to generalized linear models, such as logistic regression~\citep{grouplog}. Also, it is of interest to let the number $m$
of groups or kernels to grow unbounded and extend the results of~\citet{Zhaoyu} and~\citet{yuinfinite} to the group Lasso. Finally, similar analysis may be carried through for more general norms with different sparsity inducing properties~\citep{tracenorm}.

\vspace*{1cm}

\appendix

\section{Proof of Optimization Results}
In this appendix, we give detailed proofs of the various propositions on optimality
conditions and dual problems.

\subsection{Proof of Proposition~\ref{prop:opt}}
\label{app:opt}
We rewrite problem in \eq{problem}, in the form
$$
\min_{w \in \rb^p,\ v \in \rb^m} \ 
\frac{1}{2} \hS_{YY} -  \hS_{YX} w + \frac{1}{2} w^\top
 \hS_{XX} w  + \lambda_n \sum_{j=1}^m d_j v_j,
$$
with added constraints that $\forall j, \| w_j \| \leqslant v_j$. In order to deal with these constraints we use the tools
from conic programming with the second-order cone, also known as the ``ice cream'' cone~\citep{boyd}.
We consider the Lagrangian with dual variables $(\beta_j,\gamma_j) \in \rb^{p_j} \times \rb$ such that
$\|\beta_j \| \leqslant \gamma_j$:
$$
\mathcal{L}(w,v,\beta,\gamma) 
=  \frac{1}{2} \hS_{YY} -  \hS_{YX} w + \frac{1}{2} w^\top
 \hS_{XX} w  + \lambda_n d^\top v - \sum_{j=1}^m 
 { w_j \choose v_j}^\top { \beta_j \choose \gamma_j}.
 $$
The derivatives with respect to primal variables are
\BEAS
\nabla_{w} \mathcal{L}(w,v,\beta,\gamma)  & = &   \hS_{XX} w -  \hS_{XY} - \beta ,\\
 \nabla_{v} \mathcal{L}(w,v,\beta,\gamma) &  = &  \lambda_n d - \gamma. 
\EEAS
At optimality, primal and dual variables are completely characterized by $w$ and $\beta$. 
Since the dual and the primal problems are strictly feasible,
strong duality holds and
the KKT conditions for reduced primal/dual variables $(w,\beta)$ are
\BEA
\forall j, \ \|\beta_j\| \leqslant \lambda_n d_j & &  \mbox{ (dual feasibility) },\\
\forall j,\  \beta_j =  \hS_{X_j X} w -  \hS_{X_j Y} & & \mbox{ (stationarity) }, \\
\forall j,\  \beta_j^\top w_j + \|w_j\| \lambda_n d_j = 0 & & \mbox{ (complementary slackness) }.
\EEA
Complementary slackness for the second order cone has special consequences:
$w_j^\top \beta_j +  \|w_j\| \lambda_n d_j = 0 $ if and only if~\citep{boyd,socp}, either
(a) $w_j = 0$,
or 
 (b) $w_j \neq 0$, $  \|\beta_j\| = \lambda_n d_j $ and 
 $\exists \eta_j > 0$ such that $w_j = - \frac{\eta_j}{\lambda_n} \beta_j$ (anti-proportionality),
which implies  
 $\beta_j = - w_j \frac{ \lambda_n d_j}{\| w_j\|}$ and $\eta_j = \|w_j\|/d_j$.
This leads to the proposition.

\subsection{Proof of Proposition~\ref{prop:opt-alt}}
\label{app:opt-alt}
We follow the proof of  Proposition~\ref{prop:opt} and of~\citet{skm}.
We rewrite problem in \eq{problem-alt}, in the form
$$
\min_{w \in \rb^p,\ v \in \rb^m,\  t \in \rb} \ 
\frac{1}{2} \hS_{YY} -  \hS_{YX} w + \frac{1}{2} w^\top
 \hS_{XX} w  +  \frac{1}{2} \mu_n  t^2 ,
$$
with constraints that $\forall j, \| w_j \| \leqslant v_j$ and $ d^\top v \leqslant t$.
We consider the Lagrangian with dual variables $(\beta_j,\gamma_j) \in \rb^{p_j} \times \rb$ 
and $\delta \in \rb_+$ such that
$\|\beta_j \| \leqslant \gamma_j$, $j=1,\dots,m$:
$$
\mathcal{L}(w,v,\beta,\gamma,\delta) 
= \frac{1}{2} \hS_{YY} -  \hS_{YX} w + \frac{1}{2} w^\top
 \hS_{XX} w  + \frac{1}{2} \mu_n t^2- \beta^\top w - \gamma^\top v  + \delta( d^\top v - t ).
$$
The derivatives with respect to primal variables are
\BEAS
 \nabla_{w} \mathcal{L}(w,v,\beta,\gamma) &  =  & \hS_{XX} w -  \hS_{XY} - \beta , \\
 \nabla_{v} \mathcal{L}(w,v,\beta,\gamma)  & =  & \delta d - \gamma ,   \\
\nabla_{t} \mathcal{L}(w,v,\beta,\gamma) &  = &  \mu_n t  - \delta.
\EEAS
At optimality, primal and dual variables are completely characterized by $w$ and $\beta$.
Since the dual and the primal problems are strictly feasible,
strong duality holds and
the KKT conditions for reduced primal/dual variables $(w,\beta)$ are
\BEA
\forall j, \beta_j =  \hS_{X_j X} w -  \hS_{X_j Y} & & \mbox{ (stationarity - 1) }, \\
\label{eq:26} \forall j, \sum_{j=1}^m d_j \| w_j \|  = \frac{1}{\mu_n} \max_{ i =1,\dots,m} \frac{ \| \beta_i\|}{d_i}  & & \mbox{ (stationarity - 2) }, \\
\forall j, \left( \frac{\beta_j}{d_j} \right)^\top w_j + \|w_j\|  \max_{ i=1,\dots,m }
 \frac{ \| \beta_i \|}{d_i} = 0 & & \mbox{ (complementary slackness) }.
\EEA
Complementary slackness for the second order cone implies that:
$$\left( \frac{\beta_j}{d_j} \right)^\top w_j + \|w_j\|  \max_{ i =1,\dots,m} \frac{ \| \beta_i \|}{d_i} = 0,
$$ if and only if, either
(a) $w_j = 0$,
or
 (b) $w_j \neq 0$ and $  \frac{\|\beta_j\|}{d_j} = \displaystyle \max_{ i =1,\dots,m} \frac{ \| \beta_i \|}{d_i}  $,
   and $\exists \eta_j \geqslant 0$
 such that  $w_j = - \eta_j \beta_j / \mu_n $, which implies
 $\|w_j\| = \frac{\eta_j d_j }{\mu_n} \displaystyle  \max_{ i =1,\dots,m} \frac{ \| \beta_i \|}{d_i}$.

By writing $\eta_j=0$ if $w_j =0$ (i.e., in order to cover all cases), we have  from \eq{26}
$\sum_{j=1}^m d_j \| w_j \|  = \frac{1}{\mu_n} \displaystyle \max_{ i=1,\dots,m } \frac{ \| \beta_i \|}{d_i} $, which implies
$\sum_{j=1}^m d_j^2 \eta_j = 1$
and thus $ \forall j, \ \eta_j = \frac{ \| w_j \| / d_j }{ \sum_i d_i \| w_i\| } $. This leads
to $\forall j, \beta_j = - w_j \mu_n / \eta_j = - \frac{w_j} {\|w_j\|} \sum_{i=1}^n d_i \|w_i\|$. The proposition follows.

\subsection{Proof of Proposition~\ref{prop:opt-alt-cov}}
\label{app:opt-alt-cov}
By following the usual proof of the representer theorem~\citep{wahba}, we obtain
that each optimal function $f_j$ must be
 supported by the data points, i.e., there exists $\alpha = (\alpha_1,\dots,\alpha_m) \in \rb^{n \times m}$ such
that for all $j=1,\dots,m$, $f_j = \sum_{i=1}^n \alpha_{ij} k_j(\cdot,x_{ij})$. When using this representation
back into \eq{problem-alt-cov}, we obtain an optimization problem that only depends
on $\phi_j = G_j^\top \alpha_j$ for $j=1,\dots,m$ where $G_j$ denotes any square root of the kernel matrix
$K_j$, i.e.,  $K_j = G_j G_j^\top.$ This problem is exactly the finite dimensional
problem in \eq{problem-alt}, where $\X_j$ is replaced by $G_j$ and $w_j$ by $\phi_j$. Thus
Proposition~\ref{prop:opt-alt} applies and we can easily derive the current proposition by expressing
all terms through the functions $f_j$. Note that in this proposition, we do not show that
the $\alpha_j$, $j=1,\dots,m$, are all proportional to the same vector, as is done in \myapp{opt-alt-dual}.

\subsection{Proof of Proposition~\ref{prop:opt-alt-dual}}
\label{app:opt-alt-dual}
We prove the proposition in the linear case. Going to the general case, can be done in the same way as
done in \myapp{opt-alt-cov}. We let $\X$ denote the covariate matrix in $\rb^{n \times p}$;
we simply need to add a new variable $u =   \X w + b1_n$ and to ``dualize'' it. That is,
we rewrite problem in \eq{problem-alt}, in the form
$$
\min_{w \in \rb^p, \ b \in \rb,\ v \in \rb^m,\  t \in \rb, \ u \in \rb^n} \ 
\frac{1}{2n} \| \Y - u \|^2 +  \frac{1}{2} \mu_n  t^2 ,
$$
with constraints that $\forall j, \| w_j \| \leqslant v_j$,  $ d^\top v \leqslant t$ and
$ \X w + b 1_n= u$.
We consider the Lagrangian with dual variables $(\beta_j,\gamma_j) \in \rb^{p_j} \times \rb$ 
and $\delta \in \rb_+$ such that
$\|\beta_j \| \leqslant \gamma_j$, and $\alpha \in \rb^n$:
$$
\mathcal{L}(w,b,v,u,\beta,\gamma,\alpha,\delta) 
= \frac{1}{2n} \| \Y - u \|^2 + \mu_n \alpha^\top ( u - \X w ) +   \frac{1}{2} \mu_n t^2- \sum_{j=1}^m \left\{
\beta_j^\top w_j + \gamma_j v_j \right\} + \delta( d^\top v - t ).
$$
The derivatives with respect to primal variables are
\BEAS
 \nabla_{w} \mathcal{L}(w,v,u,\beta,\gamma,\alpha)  &  =  & - \mu_n \X^\top\alpha  - \beta \\
 \nabla_{v} \mathcal{L}(w,v,u,\beta,\gamma,\alpha)   & =  & \delta d - \gamma   \\
\nabla_{t} \mathcal{L}(w,v,u,\beta,\gamma,\alpha)  &  = &  \mu_n t  - \delta \\
\nabla_{u} \mathcal{L}(w,v,u,\beta,\gamma,\alpha)  &  = &  \frac{1}{n}( u - \Y + \mu_n n \alpha ) \\
\nabla_{b} \mathcal{L}(w,v,u,\beta,\gamma,\alpha)  &  = &   \mu_n   \alpha^\top 1_n.
\EEAS
Equating them to zero, we get the dual problem in \eq{dual}. Since the dual and the primal problems are strictly feasible,
strong duality holds and the KKT conditions for reduced primal/dual variables $(w,\alpha)$ are
\BEA
\forall j,   \X w -  \Y + \mu_n n \alpha = 0 & & \mbox{ (stationarity - 1) }, \\
\label{eq:29} \forall j, \sum_{j=1}^m d_j \| w_j \|  =  \max_{ i =1,\dots,m}
 \frac{ (\alpha^\top {K}_i \alpha)^{1/2}}{d_i}  & & \mbox{ (stationarity - 2) }, \\
 \alpha^\top 1_n = 0  & & \mbox{ (stationarity - 3) }, \\
\forall j, \left( \frac{-  \X_j^\top \alpha}{d_j} \right)^\top w_j 
+ \|w_j\|  \max_{ i=1,\dots,m } \frac{ (\alpha^\top {K}_i \alpha)^{1/2}}{d_i} = 0 & & \mbox{ (complementary slackness) }.
\EEA
Complementary slackness for the second order cone goes leads to:
$$\left( \frac{-  \X_j^\top  \alpha}{d_j}
 \right)^\top w_j + \|w_j\|  \max_{ i =1,\dots,m} \frac{ (\alpha^\top {K}_i \alpha)^{1/2}}{d_i} = 0,$$
  if and only if, either
(a) $w_j = 0$,
or
 (b) $w_j \neq 0$ and $  \frac{(\alpha^\top {K}_j \alpha)^{1/2}}{d_j} = \displaystyle \max_{ i =1,\dots,m} \frac{ (\alpha^\top {K}_i \alpha)^{1/2} }{d_i}  $, 
  and $\exists \eta_j \geqslant 0$
 such that  $w_j = - \eta_j \left( -  \X_j^\top  \alpha \right)  $, which implies
 $\|w_j\| =  \eta_j d_j   \displaystyle  \max_{ i =1,\dots,m} \frac{ (\alpha^\top {K}_i \alpha)^{1/2}}{d_i}$.
 
 By writing $\eta_j=0$ if $w_j =0$ (to cover all cases), we have  from \eq{29},
$\sum_{j=1}^m d_j \| w_j \|  =   \displaystyle \max_{ i=1,\dots,m } \frac{(\alpha^\top {K}_i \alpha)^{1/2} }{d_i} $,
which implies
$\sum_{j=1}^m d_j^2 \eta_j = 1$.  The proposition follows from the fact that at optimality, $\forall j$,
$w_j =  \eta_j   \X_j^\top  \alpha    $.

\subsection{Proof of Proposition~\ref{prop:opt-alt-dual-unique}}
\label{app:opt-alt-dual-unique}
What makes this proposition non obvious is the fact that the covariance operator $\S_{XX}$ is not invertible in general. From proposition \ref{prop:opt-alt-dual}, we know that each $f_j$ must be of the form $f_j  = \eta_j \sum_{ i=1}^n \alpha_i k_j(x_{ij},\cdot)$, where $\alpha$ is \emph{uniquely} defined. Moreover, $\eta$ is such that $
\left( \sum_{j=1}^m \eta_j {K}_j  + n \mu_n  \idm_n \right) \alpha =  \Y$ and such that 
if $   \frac{\alpha^\top {K}_j \alpha }{d_j^2} <   A$, then $\eta_j  = 0$ (where
$A = \max_{ i=1,\dots,m } \frac{\alpha^\top {K}_i \alpha }{d_i^2}
$). Thus, if the solution is not unique, there exists two vectors $\eta\neq\zeta$ such that
$\eta$ and $\zeta$ have zero components on indices $j$ such that 
$ {\alpha^\top {K}_j \alpha }  <   A d_j^2$ (we let denote $J$ the active set and thus $J^c$ this set of indices), and 
 $\sum_{j=1}^m (\zeta_j - \eta_j) {K}_j  \alpha = 0 $. This implies that the vectors
 $\Pi_n K_j \alpha = \Pi_n K_j \Pi_n \alpha$, $j \in J$ are linearly dependent. Those vectors are exactly the centered vector of values of the functions $g_j    =
 \sum_{ i=1}^n \alpha_i k_j(x_{ij},\cdot)$ at the observed data points. Thus, non unicity implies that the empirical covariance matrix of the random variables $g_j(X_j)$, $j \in J$, is non invertible. Moreover, we have
 $\|g_j\|_\Fj^2 = \alpha^\top {K}_j \alpha = d_j^2 A > 0$ and
  the empirical marginal variance of $g_j(X_j)$ is equal to  $\alpha^\top {K}_j^2 \alpha > 0$ (otherwise $\|g_j\|_\Fj^2 = 0$. By normalizing by the (non vanishing) empirical standard deviations, we thus obtain functions such that the empirical covariance matrix is singular, but the marginal empirical variance are equal to one. Because the empirical covariance operator is a consistent estimator of $\S_{XX}$ and $C_{XX}$ is invertible, we get a contradiction, which proves the  unicity of solutions.

\section{Detailed Proofs for the Group Lasso}
In this appendix, detailed proofs of the consistency results for the finite dimensional case
(Theorems~\ref{theo:theo1} and~\ref{theo:theo2}) are presented.
Some of the results presented in this appendix are corollaries of the more general results in \myapp{mklproofs}, but their
proofs in the finite dimensional case are much simpler.

\subsection{Proof of Theorem~\ref{theo:theo1}}
\label{app:theo1}

We begin with a lemma, which states that if we restrict ourselves to the covariates which
we are after (i.e., indexed by $\J$), we get a consistent estimate as soon as $\lambda_n$ tends to zero:
\begin{lemma}
\label{lemma:tilde}
Assume \hypreff{var}{model}.
Let $\tilde{w}_\J $ any minimizer of
$$\frac{1}{2n} \| \Y - \X_\J  w_\J  \|^2 + \lambda_n  \sum_{j \in \J } d_j \| w_j \|
=
\frac{1}{2} \hS_{YY} -  \hS_{YX_\J } w_\J  + \frac{1}{2} w_\J ^\top
 \hS_{X_\J X_\J  } w_\J  + \lambda_n  \sum_{j \in  \J } d_j \| w_j \|.
$$
If $\lambda_n  \to 0$, then $\tilde{w}_\J $ converges to $\wJ$ in probability.
\end{lemma}
\begin{proof}
If $\lambda_n $ tends to zero, then the cost function  defining $\tilde{w}_\J $  converges to 
$F_n(w_\J  ) = \frac{1}{2} \S_{YY} -  \S_{YX_\J } w_\J  + \frac{1}{2} w_\J ^\top
 \S_{X_\J  X_\J  } w_\J  $ whose unique (because  $\S_{X_\J  X_\J }$ is positive definite) global minimum is $\wJ$ (true generating value).
 The convergence of $\tilde{w}_\J $ is thus a simple consequence of standard results in $M$-estimation~\citep{VanDerVaart,fu}.
\end{proof}

We now prove Theorem~\ref{theo:theo1}.
Let $\tilde{w}_\J $ be defined as in Lemma~\ref{lemma:tilde}. We extend it by zeros on $\J^c$. We  already
know from 
Lemma~\ref{lemma:tilde} that we have consistency in squared norm. Since with probability tending to one, the problem has a unique solution (because $\S_{XX}$ is invertible), we now need to prove that the probability that $\tilde{w}$ is optimal for problem in \eq{problem} is tending to one.  

By definition of
$\tilde{w}_\J $, the optimality condition (\ref{eq:opt2}) is satisfied. We now need to verify 
optimality condition (\ref{eq:opt1}).
Denoting $\varepsilon = Y - \w^\top X - \b  $,  we have:
$$ \hS_{X Y}=  \hS_{X X} \w + \hS_{X \varepsilon} =
\left( \S_{X X} + O_p(n^{-1/2}) \right) \w  +   O_p(n^{-1/2}) =
 \S_{X X_\J  }\wJ  + O_p(n^{-1/2}),
 $$
because of classical results on convergence of empirical covariances to covariances~\citep{VanDerVaart}, which are applicable
because we have the fourth order moment condition \hypref{var}.
We thus have:
\BEQ
\label{eq:pp1}
\hS_{X  Y} - \hS_{X  X_\J  } \tilde{w}_\J  
=  \S_{X  X_\J  } ( \wJ  - \tilde{w}_\J   ) + O_p(n^{-1/2} ).
\EEQ
From the optimality condition $ \hS_{X_\J  Y}  - \hS_{X_\J  X_\J } \tilde{w}_\J   = 
\lambda_n \Diag ( d_j/ \| \tilde{w}_j \|) \tilde{w}_\J  $ defining $\tilde{w}_\J $ and
\eq{pp1}, we obtain:
\BEQ
\label{eq:pp2}
\tilde{w}_\J  -  \wJ  = -  \lambda_n \S_{X_\J  X_\J  }^{-1} \Diag  ( d_j/ \| \tilde{w}_j \|) \tilde{w}_\J 
+ O_p(n^{-1/2} ).
\EEQ
Therefore,
\BEAS
 \hS_{X_{\J^c} Y} -  \hS_{X_{\J^c} X_\J } \tilde{w}_\J  & = & 
\S_{X_{\J^c} X_\J  } ( \wJ - \tilde{w}_\J )
+ O_p(n^{-1/2}) \mbox{ by \eq{pp1} },\\
& = &  \lambda_n \S_{X_{\J^c} X_\J  } \S_{X_\J  X_\J  }^{-1} \Diag ( d_j/ \| \tilde{w}_j \|) \tilde{w}_\J 
+ O_p(n^{-1/2}) \mbox{ by \eq{pp2}}.
\EEAS
Since $\tilde{w}$ is consistent, and  $\lambda_n n^{1/2} \to +\infty$, then for each $i \in J^c$,
$$ \frac{1}{d_i\lambda_n} \left( \hS_{X_{i} Y} -  \hS_{X_{i} X_\J } \tilde{w}_\J  \right) $$
converges in probability to $\frac{1}{d_i} \S_{X_{i} X_\J  } \S_{X_\J  X_\J  }^{-1} \Diag ( d_j/ \| \w_j \|) \w_\J 
$ which is of
norm \emph{strictly} smaller than one because  condition (\ref{eq:condition}) is satisfied. Thus the probability that $\tilde{w}$ is
indeed optimal, which is equal to
$$\P \left\{ \forall i \in \J^c, 
 \frac{1}{d_i\lambda_n} \left\| \hS_{X_{i} Y} -  \hS_{X_{i} X_\J } \tilde{w}_\J  \right\| \leqslant 1  \right\}
 \geqslant \prod_{i \in \J^c} 
 \P \left\{ 
 \frac{1}{d_i\lambda_n} \left\| \hS_{X_{i} Y} -  \hS_{X_{i} X_\J } \tilde{w}_\J  \right\| \leqslant 1  \right\},
$$
  is tending to 1,
 which implies the theorem.

\subsection{Proof of Theorem~\ref{theo:theo2}}
\label{app:theo2}
We prove the theorem by contradiction, by assuming that there exists $i \in \J^c$ such that
$$ \frac{1}{d_i}  \left\| \S_{X_i X_\J  } \S_{X_\J  X_\J }^{-1}
\Diag ( d_j/ \| \w_j \|) \wJ  \right\| > 1 . $$
Since with probability tending to one $J(\hat{w}) = \J$, with probability tending to one, we have from 
optimality condition (\ref{eq:opt2}):
$$\hat{w}_\J  =     \hS_{X_\J  X_\J }^{-1}
\left(  \hS_{X_\J  Y} - \lambda_n \Diag ( d_j/ \| \hat{w}_j \|) \hat{w}_\J 
\right),
$$
and thus
\BEAS
 \hS_{X_i Y } -  \hS_{X_i X_\J }  \hat{w}_\J  & = & 
 (  \hS_{X_i Y } 
  -  \hS_{X_i X_\J }  \hS_{X_\J  X_\J }^{-1} \hS_{X_\J  Y}  )
+ \lambda_n \hS_{X_i X_\J }  \hS_{X_\J  X_\J }^{-1} \Diag ( d_j/ \| \hat{w}_j \|) \hat{w}_\J  \\
& = & A_n + B_n.
\EEAS
The second term $B_n$ in the last expression (divided by $\lambda_n$) converges to 
$$ v = \S_{X_i X_\J } \S_{X_\J  X_\J  }^{-1}   \Diag ( d_j/ \| \wj \|) \wJ \in \rb^{p_j},$$
because $\hat{w}$ is assumed to converge in probability to $\w$ and empirical covariance
matrices converge to  population covariance matrices. By assumption $\|v\| > d_i$, which implies
that the probability $\P\left\{  \left(\frac{v}{\|v\|}\right)^\top ( B_n/\lambda_n ) \geqslant (d_i + \| v \|)/2) \right\}$ converges to one.

The first term  is equal to (with $\varepsilon_k = y_k - \w^\top x_k - \b_k$ and $\bar{epsilon}
=\frac{1}{n} \sum_{k=1}^n \varepsilon_k$):
\BEAS
A_n & = & \hS_{X_i Y } 
  -  \hS_{X_i X_\J }  \hS_{X_\J  X_\J }^{-1} \hS_{X_\J  Y} \\
  & = &  \hS_{X_i X_\J } \w_\J
  -  \hS_{X_i X_\J }  \hS_{X_\J  X_\J }^{-1} \hS_{X_\J  X_\J} \w_\J
 +
  \hS_{X_i \varepsilon } 
  -  \hS_{X_i X_\J }  \hS_{X_\J  X_\J }^{-1} \hS_{X_\J  \varepsilon} \\
  & = &   \hS_{X_i \varepsilon } 
  -  \hS_{X_i X_\J }  \hS_{X_\J  X_\J }^{-1} \hS_{X_\J  \varepsilon} \\
  & = & \hS_{X_i \varepsilon } 
  -  \S_{X_i X_\J }  \S_{X_\J  X_\J }^{-1} \hS_{X_\J  \varepsilon} + o_p(n^{-1/2}) \\
  & = & 
  \frac{1}{n} \sum_{k=1}^n ( \varepsilon_k - \bar{\varepsilon})\left( x_{ki} -  \S_{X_i X_\J }  \S_{X_\J  X_\J }^{-1}
  x_{k\J} \right) + o_p(n^{-1/2}) =  C_n + o_p(n^{-1/2}).
\EEAS

The random variable $C_n$ is a is a U-statistic with square integrable kernel obtained from i.i.d. random vectors; it is thus asymptotically normal~\citep{VanDerVaart}.
We thus simply need to compute the mean and the variance of $C_n$.
We have $ \E C_n = 0 $ because $\E (X\varepsilon) = \Sigma_{X\varepsilon}=0$. We let denote
$D_{k} = x_{ki} -  \S_{X_i X_\J }  \S_{X_\J  X_\J }^{-1}
  x_{k\J} -  \frac{1}{n} \sum_{k=1}^n  x_{ki} -  \S_{X_i X_\J }  \S_{X_\J  X_\J }^{-1}
  x_{k\J} $. We have:
\BEAS
\var( C_n )  & = & \E C_n^2 = \E( \E (C_n^2 | \X ) ) \\
& = & \E \left[
\frac{1}{n^2} \sum_{k=1}^n E( \varepsilon_k^2| \X)    D_k D_k^\top  \right] \\
& \succcurlyeq & \E \left[
\frac{1}{n^2} \sum_{k=1}^n \sm     D_k D_k^\top  \right]  \\
  & = & \frac{1}{n} \sm  \E  \left( \hS_{X_i X_i} - \S_{X_i X_\J }  \S_{X_\J  X_\J }^{-1}
 \hS_{X_\J X_i} \right)  \\ 
& = & \frac{n-1}{n^2} \sm  \left( \S_{X_i X_i} - \S_{X_i X_\J }  \S_{X_\J  X_\J }^{-1}
 \S_{X_\J X_i}  \right),
\EEAS
where $ M \succcurlyeq N$ denotes the partial order between symmetric matrices (i.e., equivalent to 
$M - N$ positive semidefinite).

Thus $n^{1/2} C_n$ is  
asymptotically normal 
with mean $0$ and covariance matrix larger than
$\sm  \S_{X_i | X_\J } = \sm \times ( \S_{X_i X_i}
- \S_{X_i X_\J} \S_{X_\J X_\J}^{-1} \S_{X_\J X_i} )$ which is positive definite (because this is the conditional
covariance of $X_i$ given $X_\J$ and $\S_{XX}$ is assumed invertible).
Therefore
$\P( n^{1/2} v^\top A_n  > 0)$ converges
  to  a constant $a \in (0,1)$, which implies that 
  $\P \left\{  \frac{v}{\|v\|}^\top ( A_n + B_n ) / \lambda_n  \geqslant (d_i + \| v \|)/2 \right\}$ is asymptotically bounded below by $a$.
  Thus, since $ \left\|
  ( A_n + B_n ) / \lambda_n  \right\| \geqslant \frac{v}{\|v\|}^\top ( A_n + B_n ) / \lambda_n  \geqslant (d_i + \| v \|)/2 > d_i$
  implies that $\hat{w}$ is not optimal, we get a contradiction, which concludes the proof.

\subsection{Proof of Theorem~\ref{theo:theo1-refined}}

\label{app:theo1-refined}

We first prove the following refinement of Lemma~\ref{lemma:tilde}:
\begin{lemma}
\label{lemma:tilde-refined}
Assume \hypreff{var}{model}.
Let $\tilde{w}_\J $ any minimizer of
$$\frac{1}{2n} \| \Y - \X_\J  w_\J  \|^2 + \lambda_n  \sum_{j \in \J } d_j \| w_j \|
=
\frac{1}{2} \hS_{YY} -  \hS_{YX_\J } w_\J  + \frac{1}{2} w_\J ^\top
 \hS_{X_\J X_\J  } w_\J  + \lambda_n  \sum_{j \in  \J } d_j \| w_j \|.
$$
If $\lambda_n  \to 0$ and $\lambda_n n^{1/2} \to \infty$, 
then $\frac{1}{\lambda_n} ( \tilde{w}_\J - \wJ) $ converges in probability to
$$\Delta =  - \S_{X_\J X_\J  }^{-1} \Diag( d_j / \|\w_j\|) \w_\J. $$
\end{lemma}
\begin{proof}
We follow~\citet{fu} and write $ \tilde{w}_\J = \w_\J + \lambda_n \tilde{\Delta}$. The vector $\tilde{\Delta}$ is the minimizer of the following
function:
\BEAS
F(\Delta)\!\!\!\! & = \!\!\!\!& 
- \hS_{YX_\J } ( \w_\J + \lambda_n  {\Delta} )  + \frac{1}{2} ( \w_\J + \lambda_n  {\Delta}) ^\top
 \hS_{X_\J X_\J  } ( \w_\J + \lambda_n  {\Delta} )  + \lambda_n  \sum_{j \in  \J } d_j \| \w_j + \lambda_n  {\Delta}_j \| \\
 & =\!\!\!\! & -\lambda_n \hS_{YX_\J } \Delta    + \frac{\lambda_n^2}{2}  {\Delta} ^\top
 \hS_{X_\J X_\J  }{\Delta} + \lambda_n \w_\J^\top \hS_{X_\J X_\J  } {\Delta}
 + \lambda_n  \sum_{j \in  \J } d_j  \left( \| \w_j + \lambda_n  {\Delta}_j \| - \| \w_j\| \right) + \mbox{ cst} \\
  & =\!\!\!\! & - \lambda_n \hS_{\varepsilon X_\J } \Delta  + \frac{\lambda_n^2}{2}  {\Delta} ^\top
 \hS_{X_\J X_\J  }{\Delta}  
 + \lambda_n  \sum_{j \in  \J } d_j  \left( \| \w_j + \lambda_n  {\Delta}_j \| - \| \w_j\| \right) + \mbox{ cst},
\EEAS
by using $\hS_{Y X_\J } = \w_\J^\top \hS_{X_\J  X_\J} +  \hS_{\varepsilon  X_\J}  $.
The first term  is $O_p(n^{-1/2} \lambda_n) = o_p(\lambda_n^2)$, while the last ones are equal
to $ \| \w_j + \lambda_n  {\Delta}_j \| - \| \w_j\| = \lambda_n \left(
\frac{ \w_j } {\| \w_j \|} \right)^\top \Delta_j + o_p(\lambda_n)$. Thus,
$$F(\Delta)/\lambda_n^2 =  \frac{1}{2}  {\Delta} ^\top
 \S_{X_\J X_\J  }{\Delta}  
 +  \sum_{j \in  \J } \frac{ d_j \w_j } {\| \w_j \|}^\top \Delta_j   + o_p(1).$$
 By Lemma~\ref{lemma:tilde}, $\hat{w}_\J$ is $O_p(1)$ and the limiting function has an unique minimum; 
 standard results in M-estimation~\citep{VanDerVaart} shows that $\tilde{\Delta}$ converges in 
probability to the minimum of the last expression which is exactly
$\Delta = - \S_{X_\J X_\J  }^{-1} \Diag( d_j / \|\w_j\|) \w_\J$.
\end{proof}

\vspace*{1cm}

We now turn to the proof of Theorem~\ref{theo:theo1-refined}.
We follow the proof of Theorem~\ref{theo:theo1}. Given $\tilde{w}$ defined through
Lemma~\ref{lemma:tilde} and~\ref{lemma:tilde-refined}, we need to satisfy optimality condition~(\ref{eq:opt1}) for all $i \in \J^c$,
with probability tending to one. For all
those $i$ such that
$\frac{1}{d_i} \left\|  \S_{X_i X_\J  } \S_{X_\J  X_\J }^{-1}
\Diag  ( d_j/ \| \w_j \|) \wJ  \right\| < 1$, then we know from  \myapp{theo1}, that the optimality condition is indeed
satisfied with probability tending to one. We now focus on those $i$ such that  $\frac{1}{d_i} \left\|  \S_{X_i X_\J  } \S_{X_\J  X_\J }^{-1}
\Diag  ( d_j/ \| \w_j \|) \wJ  \right\| =1$, and for which we have the condition in \eq{new}.
From \eq{pp2} and the few arguments that follow, we get that for all $i\in \J^c$,
\BEQ
\label{eq:qq1}
 \hS_{X_{i} Y} -  \hS_{X_{i} X_\J } \tilde{w}_\J  = 
  \lambda_n \S_{X_{i} X_\J  } \S_{X_\J  X_\J  }^{-1} \Diag ( d_j/ \| \tilde{w}_j \|) \tilde{w}_\J + O_p(n^{-1/2})
\EEQ
Moreover, we have from Lemma~\ref{lemma:tilde-refined} and standard differential calculus, i.e., the gradient and the Hessian
of the function $v  \in \rb^q \mapsto \|v\| \in \rb$ are $v/\|v\|$ and $\frac{1}{\|v\|}\left( \idm_q - \frac{ vv^\top}{v^\top v} \right)$:
\BEQ
\label{eq:qq2}
\frac{\tilde{w}_j}{\|\tilde{w}_j\|}
= \frac{\w_j}{\|\w_j\|} + \frac{\lambda_n}{\|\w_j\|} \left( \idm_{p_j} - \frac{ \w_j \w_j^\top}{ \w_j^\top \w_j}
\right) \Delta_j + o_p(\lambda_n).
\EEQ
From \eq{qq1} and \eq{qq2}, we get:
 \BEAS
  \frac{1}{\lambda_n} (\hS_{X_{i} Y} -  \hS_{X_{i} X_\J } \tilde{w}_\J )
 & =  & O_p( n^{-1/2} \lambda_n^{-1}) + \S_{X_i X_\J  } \S_{X_\J  X_\J  }^{-1} \\
   & &  \hspace*{-4cm}
    \left\{  \Diag ( d_j/ \| \w_j \|) \w_\J  + \lambda_n \S_{X_{i} X_\J  } \S_{X_\J  X_\J  }^{-1} \Diag \left[ d_j / \|\w_j\| \left(
  \idm_{p_j} - \frac{ \w_j \w_j^\top}{ \w_j^\top \w_j}
\right) \right]  \Delta + o_p(\lambda_n)  \right\} \\
& = & A + \lambda_n B +  o_p(\lambda_n)  + O_p( n^{-1/2} \lambda_n^{-1}) .
\EEAS
 Since $\lambda_n \gg n^{-1/4}$, we have
   $O_p( n^{-1/2} \lambda_n^{-1}) = o_p(\lambda_n)$. Thus,
 since we assumed that $\| A \| = \|  \S_{X_i X_\J  } \S_{X_\J  X_\J  }^{-1} \Diag ( d_j/ \| \w_j \|) \w_\J \| = d_i$, we have:
 \BEAS
 \left\| \frac{1}{\lambda_n} (\hS_{X_{i} Y} -  \hS_{X_{i} X_\J } \tilde{w}_\J ) \right\|^2
 & =  & \| A\|^2 + 2\lambda_n A^\top B + o_p(\lambda_n)
   d_i^2 + o_p(   \lambda_n) \\ 
   & = & d_i^2 + o_p(\lambda_n) \\
   & &  \hspace*{-2cm}
 - 2 \lambda_n 
  \Delta^\top \S_{ X_\J X_i  }
 \S_{X_{i} X_\J  } \S_{X_\J  X_\J  }^{-1} \Diag \left( d_j / \|\w_j\| ( \idm_{p_j} - \frac{ \w_j \w_j^\top}{ \w_j^\top \w_j}
) \right)  \Delta,
\EEAS
(note that we have $A = -\S_{X_i X_\J} \Delta$) which is asymptotically strictly smaller than $d_i^2$ if \eq{new} is satisfied, which proves optimality and concludes the proof.

\subsection{Proof of Proposition~\ref{prop:probab}}
\label{app:probab}
As in the proof of Theorem~\ref{theo:theo1} in \myapp{theo1}, we consider the estimate $\tilde{w}$ built from the reduced problem by constraining $w_{\J^c}=0$. We consider the following event:
$$E_1 = \{ \hS_{XX} \mbox{ invertible and } \forall j \in \J, \ \tilde{w}_j \neq 0 \}.$$ 
This event has a probability converging to one. Moreover, if $E_1$ is true, then the group Lasso estimate has the correct sparsity pattern if and only if for all $i \in \J^c$,
$$\left\| \hS_{X_i X_\J } (\tilde{w}_\J - \w_\J)  -  \hS_{X_i \varepsilon} \right\| \leqslant \lambda_n d_i = \lambda_0 n^{-1/2} d_i.
$$
Moreover we have by definition of $\tilde{w}_\J$: $ \hS_{X_\J X_\J } (\tilde{w}_\J - \w_\J)  -  \hS_{X_\J \varepsilon} = - \lambda_n 
 \Diag ( d_j/ \| \tilde{w}_j \|) \tilde{w}_\J $, and thus, we get:
 \BEAS
& &  \hS_{X_i X_\J } (\tilde{w}_\J - \w_\J)  -  \hS_{X_i \varepsilon}  \\
& \!\!\!\!=\!\! \!\!&  \hS_{X_i X_\J } \hS_{X_\J X_\J }^{-1} \hS_{X_\J \varepsilon } - 
 \hS_{X_i \varepsilon }  -  \lambda_0 n^{-1/2} \hS_{X_i X_\J} \hS_{X_\J X_\J }^{-1}
 \Diag ( d_j/ \| \tilde{w}_j \|) \tilde{w}_\J  \\
& \!\!\!\!= \!\!\!\! & 
\S_{X_i X_\J } \S_{X_\J X_\J }^{-1} \hS_{X_\J \varepsilon } - 
 \hS_{X_i \varepsilon }  -   \lambda_0 n^{-1/2} 
 \S_{X_i X_\J} \S_{X_\J X_\J }^{-1}
 \Diag ( d_j/ \| \w_j \|) \w_\J + O_p(n^{-1})
 \EEAS
 The random vector $\S_{X  \varepsilon } \in \rb^p$ is a multivariate U-statistic with square integrable kernel obtained from i.i.d. random vectors; it is thus asymptotically normal~\citep{VanDerVaart} and we simply need to
 compute its mean and variance. The mean is zero, and the variance is $\frac{n-1}{n^2}\sigma^2 \S_{XX} = n^{-1}\sigma^2 \S_{XX} + o(n^{-1})$. This implies that the random vector $s$ of size ${\rm Card}(\J^c)$ defined by $$s_i = n^{1/2} \| \hS_{X_i X_\J } (\tilde{w}_\J - \w_\J)  -  \hS_{X_i \varepsilon}  \|,$$ is equal to 
 \BEAS
s_i &  = &  \left\| 
 \sigma\S_{X_i X_\J } \S_{X_\J X_\J }^{-1} u_\J - 
\sigma u_i  -   \lambda_0 \S_{X_i X_\J} \S_{X_\J X_\J }^{-1}
 \Diag ( d_j/ \| \w_j \|) \w_\J \right\| + O_p(n^{-1/2})  \\
 & = &  f_i(u) + O_p(n^{-1/2}),
 \EEAS
  where $u = \sigma^{-1} n^{-1/2}\hS_{X  \varepsilon }$ and $f_i$ are deterministic continuous functions. The vector $f(u)$ converges in distribution  to $f(v)$ where $v$ is normally distributed with mean zero and
 covariance matrix $  \S_{XX}$. By Slutsky's lemma~\citep{VanDerVaart}, this implies that the random vector $s$ has the same limiting distribution. Thus, the probability $\P(\max_{i \in \J^c} s_i/d_i \leqslant \lambda_0)$ converges to  $$\P
 \left( \max_{i \in \J^c} \frac{1}{d_i}\left\| \sigma ( \S_{X_i X_\J } \S_{X_\J X_\J }^{-1} v_\J - 
 \ v_i)  -   \lambda_0 \S_{X_i X_\J} \S_{X_\J X_\J }^{-1}
 \Diag ( d_j/ \| \w_j \|) \w_\J \right\| \leqslant \lambda_0\right).$$
 Under the event $E_1$ which has probability tending to one, we have correct pattern selection if and only if $\max_{i \in \J^c} s_i/d_i \leqslant \lambda_0$, which leads to 
  $$\P
 \left( \max_{i \in \J^c} \frac{1}{d_i}\left\|\sigma  {t}_i -  \lambda_0 \S_{X_i X_\J} \S_{X_\J X_\J }^{-1}
 \Diag ( d_j/ \| \w_j \|) \w_\J \right\| \leqslant \lambda_0\right),$$
 where $t_i  =  \S_{X_i X_\J } \S_{X_\J X_\J }^{-1} v_\J - 
 \ v_i
 $. The vector $t $ is normally distributed and a short calculation shows that its  covariance matrix is equal to 
 $\S_{X_{\J^c} X_{\J^c} | X_\J}$, which concludes the proof.

\section{Detailed Proofs for the Nonparametric Formulation}
\label{app:mklproofs}
We first prove lemmas that will be useful for further proofs, and then prove the consistency results for the non parametric case.

\subsection{Useful Lemmas on Empirical Covariance Operators}
\label{app:cov}

We first have the following lemma, proved by~\citet{kenji}, which states that the empirical covariance estimator converges in probability
at rate $O_p(n^{-1/2})$ to the population covariance operators:
\begin{lemma}
\label{lemma:cov1}
Assume \hypref{rkhs} and \hypref{model-cov}. Then $ \|\hS_{XX} - \S_{XX} \|_\F= O_p(n^{-1/2})$ (for the operator norm),
$\|\hS_{XY} - \S_{XY} \|_\F = O_p(n^{-1/2})$ and $\|\hS_{X\varepsilon} \|_\F = O_p(n^{-1/2})$.
\end{lemma}
The following lemma is useful in several proofs:
\begin{lemma}
\label{lemma:cov2}
Assume \hypref{rkhs}. Then $ \left \| \left(\hS_{XX} + \mu_n \idm\right)^{-1} \S_{XX}  -
\left(\S_{XX} + \mu_n \idm\right)^{-1} \S_{XX}  \right\|_\F = O_p(n^{-1/2} \mu_n^{-1} ), $ and
$ \left\| \left(\hS_{XX} + \mu_n \idm\right)^{-1} \hS_{XX}  -
\left(\S_{XX}+ \mu_n \idm\right)^{-1} \S_{XX}  \right\|_\F = O_p(n^{-1/2} \mu_n^{-1} ) $.
\end{lemma}
\begin{proof}
We have:
\BEAS 
& & \left(\hS_{XX} + \mu_n \idm\right)^{-1} \S_{XX}  -
\left(\S_{XX} + \mu_n \idm\right)^{-1} \S_{XX}  \\
& = &   \left(\hS_{XX} + \mu_n \idm\right)^{-1}  ( \S_{XX} - \hS_{XX} ) \left(\S_{XX} + \mu_n \idm\right)^{-1} \S_{XX}
\EEAS
This is the product of operators whose norms are respectively upper bounded by $\mu_n^{-1}$, $O_p(n^{-1/2})$ and 1, which leads to the
first inequality (we use $\| A B \|_\F \leqslant \|A\|_\F \| B\|_\F$). The second inequality follows along similar lines.
\end{proof}
Note that the two  previous lemma also hold for any suboperator of $\S_{XX}$, i.e., for $\S_{X_\J X_\J}$, or $\S_{X_i X_i}$.

\begin{lemma}
\label{lemma:cov4}
Assume \hypref{rkhs}, \hypref{compact} and \hypref{range}.  There exists $\mathbf{h}_\J \in \F_\J$ such that
$\f_\J = \S_{X_\J X_\J}^{1/2} \mathbf{h}_\J$.
\end{lemma}
\begin{proof}
The range condition implies that 
\BEAS 
\f_\J & =  & \Diag(\S_{X_j X_j}^{1/2}) \g_\J
 = \Diag(\S_{X_j X_j}^{1/2}) C_{X_\J X_\J}^{1/2} C_{X_\J X_\J}^{-1/2}  \g_\J
\EEAS
(because $C_{XX}$ is invertible).
The result follows from the identity
$$\S_{X_\J X_\J} = \Diag(\S_{X_j X_j}^{1/2}) C_{X_\J X_\J}^{1/2} (\Diag(\S_{X_j X_j}^{1/2}) C_{X_\J X_\J}^{1/2})^\ast$$ and
the fact that if $\S_{X_\J X_\J} = UU^\ast$ and $f=U\alpha$ then there exists $\beta$ such that
$f = \S_{X_\J X_\J}^{1/2} \beta$~\citep{baker}.\footnote{The adjoint operator $V^\ast$ of $V: \Fi \to \FJ$ is so that for all $f \in \Fi$ and $g \in \FJ$,
$\langle f, Vg \rangle_\Fi = \langle V^\ast f, g \rangle_\FJ $~\citep{brezis80analyse}.}
\end{proof}

\subsection{Proof of Theorem~\ref{theo:theo1-cov}}
\label{app:theo1-cov}

We now extend Lemma~\ref{lemma:tilde} to covariance operators, which requires to use the alternative formulation and a slower rate of
decrease for the regularization parameter:
\begin{lemma}
\label{lemma:tilde-cov}
Let $\tilde{f}_\J $ be any  minimizer of
$$\frac{1}{2} \hS_{YY} -  \langle\hS_{X_\J Y }, f_\J\rangle_\FJ  + \frac{1}{2} \langle f_\J ,
 \hS_{X_\J  X_\J } f_\J  \rangle_\FJ   + \frac{\mu_n}{2}  \left( \sum_{j \in \J } d_j \| f_j \|_\Fj \right)^2.
$$
If $\mu_n  \to 0$ and $\mu_n n^{1/2}  \to +\infty$, then $\| \tilde{f}_\J   - \fJ \|_\FJ$ converges to zero in probability. Moreover
for any $\eta_n$ such that $\eta_n \gg \mu_n^{1/2} + \mu_n^{-1} n^{-1/2}$ then $\| \tilde{f}_\J   - \fJ \|_\FJ = O_p(\eta_n)$.
\end{lemma}
\begin{proof}
Note that from Cauchy-Schwartz inequality, we have:
\BEAS
 \left( \sum_{j \in \J } d_j \| f_j \|_\Fj \right)^2
 & = & \left( \sum_{j \in \J }  d_j^{1/2}\|\f_j\|_\Fj^{1/2} \times \frac{d_j^{1/2} \| f_j \|_\Fj}{\|\f_j\|_\Fj^{1/2}}  \right)^2
\\
  & \leqslant& 
  \left( \sum_{j \in \J } d_j \| \fj \|_\Fj \right) \sum_{j \in \J} \frac{ d_j \|f_j\|_\Fj^2}{\|\fj\|_\Fj},
\EEAS
with equality if  and only if there exists $\alpha>0$ such that $\| f_j \|_\Fj  =  \alpha \| \fj \|_\Fj$ for all $j \in \J$.
We consider the unique minimizer $\bar{f}_\J $ of the following cost function, built by replacing the regularization by its
upperbound,
$$F(f_\J ) = \frac{1}{2} \hS_{YY} -  \langle\hS_{X_\J Y }, f_\J\rangle_\FJ   + \frac{1}{2} \langle f_\J ,
 \hS_{X_\J  X_\J } f_\J  \rangle_\FJ  + \frac{\mu_n}{2}  
 \left( \sum_{j \in \J } d_j \| \fj \|_\Fj \right) \sum_{j \in \J} \frac{d_j \| f_j\|_\Fj^2}{\|\fj\|_\Fj}.
$$
Since it is a regularized least-square  problem, we have (with $\varepsilon = Y - \sum_{j \in \J} \f_j(X) - \b$):
$$\bar{f}_\J  = \left(  \hS_{X_\J  X_\J  } + \mu_n D \right)^{-1} \left(
\hS_{X_\J  X_\J } \fJ  +  \hS_{X_\J  \varepsilon} \right), $$
where $ D =   \left( \sum_{j \in \J } d_j \| \fj \| \right)
\Diag( d_j / \| \fj \| )$. Note that $D$ is upperbounded and lowerbounded, as an
auto-adjoint operator,
 by \emph{strictly positive} constants times the
identity operator (with probability tending to one), i.e., $ D_{\max} \idm_{\FJ} \succcurlyeq D \succcurlyeq D_{\min} \idm_{\FJ}$ with $D_{\min}, D_{\max} >0$.
We now prove that $\bar{f}_\J  - \fJ$ is converging to zero in probability. 
We have:
\BEQ
\label{eq:s1}
 \left(  \hS_{X_\J  X_\J  } + \mu_n D \right)^{-1}  
\hS_{X_\J  \varepsilon} = O_p( n^{-1/2} \mu_n^{-1}), 
\EEQ
because of Lemma~\ref{lemma:cov1} and $ \left\| \left(
\hS_{X_\J  X_\J  } + \mu_n D \right)^{-1} \right\|_\FJ \leqslant D_{\min}^{-1} \mu_n^{-1}$.
Moreover, similarly, we have
\BEQ
\label{eq:s2}
\left(  \hS_{X_\J  X_\J  } + \mu_nD \right)^{-1} 
\hS_{X_\J  X_\J } f_\J  - \left(  \hS_{X_\J  X_\J  } + \mu_n D \right)^{-1} 
\S_{X_\J  X_\J } \fJ = O_p( n^{-1/2} \mu_n^{-1}).
\EEQ
Besides, by Lemma~\ref{lemma:cov2},
\BEQ
\label{eq:s3}
\left(  \hS_{X_\J  X_\J  } + \mu_nD \right)^{-1} 
\S_{X_\J  X_\J } f_\J  
-
\left(  \S_{X_\J  X_\J  } + \mu_nD \right)^{-1} 
\S_{X_\J  X_\J } f_\J  = O_p( n^{-1/2} \mu_n^{-1}).
\EEQ
Thus $ \bar{f}_\J  - \fJ = V + O_p( n^{-1/2} \mu_n^{-1})$, where
\BEAS
V & =  &  \left[
\left(   \S_{X_\J  X_\J } + \mu_n  D  \right)^{-1} \S_{X_\J  X_\J } -  \idm  \right] \fJ 
= - \left(   \S_{X_\J  X_\J } + \mu_n  D  \right)^{-1} \mu_n D \fJ.
\EEAS
We have
\BEAS
\| V\|_\FJ^2 & = & \mu_n^2  \langle \fJ, D \left(   \S_{X_\J  X_\J } + \mu_n  D  \right)^{-2} D \f_\J \rangle_\FJ \\
& \leqslant &  D_{\max}^2 \mu_n^2 
\langle  \fJ , \left(   \S_{X_\J  X_\J } + \mu_n  D_{\min} \idm  \right)^{-2}   \f_\J  \rangle_\FJ \\
& \leqslant &  D_{\max}^2 \mu_n
\langle  \fJ ,  \left(   \S_{X_\J  X_\J } + \mu_n  D_{\min} \idm  \right)^{-1}   \f_\J  \rangle_\FJ \\
& \leqslant & D_{\max}^2 \mu_n
 \langle \mathbf{h}_\J , \S_{X_\J  X_\J } \left(   \S_{X_\J  X_\J } + \mu_n  D_{\min} \idm  \right)^{-1}   \mathbf{h}_\J  \rangle_\FJ
 \mbox{ by Lemma~\ref{lemma:cov4}}, \\
 & \leqslant & D_{\max}^2 \mu_n
 \| \mathbf{h}_\J\|_\FJ^2.
\EEAS
Finally we obtain  $\|   \bar{f}_\J  - \fJ  \|_\FJ = O_p(\mu_n^{1/2} + n^{-1/2} \mu_n^{-1} )$.

\vspace*{1cm}

We now consider the cost function defining $\tilde{f}_\J$:
$$ F_n(f_\J ) = \frac{1}{2} \hS_{YY} -  \langle\hS_{X_\J Y }, f_\J\rangle_\FJ   + \frac{1}{2} \langle f_\J ,
 \hS_{X_\J  X_\J } f_\J  \rangle_\FJ   + \frac{\mu_n}{2}  \left( \sum_{j \in \J } d_j \| f_j \|_\Fj \right)^2.
$$
We have (note that although we seem to take infinite dimensional derivatives, everything can be done in the finite subspace
spanned by the data):
\BEAS
F_n(f_\J )  - F(f_\J ) & = & \frac{\mu_n}{2}  \left[
 \left( \sum_{j \in \J } d_j \| f_j \|_\Fj \right)^2  
-  \left( \sum_{j \in \J } d_j \| \fj \|_\Fj \right) \sum_{j\in \J} \frac{d_j \| f_j\|_\Fj^2}{\|\fj\|_\Fj} \right], \\
\nabla_{f_i}F_n(f_\J )  -  \nabla_{f_i} F(f_\J ) & = & 
\mu_n \left[
 \left( \sum_{j \in \J } d_j \| f_j \|_\Fj \right)  \frac{d_i f_i}{\|f_i\|_\Fj}
-  \left( \sum_{j \in \J } d_j \| \fj \|_\Fj \right)  \frac{d_i f_i}{\|\f_i\|_\Fj} \right]. \\
\EEAS
Since the right hand side of the previous equation corresponds
to a continuously differentiable function of $f_\J$ around $\fJ$ (with upper-bounded derivatives around $\fJ$), we have:
$$ \| \nabla_{f_i} F_n(\bar{f}_\J ) - 0 \|_\Fi \leqslant C \mu_n \| \fJ-\bar{f}_\J  \|_\FJ
=  \mu_n
O_p( \mu_n^{1/2} +  n^{-1/2} \mu_n^{-1} ) .$$
for some constant $C>0$.
Moreover, on the ball of center $\bar{f}_\J $ and radius $\eta_n$ such that
$\eta_n \gg \mu_n^{1/2}+ \mu_n^{-1} n^{-1/2} $ (to make sure that it asymptotically contains ${\f}_\J$, which
implies that on the ball each $f_j$, $j \in \J$ are bounded away from zero),
 and $ \eta_n \ll 1 $ (so that we get consistency), we have a 
lower bound on the second derivative of
$\left( \sum_{j \in \J} d_j \| f_j \|_\Fj \right)$. Thus
for any element of the ball,
$$  F_n( {f}_\J ) \geqslant F_n(\bar{f}_\J ) +  \langle\nabla_{f_\J }   F_n(\bar{f}_\J ),( 
{f}_\J -\bar{f}_\J ) \rangle_\FJ + C' \mu_n  \| {f}_\J -\bar{f}_\J \|_\FJ^2, $$
where $C'>0$ is a constant.
This implies that the value of 
$F_n(f_\J )$ on the edge of the ball is larger than
 $$F_n(\bar{f}_\J ) + \eta_n \mu_n O_p(\mu_n^{1/2} + n^{-1/2} \mu_n^{-1} )
+ C' \eta_n^2 \mu_n,
$$

Thus if $\eta_n^2 \mu_n \gg \eta_n  \mu_n^{3/2} $ and 
$\eta_n^2 \mu_n \gg  n^{-1/2} \eta_n  $, then we must have all  minima inside the 
ball of radius $\eta_n$ (because with probability tending to one, the value on the edge is greater than one value inside and 
the function is convex) which implies that the global
minimum of $F_n$ is at most $\eta_n$ away from $\bar{f}_\J $ and thus since $\bar{f}_\J $
 is $O(\mu_n^{1/2})$ away from $\fJ$, we have the consistency
if
$$
\eta_n \ll    1 
\mbox{ and }
\eta_n \gg    \mu_n^{1/2} 
+  n^{-1/2} \mu_n^{-1},
$$
which concludes the proof of the lemma.
\end{proof}

We now prove Theorem~\ref{theo:theo1-cov}.
Let $\tilde{f}_\J $ be defined as in Lemma~\ref{lemma:tilde}. We extend it by zeros on $\J^c$. We already know
the squared norm consistency by Lemma~\ref{lemma:tilde}. Since by Proposition~\ref{prop:opt-alt-dual-unique}, the solution is unique with probability tending to one, we need to prove that with probability
tending to one $\tilde{f}$ is optimal for problem in \eq{problem-alt-cov}.
We have by the first optimality condition for $\tilde{f}_\J$:
$$ \hS_{X_\J  Y}  - \hS_{X_\J  X_\J } \tilde{f}_\J   = 
\mu_n \|\tilde{f}\|_d \Diag  ( d_j/ \| \tilde{f}_j \|) \tilde{f}_\J, $$
 where
we use the notation
$\|f\|_d = \sum_{j=1}^m d_j \|f_j\|_\Fj$ (note the difference with
$\| f\|_\F = ( \sum_{j=1}^m \| f_j\|_\Fj^2)^{1/2}$). 
We thus have by solving for $\tilde{f}_\J$ and using $\hS_{X_\J Y} = \hS_{X_\J  X_\J  } \fJ
+\hS_{X_\J  \varepsilon}$:
$$\tilde{f}_\J  =
\left(  \hS_{X_\J  X_\J  } + 
\mu_n D_n \right)^{-1} \left( \hS_{X_\J  X_\J  }\fJ
+\hS_{X_\J  \varepsilon} \right),
$$
with the notation
$D_n = \|\tilde{f}\|_d \Diag  ( d_j/ \| \tilde{f}_j \|_\Fj)
$.
We can now put that back into
$  \hS_{X_{\J^c} Y} -  \hS_{X_{\J^c} X_\J } \tilde{f}_\J$ and show that this will have small enough norm
with
probability tending to one.
We have for all $i \in \J^c$:
\BEA
\nonumber
\hS_{X_{i} Y} -  \hS_{X_{i} X_\J } \tilde{f}_\J 
& = & 
\hS_{X_{i} Y}
-  \hS_{X_{i} X_\J }
\left(  \hS_{X_\J  X_\J  } + 
\mu_n  D_n \right)^{-1} \left( \hS_{X_\J  X_\J } \fJ
+\hS_{X_\J  \varepsilon} \right) \\
\nonumber
& = & 
-  \hS_{X_{i} X_\J }
\left(  \hS_{X_\J  X_\J  } + 
\mu_n  D_n \right)^{-1}  \hS_{X_\J  X_\J } \fJ
 \\
\nonumber
 & & + 
 \hS_{X_{i} Y}
-  \hS_{X_{i} X_\J }
\left(  \hS_{X_\J  X_\J  } + 
\mu_n  D_n \right)^{-1} \hS_{X_\J  \varepsilon} \\
\nonumber
& = & 
-  \hS_{X_{i} X_\J } \fJ + 
\hS_{X_{i} X_\J }
\left(  \hS_{X_\J  X_\J  } + 
\mu_n  D_n \right)^{-1}
\mu_n  D_n   \fJ
 \\
\nonumber
 & & + 
 \hS_{X_{i} Y}
-  \hS_{X_{i} X_\J }
\left(  \hS_{X_\J  X_\J  } + 
\mu_n  D_n \right)^{-1} \hS_{X_\J  \varepsilon}
\\
\nonumber
& = & 
   \hS_{X_{i} X_\J }
\left(  \hS_{X_\J  X_\J  } + 
\mu_n  D_n \right)^{-1} \mu_n  D_n   \fJ
 \\
 & & \label{eq:alpha} + 
 \hS_{X_{i} \varepsilon}
-  \hS_{X_{i} X_\J }
\left(  \hS_{X_\J  X_\J  } + 
\mu_n  D_n \right)^{-1} \hS_{X_\J  \varepsilon} \\
\nonumber
& = & A_n + B_n.
 \EEA
The first term $A_n$ (divided by $\mu_n  $) is equal to
$$
\frac{A_n}{\mu_n } = 
\hS_{X_{i} X_\J }
\left(  \hS_{X_\J  X_\J  } + 
\mu_n  D_n \right)^{-1} D_n   \fJ.
$$
We can replace $\hS_{X_{i} X_\J } $ in $\frac{A_n}{\mu_n }$
 by $\S_{X_{i} X_\J }$ at cost $O_p(n^{-1/2} \mu_n^{-1/2})$ because
$\langle \fJ ,  \S_{X_\J X_\J}^{-1} \fJ \rangle_\FJ < \infty$ (by Lemma~\ref{lemma:cov4}).
Also, we can replace $\hS_{X_\J X_\J } $ in $\frac{A_n}{\mu_n }$ by $\S_{X_\J X_\J }$ at cost $O_p(n^{-1/2} \mu_n^{-1})$
as a consequence of Lemma~\ref{lemma:cov2}. Those two are
$o_p(1)$ by assumptions on $\mu_n$. Thus,
$$
\frac{A_n}{\mu_n } = 
\S_{X_{i} X_\J }
\left(  \S_{X_\J  X_\J  } + 
\mu_n  D_n \right)^{-1} D_n   \fJ + o_p(1).
$$
Furthermore, we let denote $D = \|\f \|_d \Diag  ( d_j/ \| \fj \|_\Fj)$. From Lemma~\ref{lemma:tilde-cov}, we know
that $D_n - D = o_p(1)$. Thus we can replace $D_n$ by $D$ at cost $o_p(1)$ to get:
$$
\frac{A_n}{\mu_n } = 
 \S_{X_{i} X_\J }
\left(   \S_{X_\J  X_\J  } + 
\mu_n  D \right)^{-1} D    \fJ + o_p(1) = C_n + o_p(1).
$$
We now show that this last deterministic term  $C_n \in \F_i$ converges to:
$$
C = \S_{X_i X_i }^{1/2} C_{X_i X_\J  }  C_{X_\J  X_\J }^{-1}
D \gJ,
$$
where, from \hypref{range}, $\forall j\in \J$, $\f_j = \S_{X_j X_j}^{1/2} \g_j$.
We have
\BEAS
C_n - C
& = & 
\S_{X_{i} X_{i} }^{1/2} 
 C_{X_{i} X_\J }  \left[
\Diag(\S_{X_j X_j }^{1/2})
\left(   \S_{X_\J  X_\J  } + 
\mu_n  D \right)^{-1}   \Diag(\S_{X_j X_j }^{1/2})  - {C}_{X_\J  X_\J }^{-1} \right] D \gJ \\
& = & 
\S_{X_{i} X_{i} }^{1/2} 
 C_{X_{i} X_\J }  K_n  D \gJ  .
\EEAS
where $K_n =\Diag(\S_{X_j X_j }^{1/2})
\left(   \S_{X_\J  X_\J  } + 
\mu_n  D \right)^{-1}   \Diag(\S_{X_j X_j }^{1/2})  - {C}_{X_\J  X_\J }^{-1} $.
In addition, we have:
\BEAS
\Diag(\S_{X_j X_j }^{1/2} ) {C}_{X_\J  X_\J }  K_n  & = & 
  {\S}_{X_\J  X_\J }  
\left(   \S_{X_\J  X_\J  } + 
\mu_n  D \right)^{-1}   \Diag(\S_{X_j X_j }^{1/2})  - \Diag(\S_{X_j X_j }^{1/2} ) \\
& =& - \mu_n D 
\left(   \S_{X_\J  X_\J  }   + 
\mu_n  D \right)^{-1}   \Diag(\S_{X_j X_j }^{1/2}) .
\EEAS
Following~\citet{kenji},
the range of the adjoint operator $ \left( \S_{X_{i} X_{i} }^{1/2} 
 C_{X_{i} X_\J } \right)^\ast = C_{ X_\J X_{i} } \S_{X_{i} X_{i} }^{1/2}   $ 
 is included in the closure of the range of $\Diag(\S_{X_j X_j })$ (which is equal to the range
 of $\S_{X_\J X_\J}$ by Lemma~\ref{lemma:cov4}). For any $v_\J \in \F_\J$
 in the intersection of two ranges, we have $v_\J = {C}_{X_\J  X_\J }  \Diag(\S_{X_j X_j }^{1/2}) u_\J$
  (note that
 ${C}_{X_\J  X_\J } $ is invertible), and thus
 \BEAS
 \langle  K_n  D \gJ  , v_\J \rangle_\FJ  & = &
   \langle  K_n  D \gJ  ,{C}_{X_\J  X_\J }  \Diag(\S_{X_j X_j }^{1/2}) u_\J \rangle_\FJ  \\
& = &  
  \langle  - \mu_n D 
\left(   \S_{X_\J  X_\J  }   + 
\mu_n  D \right)^{-1}   \Diag(\S_{X_j X_j }^{1/2})   D \gJ  , u_\J \rangle_\FJ
 \EEAS
 which is $ O_p(\mu_n^{1/2})$  
 and thus tends to zero. Since this holds for all elements in the intersection of the ranges, Lemma~9 by~\citet{kenji} implies
 that $\| C_n - C \|_\FJ$ converges to zero.

\vspace*{.25cm}

We now simply need to show that the second term  $B_n$ is dominated by $\mu_n$. 
We have: $\| \hS_{X_{i} \varepsilon}\|_\Fi =O_p(n^{-1/2})$
and $ \| \hS_{X_{i} X_\J }
\left(  \hS_{X_\J  X_\J  } + 
\mu_n  D_n \right)^{-1} \hS_{X_\J  \varepsilon} \|_\Fi \leqslant  \| \hS_{X_{i} \varepsilon}\|_\Fi $, thus,
since $\mu_n n^{1/2} \to +\infty$,
$B_n = o_p(\mu_n)$ and therefore for
for each $i \in J^c$,
$$ \frac{1}{d_i\mu_n \| \f \|_d } \left( \hS_{X_{i} Y} -  \hS_{X_{i} X_\J } \tilde{f}_\J  \right) $$
converges in probability to $\| C\|_\FJ /d_i \| \f\|_d$ which is strictly smaller than one because
\eq{condition-cov} is satisfied. Thus 
$$\P\left\{
 \frac{1}{d_i\mu_n  \| \f \|_d  } \left\|  \hS_{X_{i} Y} -  \hS_{X_{i} X_\J } \tilde{f}_\J  \right\|_\Fi \leqslant 1 \right\}$$ is tending to 1,
 which implies the theorem (using the same arguments than in the proof of 
  Theorem~\ref{theo:theo1} in \myapp{theo1}).

 \subsection{Proof of Theorem~\ref{theo:theo2-cov}}
 
\label{app:theo2-cov}

Before proving the analog of the second group Lasso theorem, we need the following additional
proposition, which states that consistency of the patterns can only be achieved if $\mu_n n^{1/2} \to \infty$
(even if chosen in a data dependent way).

\begin{proposition}
\label{prop:sqrt}
Assume \hypreff{rkhs}{range}  and
that $\J$ is not empty.
If $\hat{f}$ is converging in probability to $\f$ and $J(\hat{f}) $ converges in probability to $\J$, then
$\mu_n n^{1/2} \to \infty$ in probability.
\end{proposition}
\begin{proof}
We give a proof by contradiction, and we thus assume that there exists $M>0$ such that
$\lim\inf_{n \to \infty} \P( \mu_n n^{1/2}  < M ) > 0 $. This imposes that there exists a subsequence which
is almost surely bounded by $M$~\citep{durrett}. Thus, we can take a further subsequence which converges 
to a limit $\mu_0  \in [0,\infty) $. We now consider such a subsequence (and still use the notation
of the original sequence for simplicity).

With probability tending to one, we have the optimality condition (\ref{eq:opt2-alt-cov}):
$$
\hS_{X_\J  \varepsilon }  + \hS_{X_\J X_\J } \f_\J = 
\hS_{X_\J  Y } =  \hS_{X_\J  X_\J }  \hat{f}_\J 
+ \mu_n \| \hat{f} \|_d \Diag  ( d_j/ \| \hat{f}_j \|_\Fj ) \hat{f}_\J.
$$
If we let denote $D_n = n^{1/2}  \mu_n \| \hat{f} \|_d \Diag  ( d_j/ \| \hat{f}_j \|_\Fj ) $, we get:
$$ D_n {\f}_\J =    \left[ \hS_{X_\J  X_\J  } + D_n n^{-1/2} \right]
 n^{1/2}  \left[ \fJ - \hat{f}_\J  \right]
+  n^{1/2}  \hS_{X_\J  \varepsilon },
$$
which can be approximated as follows (we denote $D = \|\f \|_d \Diag  ( d_j/ \| {\f}_j \|_\Fj )  $):
$$ \mu_0 D {\f}_\J + o_p(1) =     \S_{X_\J  X_\J  } 
 n^{1/2}  \left[ \fJ - \hat{f}_\J  \right] + o_p(1)
+  n^{1/2}  \hS_{X_\J  \varepsilon }.
$$
We can now write for $i \in \J^c$:
\BEAS
n^{1/2} \left ( \hS_{X_{i} Y } -  \hS_{X_{i} X_\J }  \hat{f}_\J   \right)
& = & n^{1/2} \hS_{X_{i } \varepsilon } +   
 \hS_{X_{i} X_\J } n^{1/2} ( \fJ -  \hat{f}_\J  ) \\
 & = & n^{1/2} \hS_{X_{i } \varepsilon } +   
 \S_{X_{i} X_\J } n^{1/2} ( \fJ -  \hat{f}_\J  ) + o_p(1).
\EEAS
We now consider an arbitrary vector $w_\J \in \F_\J$, such that $\S_{X_\J X_\J} w_\J$ is different from zero (such vector exists because
$\S_{X_\J X_\J} \neq 0$, as we have assumed in \hypref{rkhs} that the variables are not constant). Since the range of $\S_{X_\J X_i} $ is included in the range of $\S_{X_\J X_\J} $~\citep{baker},
 there exists $v_i \in \F_i$ such that $\S_{X_\J X_i} v_i = \S_{X_\J X_\J} w_\J$. Note that since
$\S_{X_\J X_\J} w_\J$ is different from zero, we must have $\S_{X_i X_i}^{1/2} v_i \neq 0$.
We have:
\BEAS
 n^{1/2} \langle v_i,    \hS_{X_{i} Y } -  \hS_{X_{i} X_\J }  \hat{f}_\J \rangle_\Fi & 
 = & 
 n^{1/2} \langle v_i,  \hS_{X_{i } \varepsilon }  \rangle_\Fi +   
 \langle w_\J,   \S_{X_{\J} X_\J } n^{1/2} ( \fJ -  \hat{f}_\J  ) \rangle_\FJ + o_p(1) \\
& = & n^{1/2}  \langle v_i, \hS_{X_{i } \varepsilon }  \rangle_\Fi
+ \langle w_\J,  \mu_0 D f_\J - n^{1/2}  \hS_{X_\J  \varepsilon } \rangle_\FJ  + o_p(1)\\
& = &   \langle w_\J,  \mu_0 D f_\J  \rangle_\FJ
+ n^{1/2}  \langle v_i, \hS_{X_{i } \varepsilon }  \rangle_\Fi - n^{1/2}  \langle w_\J, \hS_{X_\J  \varepsilon }   \rangle_\FJ
 + o_p(1).
\EEAS
The random variable $ E_n = 
 n^{1/2}  \langle v_i, \hS_{X_{i } \varepsilon }  \rangle - n^{1/2}  \langle w_\J, \hS_{X_\J  \varepsilon }   \rangle$  is a U-statistic with square integrable kernel obtained from i.i.d. random vectors; it is thus asymptotically normal~\citep{VanDerVaart} and we simply need to
 compute its mean and variance. The mean is zero and a short calculation similar to the one found in the proof of Theorem~\ref{theo:theo2} in \myapp{theo2} shows that we have:
 \BEAS
   \E E_n^2 & \geqslant & 
 (1-1/n) \sm \langle v_i, \S_{X_i X_i} v_i \rangle_\Fi + \sm\langle w_\J, \S_{X_\J X_\J} w_\J \rangle_\FJ
  - 2   \sm\langle  v_i, \S_{X_i X_\J} w_\J \rangle_\Fi \\
  & = &  (1-1/n)  (  \sm\langle v_i, \S_{X_i X_i} v_i \rangle_\Fi 
  -  \sm \langle  v_i, \S_{X_i X_\J} w_\J \rangle_\Fi).
  \EEAS
  The operator $C_{X_\J X_\J}^{-1}  C_{X_\J X_i} $ has the same range as  $C_{X_\J X_\J}$ (because $C_{XX}$ is invertible),
  and is thus included in the closure of the range of $\Diag( \S_{X_j X_j}^{1/2})$~\citep{baker}. Thus, for any
  $ u \in \F_i$, $C_{X_\J X_\J}^{-1}  C_{X_\J X_i} u$
   can be expressed as a limit of terms of the form  $\Diag( \S_{X_j X_j}^{1/2}) t$ where $t \in \FJ$. We thus have
   that 
  \BEAS
\langle u, C_{X_i X_\J}  \Diag( \S_{X_j X_j}^{1/2}) w_\J \rangle_\Fi
=
\langle u, C_{X_i X_\J} C_{X_\J X_\J}^{-1} C_{X_\J X_\J} \Diag( \S_{X_j X_j}^{1/2}) w_\J \rangle_\Fi
  \EEAS
  can be expressed as a limit of terms of the form
\begin{multline*}
 \langle t,   \Diag( \S_{X_j X_j}^{1/2})  C_{X_\J X_\J} \Diag( \S_{X_j X_j}^{1/2}) w_\J \rangle_\FJ
  = \langle t,  \S_{X_\J X_\J} w_\J \rangle_\FJ  
  = \langle t,  \S_{X_\J X_i} v_i \rangle_\FJ  \\
  = 
  \langle t,  \Diag( \S_{X_j X_j}^{1/2}) C_{X_\J X_i}  \S_{X_i X_i}^{1/2}  v_i \rangle_\FJ 
  \to \langle u,  C_{X_i  X_\J}  C_{X_\J X_\J}^{-1} C_{X_\J X_i}  \S_{X_i X_i}^{1/2}  v_i \rangle_\Fi .
  \end{multline*}
  This implies that $C_{X_i X_\J}   \Diag( \S_{X_j X_j}^{1/2}) w_\J = 
  C_{X_i  X_\J} C_{X_\J X_\J}^{-1}  C_{X_\J X_i}  \S_{X_i X_i}^{1/2}  v_i$, and thus we have:
  \BEAS
 \E E_n^2  & \geqslant & 
   \sm\langle v_i, \S_{X_i X_i} v_i \rangle_\Fi  
   -  \sm \langle  v_i, \S_{X_i X_i}^{1/2} C_{X_i X_\J} \Diag( \S_{X_j X_j}^{1/2} ) w_\J \rangle_\Fi  \\
  & = & 
   \sm\langle v_i, \S_{X_i X_i} v_i \rangle_\Fi  
   -  \sm \langle  v_i, \S_{X_i X_i}^{1/2} C_{X_i X_\J} C_{X_\J X_\J}^{-1} C_{X_\J X_i} \S_{X_i X_i}^{1/2} v_i \rangle_\Fi  \\
  & = & \sm\langle  \S_{X_i X_i}^{1/2} v_i,   (\idm_\Fi - C_{X_i X_\J} C_{X_\J X_\J}^{-1} C_{X_\J X_i} ) \S_{X_i X_i}^{1/2}  v_i \rangle_\Fi.
 \EEAS
 By assumption \hypref{compact}, the operator $\idm_\Fi - C_{X_i X_\J} C_{X_\J X_\J}^{-1} C_{X_\J X_i}$ is lower bounded
 by a strictly positive constant times the identity matrix, and thus, since $\S_{X_i X_i}^{1/2} v_i \neq 0$, we have $\E E_n^2 > 0$.
 This implies that  $n^{1/2} \langle v_i,    \hS_{X_{i} Y } -  \hS_{X_{i} X_\J }  \hat{f}_\J \rangle$ converges to a normal distribution
 with strictly positive variance. Thus the probability
 $ \P \left(
  n^{1/2} \langle v_i,    \hS_{X_{i} Y } -  \hS_{X_{i} X_\J }  \hat{f}_\J \rangle_\Fi \geqslant d_i \| \hat{f}\|_d \| v_i \|_\Fi+ 1\right)$
  converges to a strictly positive limit (note that $\| \hat{f}\|_d$ can be replaced by $\| \f\|_d$ without changing the result). Since
  $ \mu_n n^{1/2} \to \mu_0 < \infty$, this implies
  that 
   $$ \P \left(
  \mu_n^{-1} \langle v_i,    \hS_{X_{i} Y } -  \hS_{X_{i} X_\J }  \hat{f}_\J \rangle_\Fi > d_i \| \hat{f}\|_d \| v_i \|_\Fi \right)$$ 
  is asymptotically strictly positive (i.e., has a strictly positive $\lim \inf$). Thus the optimality
  condition (\ref{eq:opt1-alt-cov}) is not satisfied
  with non vanishing probability, which is a contradiction and proves the proposition.

\end{proof}

We now go back to the proof of Theorem~\ref{theo:theo2-cov}.
We prove by contradiction, by assuming that there exists $i \in \J^c$ such that
$$ \frac{1}{d_i} \left\| \S_{X_i X_i }^{1/2} C_{X_i X_\J  }  C_{X_\J  X_\J }^{-1}
\Diag ( d_j/ \| \fj \|_\Fj ) \gJ \right\|_\Fi > 1 . $$
Since with probability tending to one $J(\hat{f})  = \J$, with probability tending to one, we have from 
optimality condition (\ref{eq:opt2-alt-cov}),
and the usual line of arguments (see \eq{alpha} in \myapp{theo2}) that for every $i \in \J^c$:
\BEAS
\hS_{X_i Y } -  \hS_{X_i X_\J }  \hat{f}_\J 
& = & 
\mu_n \hS_{X_{i} X_\J }
\left(  \hS_{X_\J  X_\J  } + 
\mu_n D_n\right)^{-1}   D_n \f
 \\
 & & + 
 \hS_{X_{i} \varepsilon}
-  \hS_{X_{i} X_\J }
\left(  \hS_{X_\J  X_\J  } +   
\mu_n D_n  \right)^{-1} \hS_{X_\J  \varepsilon},
 \EEAS
 where $D_n = \| \hat{f}\|_d \Diag  ( d_j/ \| \hat{f}_j \|) $.
Following the same argument as in the proof of
Theorem~\ref{theo:theo1-cov}, (and because $\mu_n n^{1/2} \to +\infty$ as a consequence of
Proposition~\ref{prop:sqrt}),
 the first term in the last expression (divided by $\mu_n$) converges to 
$$ v_i = \S_{X_i X_i }^{1/2} C_{X_i X_\J  }  C_{X_\J  X_\J }^{-1} \| \f \|_d
\Diag ( d_j/ \| \fj \|_\Fj  ) \gJ 
 $$
 By assumption $\|v_i\| > d_i \| \f \|_d$.
We have the second term:
\BEAS
& &  \hS_{X_{i} \varepsilon}
-  \hS_{X_{i} X_\J }
\left(  \hS_{X_\J  X_\J  } + 
\mu_n \| \hat{f} \|_d \Diag  ( d_j/ \| \hat{f}_j \|_\Fj) \right)^{-1} \hS_{X_\J  \varepsilon}
\\
& = & O_p(n^{-1/2})
-  \hS_{X_{i} X_\J }
\left(   \hS_{X_\J  X_\J  } + 
\mu_n \| \f \|_d \Diag  ( d_j/ \|  \fj \|_\Fj) \right)^{-1} \hS_{X_\J  \varepsilon} + O_p(n^{-1/2}).
\EEAS
The remaining term can be bounded as follows (with $D = \| \f \|_d \Diag  ( d_j/ \|  \fj \|_\Fj)$):
\BEAS
 & & \E \left(\left\|
\hS_{X_{i} X_\J }
\left(   \hS_{X_\J  X_\J  } + 
\mu_n D \right)^{-1} \hS_{X_\J  \varepsilon} \right\|^2_\Fi | \X \right)
\\
& \leqslant & \frac{\sM}{n} \tr
  \hS_{X_{i} X_\J }
\left(   \hS_{X_\J  X_\J  } + 
\mu_n D \right)^{-1}
\hS_{X_\J  X_\J} \left(   \hS_{X_\J  X_\J  } + 
\mu_n D \right)^{-1}\hS_{X_{\J} X_i } \\
& \leqslant & \frac{\sM}{n}
\tr \hS_{X_{i} X_{i} },
\EEAS
which implies that the full expectation is $O(n^{-1})$ (because our operators are trace-class, i.e., have finite trace). 
Thus the remaining term
is $O_p(n^{-1/2})$ and thus negligible compared to $\mu_n$, therefore
$\frac{1}{\mu_n \|\hat{f} \|_d} \left( \hS_{X_i Y } -  \hS_{X_i X_\J }  \hat{f}_\J \right) $ converges in probability to a limit which is
of norm strictly greater than $d_i$. Thus there is a non vanishing probability of being strictly larger than $d_i$,
which implies that 
with non vanishing probability, the optimality condition (\ref{eq:opt1-alt-cov}) is not satisfied, which is a contradiction.
 This concludes the proof.

\subsection{Proof of Proposition~\ref{prop:correst}}
\label{app:correst}
Note that the estimator defined in \eq{correst} is exactly equal to
$$\left\| \hS_{X_i X_\J  }  ( \hS_{X_\J  X_\J } +  \kappa_n \idm )^{-1}
\Diag ( d_j/ \| (\hat{f}^{LS}_{\kappa_n})_j \|_\Fj  ) (\hat{f}^{LS}_{\kappa_n})_\J  \right\|_\Fi.
$$
Using Proposition~\ref{prop:ls} and the arguments from \myapp{theo1-cov} by replacing $\tilde{f}$ by $\hat{F}_{LS}$, we get the consistency result.

\section{Proof of Results on Adaptive Group Lasso}
\subsection{Proof of Theorem~\ref{theo:adapGL}}

\label{app:adapGL}

We define $\tilde{w}$ as the minimizer of the same  cost function restricted to $w_{\J^c}=0$. Because
$\hat{w}^{LS}$ is consistent, the norms of $\hat{w}^{LS}_j$ for $j \in \J$ are bounded away from zero, and we get from standard results on M-estimation~\citep{VanDerVaart} the normal limit distribution with given covariance matrix if $\mu_n \ll n^{-1/2}$.

Moreover, the patterns of zeros (which is obvious by construction of $\tilde{w}$) converges in probability. What remains to be shown is that with probability tending to one, $\tilde{w}$ is optimal for the full problem.
We just need to show that with probability tending to one, for all $i \in \J^c$, 
\BEQ
\label{eq:ad2}
 \| \hS_{X_i \varepsilon } - \hS_{X_i X_\J} ( \tilde{w}_\J - w_\J )  \| \leqslant \mu_n \| \tilde{w} \|_d
 \| \hat{w}^{LS}_i\|^{-\gamma}.
\EEQ
Note that $\| \tilde{w} \|_d$ converges in probability to $\| \w\|_d >0$. Moreover,    $ \|\hat{w}^{LS}_i - \w_i \| = O_p(n^{-1/2})$. Thus,
if $ i \in \J^c$, i.e., if $\f_i=0$, then $\|\hat{w}^{LS}_i \| = O_p(n^{-1/2})$. The left hand side
in \eq{ad2}
is thus upper bounded by $O_p(n^{-1/2} )$ while the right hand side is lower bounded asymptotically by
$ \mu_n n^{\gamma/2}$. Thus if
$n^{-1/2} = o(\mu_n n^{\gamma/2} )$, then with probability tending to one we get the correct optimality condition, which concludes the proof.

\subsection{Proof of Proposition~\ref{prop:ls}}
\label{app:ls}
We have:
$$\hat{f}^{LS}_{\kappa_n} = \left( \hS_{XX} + \kappa_n \idm_\F  \right)^{-1} \hS_{X Y},$$ 
and thus:
\BEAS
\hat{f}^{LS}_{\kappa_n} - \f & = &  
\left( \hS_{XX} + \kappa_n \idm_\F \right)^{-1} \hS_{X X} \f - \f +
\left( \hS_{XX} + \kappa_n \idm_\F \right)^{-1} \hS_{X \varepsilon} \\
& = &   \left( \S_{XX} + \kappa_n \idm \right)^{-1} \S_{X X} \f  - \f + O_p(n^{-1/2} \kappa_n^{-1}) \mbox{ from Lemma~\ref{lemma:cov2} } \\
& = &  -\left( \S_{XX} + \kappa_n \idm_\F \right)^{-1} \kappa_n  \f  + O_p(n^{-1/2} \kappa_n^{-1}) .
\EEAS
Since $\f = \Sigma_{XX}^{1/2} \g$, we have $ \| -\left( \S_{XX} + \kappa_n \idm_\F \right)^{-1} \kappa_n  \f  \|_\F^2  \leqslant 
C \kappa_n \| \g \|_\F^2$,  which concludes the proof.

\subsection{Proof of Theorem~\ref{theo:adapt}}
\label{app:adapt}
We define $\tilde{f}$ as the minimizer of the same  cost function restricted to $f_{\J^c}=0$. Because
$\hat{f}^{LS}_{n^{-1/3}}$ is consistent, the norms of $(\hat{f}^{LS}_{n^{-1/3}})_j$ for $j \in \J$ are bounded away from zero, 
and Lemma~\ref{lemma:tilde-cov} applies with $\mu_n = \mu_0 n^{-1/3}$, i.e., $\tilde{f}$ 
converges in probability to $\f$ and so are the patterns of zeros (which is obvious by construction of $\tilde{f}$).
Moreover,  for any $\eta>0$, from Lemma~\ref{lemma:tilde-cov}, we have
$ \| \tilde{f}_\J   - f_\J \| = O_p(n^{-1/6 + \eta} )$ (because
$\mu_n^{-1/2} + n^{-1/2} \mu_n^{-1} = O_p(n^{-1/6})$).

What remains to be shown is that with probability tending to one, $\tilde{f}$ is optimal for the full problem.
We just need to show that with probability tending to one, for all $i \in \J^c$, 
\BEQ
\label{eq:ad}
 \| \hS_{X_i \varepsilon } - \hS_{X_i X_\J} ( \tilde{f}_\J - f_\J )  \| \leqslant \mu_n \| \tilde{f} \|_d
 \| (\hat{f}^{LS}_{n^{-1/3}})_i\|_\Fi^{-\gamma}.
\EEQ
Note that $\| \tilde{f} \|_d$ converges in probability to $\| \f \|_d >0$. Moreover, by Proposition~\ref{prop:ls},  $ \|( \hat{f}^{LS}_{n^{-1/3}})_i - \f_i \| = O_p(n^{-1/6})$. Thus,
if $ i \in \J^c$, i.e., if $\f_i=0$, then $\|( \hat{f}^{LS}_{n^{-1/3}})_i \|_\Fi= O_p(n^{-1/6})$. The left hand side
in \eq{ad}
is thus upper bounded by $O_p(n^{-1/2} + n^{-1/6 + \eta} )$ while the right hand side is lower bounded asymptotically by
$n^{-1/3} n^{\gamma/6}$. Thus if
$-1/6 + \eta < -1/3 + \gamma/6 $, then with probability tending to one we get the correct optimality condition.
As soon as $\gamma>1$, we can find $\eta$  small enough and strictly positive, which concludes the proof.

\subsection{Range Condition of Covariance Operators}
\label{app:range}

We let denote $C(q)$ the convolution operator by $q$ on the space of real functions on $\rb^p$ 
and $T(p)$ the pointwise multiplication by $p(x)$.
In this appendix,   we look at different Hilbertian products of functions
on $\rb^p$, we use the notations $\langle \cdot, \cdot \rangle_\F$ and $\langle \cdot, \cdot \rangle_{L^2(p_X)}$ and
$\langle \cdot, \cdot \rangle_{L^2(\rb^p)}$ for the dot products in the RKHS $\F$, the space $L^2(p_X)$ of square integrable functions
with respect to $p(x) dx$, and the space $L^2(\rb^p)$ of square integrable functions with respect to the Lebesgue measure.
With
our assumptions,  for
all $\tilde{f}, \tilde{g} \in L^2(\rb^p)$, we have:
$$\langle \tilde{f},\tilde{g}\rangle_{L^2} = \langle C(q)^{1/2} \tilde{f}, C(q)^{1/2}  \tilde{g}\rangle_{\F}.$$

Denote by $\{\lambda_k\}_{k \geq 1}$ and $\{e_k\}_{k \geq 1}$ the positive eigenvalues and the eigenvectors of the covariance operator $\S_{XX}$,
respectively. Note that since $p_X(x)$ was assumed to be strictly positive, all eigenvalues are strictly positive (the RKHS cannot contain any non zero constant functions on $\rb^p$).
 For $k \geqslant 1 $, set
$f_k= \lambda_k^{-1/2}   ( e_k  - \int_{\rb^p} e_k(x) p_X(x) dx)$. By construction, for any $k,\ell  \geqslant 1 $,
\begin{multline*}
\lambda_k \delta_{k,\ell} = \pscal{e_k}{\Sigma e_\ell}_{\F}= \int_{\rb^p} p_X(x) 
  ( e_k  - \textstyle \int_{\rb^p} e_k(x) p_X(x) dx)  ( e_\ell  - \int_{\rb^p} e_\ell(x) p_X(x) dx)  dx \\
= \displaystyle \lambda_k^{1/2} \lambda_\ell^{1/2} \int_{\rb^p} p_X(x) f_k(x) f_\ell(x) dx = \lambda_k^{1/2} \lambda_\ell^{1/2} \pscal{f_k}{f_\ell}_{L^2(p_X)}.
\end{multline*}
Thus  $\{f_k\}_{k \geqslant 1 }$ is an orthonormal sequence in $L^2(p_X)$. Let $f = C(q) g $ for $g \in L^2(\rb^p)$ such that $\int_{\rb^p} g(x) dx=0$. Note that
$f$ is in the range of $\S_{XX}^{1/2}$ if and only if $\langle f,\Sigma^{-1}f\rangle_\F$ is finite. We have:
\begin{multline*}
\langle f,\Sigma^{-1}f\rangle_\F = 
\sum_{p =1 }^\infty \lambda_p^{-1} \pscal{e_p}{f}_{\F}^2
= \sum_{p =1 }^\infty\lambda_p^{-1} \pscal{e_p}{g}_{L^2(\rb^p)}^2
= \sum_{p =1 }^\infty \lambda_p^{-1} \left( \int_{\rb^p} g(x) e_p(x) dx \right)^2 \\
= \sum_{p =1 }^\infty  \pscal{p_X^{-1}g}{f_p}_{L^2(p_X)}^2 \leqslant  \|p_X^{-1}g\|^2_{L^2(p_X)} = \int_{\rb^p} \frac{g^2(x)}{p_X(x)} dx ,
\end{multline*}
because   $\{f_k\}_{k \geqslant 1 }$ is an orthonormal sequence in $L^2(p_X)$. This concludes the proof.

\section{Gaussian Kernels and Gaussian Variables}
\label{app:covgauss}
In this section, we consider $X \in \rb^m$ with normal distribution with zero mean and covariance matrix $S$. We also  consider Gaussian kernels $k_j(x_j,x_j') = \exp( - b_i(x_j-x_j')^2)$ on each of its component. In this situation, we can find orthonormal basis of the Hilbert spaces $\F_j$ where we can compute the coordinates of all covariance operators. This thus allows to check conditions (\ref{eq:condition-cov}) or (\ref{eq:condition-cov-weak}) without using sampling.

We consider the eigenbasis of the non centered covariance operators on each $\mathcal{F}_j$, $j=1,\dots,m$, which is equal to~\citep{hermitegauss}:
$$ e_k^j(x_j) = (\lambda_k^j)^{1/2}\left( \frac{c_j^{1/2}}{a_j^{1/2} 2^k k!} \right)^{1/2} e^{-(c_j-a_j)u^2} H_k((2c_j)^{1/2} x_j)$$
with eigenvalues $\lambda_k^j = \left( \frac{2a_j}{A_j} \right)^{1/2} (B_j)^k$, where $a_i = 1 / 4 S_{ii}$, $c_j = (a_j^2+2a_j b_j)^{1/2}$, $A_j=a_j+b_j+c_j$ and $B_j = b_j / A_j$, and $H_k$  is the $k$-th Hermite polynomial.

We can then compute all required expectations as follows (note that by definition  we have
$\E e_k^j(X_j)^2 = \lambda_k^i$):
\BEAS
\E e_{2k+1}^j(X_j) &  = &  0
\\
\E e_{2k}^j(X_j) & = &   \left(   \lambda_{2k}^j \frac{2a_j^{1/2}  c_j^{1/2}   }{  (a_j+c_j) }{ 2k \choose k}  \right)^{1/2} \left( \frac{c_j - a_j}{2(c_j + a_j)} \right)^k
\EEAS
$$
\E e_k^j(X_j) e_\ell^i(X_i) = 
 \left(   \lambda_{2k}^j   \lambda_{2\ell}^i  \frac{  c_j^{1/2} c_i^{1/2}   }{ 
 a_j^{1/2}a_i^{1/2} 2^k 2^\ell k! \ell!  }   \right)^{1/2} \frac{ (S_{ii} S_{jj} - S_{ij}^2 )^{-1/2}}{4\pi c_i^{1/2} c_j^{1/2}
}  D_{k\ell}(Q_{ij}),
$$
 where $Q_{ij} = \left( \begin{array}{cc}
 \frac{1}{2}(1-a_i/c_i)  & 
0 \\ 
 0 &\frac{1}{2} (1-a_j/c_j)  \end{array}\right)
 + \frac{1}{4}
 \left( \begin{array}{cc}
 S_{ii} c_i & S_{ij} c_i^{1/2} c_j^{1/2} \\
 S_{ij} c_i^{1/2} c_j^{1/2} & S_{jj} c_j \end{array}\right)^{-1}
 $ and
 $$
D_{k\ell}(Q) = 
 \int_{\rb^2}
 \exp \left[ -
 \left( \begin{array}{c}u \\ v 
 \end{array}\right)^\top Q
 \left( \begin{array}{c} u \\ v
 \end{array}\right) \right]
  H_k(u)
H_\ell(v)  du dv,
$$
for any  positive matrix $Q$. For any given $Q$, $D_{k\ell}(Q)$ can be computed exactly by using a singular value decomposition of $Q$ and the appropriate change of variables.\footnote{Matlab code to compute $D_{k\ell}(Q)$ can be downloaded from the author's webpage.}


\bibliography{grouplasso}

\begin{thebibliography}{47}
\providecommand{\natexlab}[1]{#1}
\providecommand{\url}[1]{\texttt{#1}}
\expandafter\ifx\csname urlstyle\endcsname\relax
  \providecommand{\doi}[1]{doi: #1}\else
  \providecommand{\doi}{doi: \begingroup \urlstyle{rm}\Url}\fi

\bibitem[Bach(2007)]{tracenorm}
F.~R. Bach.
\newblock Consistency of trace norm minimization.
\newblock Technical Report 00179522, HAL, 2007.
\newblock URL \url{http://hal.archives-ouvertes.fr/hal-00200109/fr/}.

\bibitem[Bach et~al.(2004{\natexlab{a}})Bach, Lanckriet, and Jordan]{skm}
F.~R. Bach, G.~R.~G. Lanckriet, and M.~I. Jordan.
\newblock Multiple kernel learning, conic duality, and the {SMO} algorithm.
\newblock In \emph{Proceedings of the International Conference on Machine
  Learning (ICML)}, 2004{\natexlab{a}}.

\bibitem[Bach et~al.(2004{\natexlab{b}})Bach, Thibaux, and
  Jordan]{bach_thibaux}
F.~R. Bach, R.~Thibaux, and M.~I. Jordan.
\newblock Computing regularization paths for learning multiple kernels.
\newblock In \emph{Advances in Neural Information Processing Systems 17},
  2004{\natexlab{b}}.

\bibitem[Baker(1973)]{baker}
C.~Baker.
\newblock Joint measures and cross-covariance operators.
\newblock \emph{Transactions of the American Mathematical Society},
  186:\penalty0 273--289, 1973.

\bibitem[Berlinet and Thomas-Agnan(2003)]{rkhs}
A.~Berlinet and C.~Thomas-Agnan.
\newblock \emph{Reproducing Kernel {H}ilbert Spaces in Probability and
  Statistics}.
\newblock Kluwer Academic Publishers, 2003.

\bibitem[Bousquet and Herrmann(2003)]{bousquet}
O.~Bousquet and D.~J.~L. Herrmann.
\newblock On the complexity of learning the kernel matrix.
\newblock In \emph{Advances in Neural Information Processing Systems 17}, 2003.

\bibitem[Boyd and Vandenberghe(2003)]{boyd}
S.~Boyd and L.~Vandenberghe.
\newblock \emph{Convex Optimization}.
\newblock Cambridge Univ. Press, 2003.

\bibitem[Br\'emaud(1999)]{bremaud}
P.~Br\'emaud.
\newblock \emph{Markov chains, {G}ibbs fields, {M}onte {C}arlo simulation, and
  queues}.
\newblock Springer-Verlag, 1999.

\bibitem[Brezis(1980)]{brezis80analyse}
H.~Brezis.
\newblock \emph{Analyse Fonctionelle}.
\newblock Masson, 1980.

\bibitem[Caponnetto and de~Vito(2005)]{capo}
A.~Caponnetto and E.~de~Vito.
\newblock Fast rates for regularized least-squares algorithm.
\newblock Technical Report 248/AI Memo 2005-013, CBCL, Massachusetts Institute
  of Technology, 2005.

\bibitem[Cucker and Smale(2002)]{smale}
F.~Cucker and S.~Smale.
\newblock On the mathematical foundations of learning.
\newblock \emph{Bulletin of the American Mathematical Society}, 39\penalty0
  (1), 2002.

\bibitem[Durrett(2004)]{durrett}
R.~Durrett.
\newblock \emph{Probability: theory and examples}.
\newblock Duxbury Press, third edition, 2004.

\bibitem[Efron et~al.(2004)Efron, Hastie, Johnstone, and Tibshirani]{lars}
B.~Efron, T.~Hastie, I.~Johnstone, and R.~Tibshirani.
\newblock Least angle regression.
\newblock \emph{Annals of Statistics}, 32:\penalty0 407, 2004.

\bibitem[Fu and Knight(2000)]{fu}
W.~Fu and K.~Knight.
\newblock Asymptotics for {L}asso-type estimators.
\newblock \emph{Annals of Statistics}, 28\penalty0 (5):\penalty0 1356--1378,
  2000.

\bibitem[Fukumizu et~al.(2004)Fukumizu, Bach, and Jordan]{fukumizu}
K.~Fukumizu, F.~R. Bach, and M.~I. Jordan.
\newblock Dimensionality reduction for supervised learning with reproducing
  kernel {H}ilbert spaces.
\newblock \emph{Journal of Machine Learning Research}, 5:\penalty0 73--99,
  2004.

\bibitem[Fukumizu et~al.(2007)Fukumizu, Bach, and Gretton]{kenji}
K.~Fukumizu, F.~R. Bach, and A.~Gretton.
\newblock Statistical convergence of kernel canonical correlation analysis.
\newblock \emph{Journal of Machine Learning Research}, 8\penalty0 (8), 2007.

\bibitem[Gretton et~al.(2005)Gretton, Herbrich, Smola, Bousquet, and
  Sch{\"o}lkopf]{gretton}
A.~Gretton, R.~Herbrich, A.~Smola, O.~Bousquet, and B.~Sch{\"o}lkopf.
\newblock Kernel methods for measuring independence.
\newblock \emph{Journal of Machine Learning Research}, 6:\penalty0 2075--2129,
  12 2005.

\bibitem[Harchaoui and Bach(2007)]{graphkernel}
Z.~Harchaoui and F.~R. Bach.
\newblock Image classification with segmentation graph kernels.
\newblock In \emph{Proceedings of the Conference on Computer Vision and Pattern
  Recognition (CVPR)}, 2007.

\bibitem[Hastie et~al.(2001)Hastie, Tibshirani, and Friedman]{hastie}
T.~Hastie, R.~Tibshirani, and J.~Friedman.
\newblock \emph{The Elements of Statistical Learning}.
\newblock Springer-Verlag, 2001.

\bibitem[Hastie and Tibshirani(1990)]{hastie_GAM}
T.~J. Hastie and R.~J. Tibshirani.
\newblock \emph{Generalized Additive Models}.
\newblock Chapman \& Hall, 1990.

\bibitem[Juditsky and Nemirovski(2000)]{agreg}
A.~Juditsky and A.~Nemirovski.
\newblock Functional aggregation for nonparametric regression.
\newblock \emph{Annals of Statistics}, 28\penalty0 (3):\penalty0 681--712,
  2000.

\bibitem[Lanckriet et~al.(2004{\natexlab{a}})Lanckriet, Bie, Cristianini,
  Jordan, and Noble]{genomic_fusion}
G.~R.~G. Lanckriet, T.~De Bie, N.~Cristianini, M.~I. Jordan, and W.~S. Noble.
\newblock A statistical framework for genomic data fusion.
\newblock \emph{Bioinformatics}, 20:\penalty0 2626--2635, 2004{\natexlab{a}}.

\bibitem[Lanckriet et~al.(2004{\natexlab{b}})Lanckriet, Cristianini, Ghaoui,
  Bartlett, and Jordan]{gert}
G.~R.~G. Lanckriet, N.~Cristianini, L.~El Ghaoui, P.~Bartlett, and M.~I.
  Jordan.
\newblock Learning the kernel matrix with semidefinite programming.
\newblock \emph{Journal of Machine Learning Research}, 5:\penalty0 27--72,
  2004{\natexlab{b}}.

\bibitem[Lobo et~al.(1998)Lobo, Vandenberghe, Boyd, and L{\'e}bret]{socp}
M.~S. Lobo, L.~Vandenberghe, S.~Boyd, and H.~L{\'e}bret.
\newblock Applications of second-order cone programming.
\newblock \emph{Linear Algebra and its Applications}, 284:\penalty0 193--228,
  1998.

\bibitem[McAuley et~al.(2005)McAuley, Ming, Stewart, and Hanna]{speech}
J.~McAuley, J.~Ming, D.~Stewart, and P.~Hanna.
\newblock Subband correlation and robust speech recognition.
\newblock \emph{IEEE Transactions on Speech and Audio Processing}, 13\penalty0
  (5):\penalty0 956--964, 2005.

\bibitem[Meier et~al.(2006)Meier, van~de Geer, and B{\"u}hlmann]{grouplog}
L.~Meier, S.~van~de Geer, and P.~B{\"u}hlmann.
\newblock The group {L}asso for logistic regression.
\newblock Technical Report 131, Eidgen{\"o}ossische Technische Hochschule
  (ETH), Z{\"u}rich, Switzerland, 2006.

\bibitem[Meinshausen and Yu(2006)]{yuinfinite}
N.~Meinshausen and B.~Yu.
\newblock {L}asso-type recovery of sparse representations for high-dimensional
  data.
\newblock Technical Report 720, Departement of Statisics, UC Berkeley, 2006.

\bibitem[Osborne et~al.(2000)Osborne, Presnell, and Turlach]{osborne}
M.~R. Osborne, B.~Presnell, and B.~A. Turlach.
\newblock On the lasso and its dual.
\newblock \emph{Journal of Computational and Graphical Statistics}, 9\penalty0
  (2):\penalty0 319--337, 2000.

\bibitem[Rakotomamonjy et~al.(2007)Rakotomamonjy, Bach, Canu, and
  Grandvalet]{rakoto}
A.~Rakotomamonjy, F.~R. Bach, S.~Canu, and Y.~Grandvalet.
\newblock More efficiency in multiple kernel learning.
\newblock In \emph{Proceedings of the International Conference on Machine
  Learning (ICML)}, 2007.

\bibitem[Ravikumar et~al.(2008)Ravikumar, Liu, Lafferty, and Wasserman]{spam}
P.~Ravikumar, H.~Liu, J.~Lafferty, and L.~Wasserman.
\newblock {SpAM}: Sparse additive models.
\newblock In \emph{Advances in Neural Information Processing Systems 22}, 2008.

\bibitem[Renyi(1959)]{renyi}
A.~Renyi.
\newblock {On Measures of Dependence}.
\newblock \emph{Acta Mathematica Academy Sciences Hungary}, 10:\penalty0
  441--451, 1959.

\bibitem[Sch{\"o}lkopf and Smola(2001)]{scholkopf-smola-book}
B.~Sch{\"o}lkopf and A.~J. Smola.
\newblock \emph{Learning with Kernels}.
\newblock MIT Press, 2001.

\bibitem[Sonnenburg et~al.(2006)Sonnenburg, R{\"a}tsch, Sch{\"a}fer, and
  Sch{\"o}lkopf]{sonnenburg}
S.~Sonnenburg, G.~R{\"a}tsch, C.~Sch{\"a}fer, and B.~Sch{\"o}lkopf.
\newblock Large scale multiple kernel learning.
\newblock \emph{Journal of Machine Learning Research}, 7:\penalty0 1531--1565,
  07 2006.

\bibitem[Steinwart(2001)]{steinwart}
I.~Steinwart.
\newblock On the influence of the kernel on the consistency of support vector
  machines.
\newblock \emph{Journal of Machine Learning Research}, 2:\penalty0 67--93,
  2001.

\bibitem[Tibshirani(1994)]{lasso}
R.~Tibshirani.
\newblock Regression shrinkage and selection via the {l}asso.
\newblock \emph{Journal of The Royal Statistical Society Series B}, 58\penalty0
  (1):\penalty0 267--288, 1994.

\bibitem[Tikhonov and Arsenin(1997)]{tikhonov}
A.~N. Tikhonov and V.~Y. Arsenin.
\newblock \emph{Solutions of ill-posed problems}.
\newblock V. H. Winston and Sons, 1997.

\bibitem[{Van der Vaart}(1998)]{VanDerVaart}
A.~W. {Van der Vaart}.
\newblock \emph{Asymptotic Statistics}.
\newblock Cambridge Univ. Press, 1998.

\bibitem[Varma and Ray(2007)]{Varma07c}
M.~Varma and D.~Ray.
\newblock Learning the discriminative power-invariance trade-off.
\newblock In \emph{Proceedings of the {IEEE} International Conference on
  Computer Vision (CVPR)}, 2007.

\bibitem[Wahba(1990)]{wahba}
G.~Wahba.
\newblock \emph{Spline Models for Observational Data}.
\newblock SIAM, 1990.

\bibitem[Wainwright(2006)]{martin}
M.~J. Wainwright.
\newblock Sharp thresholds for noisy and high-dimensional recovery of sparsity
  using $\ell_1$-constrained quadratic programming.
\newblock Technical Report 709, Department of Statistics, UC Berkeley, 2006.

\bibitem[Wu et~al.(2007)Wu, Ying, and Zhou]{multi}
Q.~Wu, Y.~Ying, and D.-X. Zhou.
\newblock Multi-kernel regularized classifiers.
\newblock \emph{Journal of Complexity}, 23\penalty0 (1):\penalty0 108--134,
  2007.

\bibitem[Yuan and Lin(2006)]{grouped}
M.~Yuan and Y.~Lin.
\newblock Model selection and estimation in regression with grouped variables.
\newblock \emph{Journal of The Royal Statistical Society Series B}, 68\penalty0
  (1):\penalty0 49--67, 2006.

\bibitem[Yuan and Lin(2007)]{yuanlin}
M.~Yuan and Y.~Lin.
\newblock On the non-negative garrotte estimator.
\newblock \emph{Journal of The Royal Statistical Society Series B}, 69\penalty0
  (2):\penalty0 143--161, 2007.

\bibitem[Zhao and Yu(2006)]{Zhaoyu}
P.~Zhao and B.~Yu.
\newblock On model selection consistency of {L}asso.
\newblock \emph{Journal of Machine Learning Research}, 7:\penalty0 2541--2563,
  2006.

\bibitem[Zhou and Burges(2007)]{views}
D.~Zhou and C.~J.~C. Burges.
\newblock Spectral clustering and transductive learning with multiple views.
\newblock In \emph{Proceedings of the International Conference on Machine
  Learning (ICML)}, 2007.

\bibitem[Zhu et~al.(1998)Zhu, Williams, Rohwer, and Morciniec]{hermitegauss}
H.~Zhu, C~K.~I. Williams, R.~Rohwer, and M.~Morciniec.
\newblock Gaussian regression and optimal finite dimensional linear models.
\newblock In \emph{Neural Networks and Machine Learning}. Springer-Verlag,
  1998.

\bibitem[Zou(2006)]{zou}
H.~Zou.
\newblock The adaptive l{L}sso and its oracle properties.
\newblock \emph{Journal of the American Statistical Association}, 101:\penalty0
  1418--1429, December 2006.

\end{thebibliography}

\end{document}